%% file: main.tex
\documentclass[manuscript,screen]{acmart}

\usepackage[utf8]{inputenc}
\usepackage[T1]{fontenc}
\usepackage{setspace}
\usepackage{graphicx}
\graphicspath{ {./figures/} }
\usepackage{subcaption}
\usepackage{amsmath}
\usepackage{lineno}
\usepackage{paralist}
\usepackage{xspace} 
\usepackage{tcolorbox}
\usepackage{algorithm}
\usepackage{algpseudocode}
\usepackage{url} 
\usepackage{bm}
\usepackage{amsfonts}
\usepackage{booktabs}
\usepackage{graphicx}
\usepackage{multirow}
\usepackage{hyperref}
\usepackage{todonotes}
\usepackage{accents}
\usepackage{mathtools}
\usepackage{longtable}
\usepackage{threeparttable}
\usepackage{threeparttablex}
\usepackage{pdflscape}
\usepackage{color}
\usepackage{float}
\usepackage{array}
\usepackage{tabto}
\usepackage{diagbox}
\usepackage{makecell}
\usepackage{mfirstuc}
\usepackage{boldline}
\usepackage{enumitem}
\usepackage{realboxes}
\usepackage{tabularx}

\newcolumntype{L}[1]{>{\raggedright\let\newline\\\arraybackslash\hspace{0pt}}m{#1}}
\newcolumntype{C}[1]{>{\centering\let\newline\\\arraybackslash\hspace{0pt}}m{#1}}
\newcolumntype{R}[1]{>{\raggedleft\let\newline\\\arraybackslash\hspace{0pt}}m{#1}}

\algdef{SE}{Begin}{End}{\textbf{begin}}{\textbf{end}}

\newcommand{\pvec}[1]{\vec{#1}\mkern2mu\vphantom{#1}}
\makeatletter
\renewenvironment{cases}[1][\lbrace]{%
	\def\@ldelim{#1}
	\matrix@check\cases\env@cases
}{%
	\endarray\right.%
}
\patchcmd{\env@cases}{\lbrace}{\@ldelim}{}{}
\makeatother

\DeclareUnicodeCharacter{0301}{\'{e}}
\DeclareUnicodeCharacter{0308}{\"{i}}

\newcommand{\mh}{metaheuristic\xspace}
\newcommand{\mhs}{metaheuristics\xspace}
\newcommand{\Mh}{Metaheuristic\xspace}
\newcommand{\LS}{LS\xspace}

\newcommand{\ls}{local search\xspace}

\newcommand{\pbmhs}{population-based \mhs}
\newcommand{\Pbmhs}{Population-based \mhs}

\newcommand{\op}{optimization problem\xspace}
\newcommand{\ops}{optimization problems\xspace}

\newcommand{\cop}{continuous \op}
\newcommand{\cops}{continuous \ops}

\newcommand{\Cops}{Continuous \ops}
\newcommand{\Ad}{Automatic design\xspace}
\newcommand{\ad}{automatic design\xspace}
\newcommand{\aacts}{automatic algorithm configuration tools\xspace}
\newcommand{\aact}{automatic algorithm configuration tool\xspace}

\newcommand{\AACT}{AACT\xspace}

\newcommand{\msf}{\mh software framework\xspace}
\newcommand{\Msfs}{\Mh software frameworks\xspace}
\newcommand{\msfs}{\mh software frameworks\xspace}

\newcommand{\MSF}{MSF\xspace}
\newcommand{\MSFs}{MSFs\xspace}

\newcommand{\Tbl}{Table\xspace}
\newcommand{\Tbls}{Tables\xspace}
\newcommand{\Alg}{Algorithm\xspace}

\newcommand{\Sect}{Section\xspace}

\newcommand{\Eq}{Equation\xspace}

\newcommand{\etal}{\textit{et al.}\xspace}

\newcommand{\PSOX}{PSO\textsl{-X}\xspace}
\newcommand{\irace}{\texttt{irace}\xspace}
\newcommand{\aCMAES}{autoCMAES\xspace}
\newcommand{\aDE}{autoDE\xspace}
\newcommand{\modPSODE}{modPSODE\xspace}

\newcommand{\aco}{ant colony optimization\xspace}
\newcommand{\ACO}{ACO\xspace}

\newcommand{\PSOall}{Particle Swarm Optimization\xspace}
\newcommand{\pso}{particle swarm optimization\xspace}
\newcommand{\PSO}{PSO\xspace}

\newcommand{\abc}{artificial bee colony\xspace}

\newcommand{\ec}{evolutionary computation\xspace}
\newcommand{\EC}{EC\xspace}

\newcommand{\ESall}{Evolution Strategy\xspace}

\newcommand{\ES}{ES\xspace}

\newcommand{\ess}{evolution strategies\xspace}
\newcommand{\ESs}{ESs\xspace}

\newcommand{\gas}{genetic algorithms\xspace}
\newcommand{\GAs}{GAs\xspace}

\newcommand{\ssii}{swarm intelligence\xspace}
\newcommand{\SSII}{SI\xspace}

\newcommand{\DEall}{Differential Evolution\xspace}
\newcommand{\de}{differential evolution\xspace}
\newcommand{\DE}{DE\xspace}
\newcommand{\CMAESall}{Covariance Matrix Adaptation--\ESall}

\newcommand{\cmaes}{covariance matrix adaptation--\ES}
\newcommand{\CMAES}{CMA--\ES}
\newcommand{\mtsls}{multiple trajectory search \ls}

\newcommand{\MTSLS}{MTS\textsf{ls}\xspace}

\newcommand{\stanPSO}{standard \PSO}
\newcommand{\StanPSO}{Standard \PSO}
\newcommand{\SPSO}{Sta\PSO}
\newcommand{\PsoParam}{$\omega$, $\varphi_1$ and $\varphi_2$\xspace}

\newcommand{\StdPSO}{StdPSO\xspace}

\newcommand{\GwoAll}{Grey Wolf Optimizer\xspace}

\newcommand{\WoaAll}{Whale Optimization Algorithm\xspace}

\newcommand{\MfAll}{Moth-Flame Algorithm\xspace}

\newcommand{\Topo}{{\textsf{Topology}}\xspace}
\newcommand{\Moi}{{\textsf{Model of influence}}\xspace}
\newcommand{\Pop}{{\textsf{Population}}\xspace}
\newcommand{\Init}{{\textsf{Initialization}}\xspace}
\newcommand{\DNPP}{{\textsf{DNPP}}\xspace}
\newcommand{\PertRnd}{{\textsf{Pert$_{\mathrm{\textsf{rand}}}$}}\xspace}
\newcommand{\PertInf}{{\textsf{Pert$_{\mathrm{\textsf{info}}}$}}\xspace}
\newcommand{\Mtx}{{\textsf{Mtx}}\xspace}

\newcommand{\PopConst}{{\textsf{Pop-constant}}\xspace}
\newcommand{\PopIncre}{{\textsf{Pop-incremental}}\xspace}
\newcommand{\PopTV}{{\textsf{Pop-time-varying}}\xspace}
\newcommand{\InitRandom}{{\textsf{Init-random}}\xspace}
\newcommand{\InitHorizontal}{{\textsf{Init-horizontal}}\xspace}
\newcommand{\DNPPRect}{{\textsf{DNPP-rectangular}}\xspace}
\newcommand{\DNPPSphe}{{\textsf{DNPP-spherical}}\xspace}

\newcommand{\OperatorS}{{\textsf{DNPP-standard}}\xspace}
\newcommand{\OperatorD}{{\textsf{DNPP-discrete}}\xspace}
\newcommand{\OperatorN}{{\textsf{DNPP-Gaussian}}\xspace}
\newcommand{\OperatorCG}{{\textsf{DNPP-Cauchy--Gaussian}}\xspace}
\newcommand{\OperatorQrandNeigh}{$\textsf{random\_neighbor}$\xspace}

\newcommand{\PertGau}{{\textsf{\PertInf-Gaussian}}\xspace}
\newcommand{\PertLev}{{\textsf{\PertInf-L{\'e}vy}}\xspace}
\newcommand{\PertUni}{{\textsf{\PertInf-uniform}}\xspace}
\newcommand{\PertRect}{{\textsf{\PertRnd-rectangular}}\xspace}
\newcommand{\PertNoi}{{\textsf{\PertRnd-noisy}}\xspace}
\newcommand{\MagCons}{{\textsf{PM-constant value}}\xspace}
\newcommand{\MagEucli}{{\textsf{PM-Euclidean distance}}\xspace}
\newcommand{\MagOFd}{{\textsf{PM-obj.func. distance}}\xspace}
\newcommand{\MagSucc}{{\textsf{PM-success rate}}\xspace}

\newcommand{\MtxDiagonal}{{\textsf{\Mtx-random diagonal}}\xspace}
\newcommand{\MtxLinear}{{\textsf{\Mtx-random linear}}\xspace}
\newcommand{\MtxExponential}{{\textsf{\Mtx-exponential map}}\xspace}

\newcommand{\MtxEuclideanOne}{{\textsf{\Mtx-Euclidean rotation$_{one}$}}\xspace}
\newcommand{\MtxEuclideanAll}{{\textsf{\Mtx-Euclidean rotation$_{all}$}}\xspace}
\newcommand{\MtxIncreasingGroupBased}{{\textsf{\Mtx-Increasing group-based}}\xspace}

\newcommand{\PhiConstant}{{\textsf{AC-constant}}\xspace}
\newcommand{\PhiRandom}{{\textsf{AC-random}}\xspace}
\newcommand{\PhiTV}{{\textsf{AC-time-varying}}\xspace}
\newcommand{\PhiExtra}{{\textsf{AC-extrapolated}}\xspace}
\newcommand{\MoiBoN}{{\textsf{MoI-best-of-neighborhood}}\xspace}
\newcommand{\MoiFI}{{\textsf{MoI-fully informed}}\xspace}
\newcommand{\MoiRFI}{{\textsf{MoI-ranked fully informed}}\xspace}

\newcommand{\TopFC}{{\textsf{Top-fully-connected}}\xspace}
\newcommand{\TopRing}{{\textsf{Top-ring}}\xspace}
\newcommand{\TopWheel}{{\textsf{Top-wheel}}\xspace}
\newcommand{\TopRandom}{{\textsf{Top-random edge}}\xspace}
\newcommand{\TopNeum}{{\textsf{Top-Von Neumann}}\xspace}
\newcommand{\TopHie}{{\textsf{Top-hierarchical}}\xspace}
\newcommand{\TopTVFI}{{\textsf{Top-time-varying}}\xspace}

\newcommand{\OmegaCons}{{\textsf{IW-constant}}\xspace}
\newcommand{\OmegaLinDec}{{\textsf{IW-linear decreasing}}\xspace}
\newcommand{\OmegaLinInc}{{\textsf{IW-linear increasing}}\xspace}
\newcommand{\OmegaRnd}{{\textsf{IW-random}}\xspace}
\newcommand{\OmegaSelfReg}{{\textsf{IW-self-regulating}}\xspace}
\newcommand{\OmegaAdapVel}{{\textsf{IW-adaptive based on velocity}}\xspace}
\newcommand{\OmegaDoubExp}{{\textsf{IW-double exponential self-adaptive}}\xspace}
\newcommand{\OmegaRnkBsd}{{\textsf{IW-rank-based}}\xspace}
\newcommand{\OmegaSuccBsd}{{\textsf{IW-success-based}}\xspace}
\newcommand{\OmegaConvBsd}{{\textsf{IW-convergence-based}}\xspace}
\newcommand{\VelClp}{{\textsf{Velocity Clamping}}\xspace}
\newcommand{\StagDet}{{\textsf{Unstuck Velocity}}\xspace} 
\newcommand{\IgnPbest}{\textsf{Ignore Pbest}\xspace} 

\newcommand{\CMAESMtx}{{\textsf{CMAES-\Mtx}}\xspace} 
\newcommand{\CMAESMtxDia}{{\textsf{CMAES-\Mtx-diagonal}}\xspace} 
\newcommand{\CMAESMtxCov}{{\textsf{CMAES-\Mtx-covariance}}\xspace} 
\newcommand{\CMAESMtxCovDia}{{\textsf{CMAES-\Mtx-cov-then-diag}}\xspace} 
\newcommand{\CMAESPop}{{\textsf{CMAES-population}}\xspace}
\newcommand{\CMAESPopConst}{{\textsf{CMAES-pop-constant}}\xspace}
\newcommand{\CMAESPopInc}{{\textsf{CMAES-pop-incremental}}\xspace}
\newcommand{\CMAESRW}{{\textsf{CMAES-recombination-weights}}\xspace} 
\newcommand{\CMAESRWlog}{{\textsf{CMAES-RW-logarithmic}}\xspace} 
\newcommand{\CMAESRWlindec}{{\textsf{CMAES-RW-linear-decreasing}}\xspace} 
\newcommand{\CMAESRWeq}{{\textsf{CMAES-RW-equal}}\xspace} 
\newcommand{\CMAESRes}{{\textsf{CMAES-restart}}\xspace}
\newcommand{\CMAESpara}{{\textsf{cmaes\_par\_a}}\xspace} 
\newcommand{\CMAESparb}{{\textsf{cmaes\_par\_b}}\xspace} 
\newcommand{\CMAESparc}{{\textsf{cmaes\_par\_c}}\xspace} 
\newcommand{\CMAESpard}{{\textsf{cmaes\_par\_d}}\xspace} 
\newcommand{\CMAESpare}{{\textsf{cmaes\_par\_e}}\xspace} 
\newcommand{\CMAESparf}{{\textsf{cmaes\_par\_f}}\xspace} 
\newcommand{\CMAESparg}{{\textsf{cmaes\_par\_g}}\xspace} 

\newcommand{\DEBaseVect}{{\textsf{DE-Base Vector}}\xspace}

\newcommand{\DEBVrand}{{\textsf{BV-random}}\xspace}
\newcommand{\DEBVbest}{{\textsf{BV-best}}\xspace}
\newcommand{\DEBVttbest}{{\textsf{BV-target-to-best}}\xspace}
\newcommand{\DEBVdirRand}{{\textsf{BV-directed-random}}\xspace}
\newcommand{\DEBVdirBest}{{\textsf{BV-directed-best}}\xspace}
\newcommand{\DERcb}{{\textsf{DE-Recombination}}\xspace}
\newcommand{\DERcbcomp}{{\textsf{RCB}}\xspace}
\newcommand{\DERcbBinomial}{{\textsf{RCB-binomial}}\xspace}
\newcommand{\DERcbExponential}{{\textsf{RCB-exponential}}\xspace}

\newcommand{\DEVect}{{\textsf{DE-Vectors}}\xspace}
\newcommand{\DEVectPos}{{\textsf{V-positions}}\xspace}
\newcommand{\DEVectPbest}{{\textsf{V-pbest}}\xspace}
\newcommand{\DEVectMix}{{\textsf{V-mixture}}\xspace}

\newcommand{\DERecompVel}{{\textsf{DE-Recompute Velocity}}\xspace}
\newcommand{\DERVnone}{{\textsf{RV-none}}\xspace}
\newcommand{\DERVgoBack}{{\textsf{RV-goBack}}\xspace}
\newcommand{\DERVrandom}{{\textsf{RV-random}}\xspace}
\newcommand{\DERVposition}{{\textsf{RV-position}}\xspace}

\newcommand{\PSOonFail}{{\textsf{PSO-only-on-fail}}\xspace}

\newcommand{\DEVectDifferences}{{\textsf{DE-vector-differences}}\xspace}

\newcommand{\LSbudget}{{\textsf{LS-budget}}\xspace}
\newcommand{\LSdivide}{{\textsf{LS-divide}}\xspace}
\newcommand{\MTSLSinitss}{{\textsf{MTSLS-init-ss}}\xspace}

\newcommand{\MTSLSiterations}{{\textsf{MTSLS-iterations}}\xspace}
\newcommand{\MTSLSbias}{{\textsf{MTSLS-bias}}\xspace}

\newcommand{\VectBasis}{{\textsf{Vector-Basis}}\xspace}

\newcommand{\VBeigen}{{\textsf{VB-eigenvector}}\xspace}
\newcommand{\VBnatural}{{\textsf{VB-natural}}\xspace}

\newcommand{\Reinit}{{\textsf{Re-initialization}}\xspace}
\newcommand{\RIchange}{{\textsf{RI-change}}\xspace}
\newcommand{\RIsimilarity}{{\textsf{RI-similarity}}\xspace}

\newcommand{\algInPaper}{\PSO, \CMAES and \DE}
\newcommand{\algInPaperLS}{\PSO, \CMAES, \DE and \LS algorithms\xspace}
\newcommand{\algInPaperHyb}{\PDHyb, \DCHyb and \PCHyb}
\newcommand{\MetafoR}{\textsf{\textbf{METAFO}}$\mathbb{R}$\xspace}
\newcommand{\PDHyb}{\PSO--\DE hybrid\xspace}
\newcommand{\PDHybs}{\PSO--\DE hybrids\xspace}
\newcommand{\DCHyb}{\DE--\CMAES hybrid\xspace}
\newcommand{\DCHybs}{\DE--\CMAES hybrids\xspace}
\newcommand{\PCHyb}{\PSO--\CMAES hybrid\xspace}
\newcommand{\PCHybs}{\PSO--\CMAES hybrids\xspace}
\newcommand{\MTF}{MTF\xspace}

\newcommand{\PSOmod}{{\textsf{\textbf{PSOmod}}}\xspace}
\newcommand{\DEmod}{{\textsf{\textbf{DEmod}}}\xspace}
\newcommand{\CMAESmod}{{\textsf{\textbf{CMA-ESmod}}}\xspace}
\newcommand{\LSmod}{{\textsf{\textbf{LSmod}}}\xspace}
\newcommand{\MetafoRModules}{\PSOmod, \DEmod and \CMAESmod}
\newcommand{\MetafoRModulesAlg}{\PSOmod, \DEmod, \CMAESmod, \LSmod}
\newcommand{\Execution}{{\textsf{Execution}}\xspace}
\newcommand{\ExecutionCB}{{\textsf{EXE-component-based}}\xspace}
\newcommand{\ExecutionPB}{{\textsf{EXE-probabilistic}}\xspace}
\newcommand{\ExecutionPBuni}{{\textsf{EXE-PB-uniform}}\xspace}
\newcommand{\ExecutionPBnor}{{\textsf{EXE-PB-normal}}\xspace}
\newcommand{\ExecutionPBlev}{{\textsf{EXE-PB-levy}}\xspace}
\newcommand{\ExecutionParStd}{$\textsf{par\_std}$\xspace}
\newcommand{\ExecutionMP}{{\textsf{EXE-multiple phases}}\xspace}
\newcommand{\IWcomp}{{\textsf{IW}}\xspace}
\newcommand{\ACcomp}{{\textsf{AC}}\xspace}
\newcommand{\PMcomp}{{\textsf{PM}}\xspace}

\newcommand{\TFOUT}{{\texttt{25OUT}}\xspace}

\newcommand{\LDO}{{\texttt{LDOUT}}\xspace}
\newcommand{\PSOtfo}{PSO-\textsl{X}$_{\TFOUT}$\xspace}
\newcommand{\DEtfo}{DE$_{\TFOUT}$\xspace}
\newcommand{\CMAEStfo}{CMAES$_{\TFOUT}$\xspace}
\newcommand{\PDHybtfo}{PSO--DE$_{\TFOUT}$\xspace}
\newcommand{\DCHybtfo}{DE--CMAES$_{\TFOUT}$\xspace}
\newcommand{\PCHybtfo}{PSO--CMAES$_{\TFOUT}$\xspace}
\newcommand{\MTFtfo}{MTF$_{\TFOUT}$\xspace}
\newcommand{\PSOldo}{PSO-\textsl{X}$_{\LDO}$\xspace}
\newcommand{\DEldo}{DE$_{\LDO}$\xspace}
\newcommand{\CMAESldo}{CMAES$_{\LDO}$\xspace}
\newcommand{\PDHybldo}{PSO--DE$_{\LDO}$\xspace}
\newcommand{\DCHybldo}{DE--CMAES$_{\LDO}$\xspace}
\newcommand{\PCHybldo}{PSO--CMAES$_{\LDO}$\xspace}
\newcommand{\MTFldo}{MTF$_{\LDO}$\xspace}
\newcommand{\StdPSOdft}{StdPSO$_{\texttt{DFT}}$\xspace}
\newcommand{\DEdft}{DE$_{\texttt{DFT}}$\xspace}
\newcommand{\CMAESdft}{CMAES$_{\texttt{DFT}}$\xspace}
\newcommand{\PSOtfoTBL}{PSO-\textsl{X}\xspace({\TFOUT})\xspace}
\newcommand{\DEtfoTBL}{DE\newline({\TFOUT})\xspace}
\newcommand{\CMAEStfoTBL}{CMAES\xspace({\TFOUT})\xspace}
\newcommand{\PDHybtfoTBL}{PSO--\newline DE\xspace({\TFOUT})\xspace}
\newcommand{\DCHybtfoTBL}{DE--CMAES\xspace({\TFOUT})\xspace}
\newcommand{\PCHybtfoTBL}{PSO--CMAES\xspace({\TFOUT})\xspace}
\newcommand{\MTFtfoTBL}{MTF\xspace({\TFOUT})\xspace}
\newcommand{\PSOldoTBL}{PSO-\textsl{X}\xspace(\LDO)\xspace}
\newcommand{\DEldoTBL}{DE\newline(\LDO)\xspace}
\newcommand{\CMAESldoTBL}{CMAES\xspace(\LDO)\xspace}
\newcommand{\PDHybldoTBL}{PSO--\newline DE\newline(\LDO)\xspace}
\newcommand{\DCHybldoTBL}{DE--CMAES\xspace(\LDO)\xspace}
\newcommand{\PCHybldoTBL}{PSO--CMAES\xspace(\LDO)\xspace}
\newcommand{\MTFldoTBL}{MTF\xspace(\LDO)\xspace}

\newcommand{\StdPSOdftTBL}{StdPSO\xspace(\texttt{DFT})\xspace}
\newcommand{\DEdftTBL}{DE\newline(\texttt{DFT})\xspace}
\newcommand{\CMAESdftTBL}{CMAES\xspace(\texttt{DFT})\xspace}

\newcommand{\fSOCO}{$f_\text{SOCO}$\xspace}
\newcommand{\fCECFIVE}{$f_\text{CEC'05}$\xspace}
\newcommand{\fCECFOURTEEN}{$f_\text{CEC'14}$\xspace}
\newcommand{\fBENCHMARK}{$f_\text{BENCHMARK}$\xspace}

\AtBeginDocument{%
	}

\setcopyright{acmlicensed}
\copyrightyear{2018}
\acmYear{2018}
\acmDOI{XXXXXXX.XXXXXXX}

\begin{document}
	
	\title[\texorpdfstring{\MetafoR}~: A Hybrid Metaheuristics Software Framework for Continuous Optimization Problems]{\texorpdfstring{\MetafoR}~: A Hybrid Metaheuristics Software Framework for Single-Objective Continuous Optimization Problems}

\author{Christian Camacho-Villal\'on}
\affiliation{%
	\institution{Department of Knowledge Technologies, Jožef Stefan Institute}
	\city{Ljubljana}
	\country{Slovenia}
}
\affiliation{%
	\institution{Institut de Recherches Interdisciplinaires et de D{\'e}veloppements en Intelligence Artificielle (IRIDIA), Universit{\'e} Libre de Bruxelles}
	\city{Brussels}
	\country{Belgium}
}
\email{christian.camacho.villalon@ijs.si}
\orcid{0000-0002-0182-3469}
\authornotemark[1]

\author{Marco Dorigo}
\email{mdorigo@ulb.ac.be}
\orcid{0000-0002-3971-0507} 
\author{Thomas Stützle}
\email{thomas.stuetzle@ulb.be}
\orcid{0000-0002-5820-0473}
\affiliation{%
	\institution{Institut de Recherches Interdisciplinaires et de D{\'e}veloppements en Intelligence Artificielle (IRIDIA), Universit{\'e} Libre de Bruxelles}
	\city{Brussels}
	\country{Belgium}
}



\begin{abstract}
	Hybrid \mhs are powerful techniques for solving difficult optimization problems that exploit the strengths of different approaches in a single implementation. 
	For algorithm designers, however, creating hybrid metaheuristic implementations has become increasingly challenging due to the vast number of design options available in the literature and the fact that they often rely on their knowledge and intuition to come up with new algorithm designs.
	In this paper, we propose a modular \msf, called \MetafoR, that can be coupled with an \aact to automatically design hybrid \mhs.
	\MetafoR is specifically designed to hybridize \PSOall, \DEall and \CMAESall, and includes a \ls module that allows their execution to be interleaved with a subordinate \ls.
	We use the configuration tool \irace to automatically generate 17 different \mh implementations and evaluate their performance on a diverse set of continuous optimization problems.
	Our results show that, across all the considered problem classes, automatically generated hybrid implementations are able to outperform configured single-approach implementations, while these latter offer advantages on specific classes of functions. 
	We provide useful insights on the type of hybridization that works best for specific problem classes, the algorithm components that contribute to the performance of the algorithms,
	and the advantages and disadvantages of two well-known instance separation strategies, creating stratified training set using a fix percentage and leave-one-class-out cross-validation.
\end{abstract}




\maketitle

\section{Introduction}
\Cops are ubiquitous in many areas of science and technology, including engineering, finance, education, e-commerce and healthcare, to name a few.
In a $d$-dimensional \cop, the goal is to find a $d$-dimensional vector that minimizes a continuous objective function.
Due to their complexity, many \cops cannot be efficiently approached using traditional analytical methods, such as gradient search or the Hessian matrix computation.
These methods often have limitations when \cops involve multimodal, non-differentiable functions~\cite{Andreasson2020:Book-continuousOptimization}, nonlinear constraints and/or a large number of dimensions~\cite{LueYe1984:Book-linear-nonlinear,Deb2012:book-OptimizationED}, or when the objective function is not explicitly given, as in the case when it is computed using a computer simulation~\cite{AudHar2017:Book-BBO-DFO}.

To tackle difficult \ops of all kinds, researchers have proposed techniques, such as heuristics or metaheuristics~\cite{Glo1986,SorGlo2013,Handbook2019,HooStu05sls-mk}, which are capable of finding "good solutions" in a "reasonable time", but offer no guarantee about the optimality of the solutions generated.
Among the two largest groups of \mhs that can be used to tackle \cops are
\textit{\ec} (\EC)~\cite{FogOweWal1966,Hol75,Rec1973,Schwefel1977,Sch1981}, which include techniques such as \ess (\ESs)~\cite{Rec1971PhD,Schwefel1977}, \gas (\GAs)~\cite{Hol75,Goldberg89},  and \de (\DE)~\cite{StoPri1997:de};
and
\textit{\ssii} (\SSII)~\cite{BonDorThe1999swarm,KenEbeShi01,DorBir2007:sch-si}, which includes techniques such as 
\pso (\PSO)~\cite{KenEbe1995pso,EbeKen1995:pso},
\aco (\ACO)~\cite{SocDor2008:ejor,LegCoe2010:ants-alternativeACO,LiaStuMonDor2014}
and artificial bee colonies~\cite{Kar2005idea,KarBas2007}.
Most \ec and \ssii \mhs are population-based, i.e., they handle multiple solutions at the same time in order to perform a parallel exploration of the search space.

In this paper, we consider some of the most successful \pbmhs 
and investigate how they can be combined to create high-performance hybrid implementations for single-objective continuous optimization problems using an \ad approach.
The \ad approach consist in delegating the task of designing and configuring a metaheuristic implementations to an \aact, e.g., 
ParamILS~\cite{HutHooLeyStu2009jair}, SMAC~\cite{HutHooLey2011lion} or \irace~\cite{LopDubStu2011irace,LopDubPerStuBir2016irace},
rather than to a human algorithm designer.
This approach has already been used to design various state-of-the-art metaheuristics, including 
stochastic \ls~\cite{FraStu2019:cor,PagStu2019:ejor}, 
multi-objective evolutionary algorithms~\cite{BezLopStu2015tec,BezLopStu20:ecj}, 
\DE~\cite{VerCarKon2023:autoDE:gecco}, 
\CMAES~\cite{deNVerWan2021:gecco-CMAESframework,LiaMonStu13:soco},
\ACO~\cite{LopStu2012tec,LiaMolMonStu2014,LopStuDor2017aco}, and 
\PSO~\cite{CamDorStu2022:tec,NebLopGar23:si:mopso}.
However, there are relatively few papers investigating how to automatically combine the strengths of multiple metaheuristics into single implementations \cite{BokWanBac2020:gecco-PSO-DE,LiaStu2013cec}.
In this work we develop a \msf that allows users to instantiate hybrid metaheuristics to solve \cops in a robust and efficient manner.

Our main contribution is the development of a \msf called \MetafoR, which stands for \textsf{\textbf{META}}heuristic \textsf{\textbf{F}}ramework for \textsf{\textbf{O}}ptimization in $\mathbb{R}$eal domains, and incorporates more than one hundred components proposed in the literature for \PSO, \cmaes (\CMAES), \DE and \ls (\LS) algorithms.
\MetafoR allows users not only to replicate many state-of-the-art variants of these metaheuristics, including many of their hybrids, but also to combine their components in new ways using a flexible algorithm template specifically designed for combining \pbmhs with a \ls procedure.
Due to its modular design, \MetafoR can be easily extended with more components allowing users to try novel ideas and assess their usefulness in a systematic manner.
Also, \MetafoR is a parameterized framework: executing it with different parameters results in different implementations, which facilitates its integration with \aacts to explore its design space and find high-performance implementations.

We defined two machine learning-type training scenarios (one in which we included only a fixed percentage of 75\% of the problem instances and another in which we left out an entire problem class) and used \irace to automatically generate 17 implementations of \algInPaper, which we experimentally compared using an heterogeneous benchmark set of 50 continuous functions.
The benchmark set includes unimodal, multimodal and hybrid composition functions, mathematical transformations (rotation, translation and scaling) and varying dimensionalities (from 50 to 1250 dimensions).
We find that automatically generated hybrids involving \CMAES perform better than single-approach implementations across the entire set of 50 functions, whereas configured single-approach implementations of \DE excel in functions with a large number of dimensions.
We study the differences in the algorithms designs and discuss how these differences contribute to the differences in their performance.

The rest of this paper is structured as follows.
\Sect~\ref{sec:Background} provides an overview of \algInPaperLS and briefly discusses the fundamentals of \ad.
\Sect~\ref{sec:Previous works on hybridization} reviews the literature on hybrid implementations combining \algInPaper.
In \Sect~\ref{sec:Metafor}, we introduce \MetafoR and describe the way it has been designed.
In \Sect~\ref{sec:ExperimentalStudy}, 
we experimentally compare default, configured and hybrid implementations of \algInPaper created using \MetafoR. 
Finally, in \Sect~\ref{sec:conclusions}, we conclude the paper and provide ideas for future research.

\section{\Cops, \mhs, and \ad}\label{sec:Background}
Without loss of generality, in a \cop, the goal is to minimize a $d$-dimensional continuous objective function $f : S \subseteq \mathbb{R}^d \rightarrow \mathbb{R}$ by finding a vector $\vec{o}$ in the search space $S$, such that $\forall \:\vec{x} \in S$, $f(\vec{o}) \leq f(\vec{x})$.
The search space $S$ is a subset of $R^d$ in which a solution is represented by a real-valued vector $\vec{x}$, and each component $x_j$ of $\vec{x}$ is constrained by a lower and upper bound: $lb_j \leq x_j \leq ub_j$, for $j = 1,\dots, d$.  
Although there are many different approaches that can be used to deal with \cops \cite{SerOsaMol2019:sec-BioComputation}, in our work we focused on \algInPaperLS, which are among the best performing ones reported in the literature \cite{LiaMolStu2015,VskEftKor19:ieee-cec}.
The main characteristic of \algInPaper is that they are \pbmhs, that is, at each iteration, they keep a population of solutions and perform a parallel exploration-exploitation of the search space.
Differently, \ls algorithms, such as \mtsls \cite{TseChe08:ieee-ec-ci}, make small changes to a single solution based on a step-size parameter that determines how far a new candidate solution will be created.
Since \pbmhs and \ls have complementary characteristics, they are often combined to create algorithms that are better equipped to solve difficult optimization problems \cite{MulBauSba09:ieee-cec,MolLozSan11:soco,GhoDasRoy12:is,LiaStu2013cec}.

\subsection{\Pbmhs}\label{sec:pbmhs}
\subsubsection{\PSOall}\label{sec:PSOall}
In \pso (\PSO) \cite{KenEbe1995pso,EbeKen1995:pso}, particles try to discover the region of the search space where the best quality solutions are located by moving in directions that are estimated based on the best locations that they and their neighboring particles have visited in the past.
The standard version of \PSO (\SPSO) \cite{ShiEbe1998modifiedPSO} 
uses a computational model composed of three main elements: 
(i) a \textit{cognitive component} that allows each particle $i$ in the swarm to memorize the best position it has visited so far, called personal best position $\vec{p}^{i}$;
(ii) a \textit{social component} that allows a particle to know the best position $\vec{l}^{i}$ ever found by any of the particles in its neighborhood; and
(iii) a \textit{velocity update rule} and a \textit{position update rule} that specify how the particles move in the search space and that are defined respectively as follows:
\begin{align}
	\label{eq:psoVelocity}
	\vec{v}^{i}_{t+1} &= \omega \vec{v}^{i}_{t} 
	+ \varphi_1 U^i_{1t} \big(\vec{p}^{i}_t - \vec{x}^{i}_t\big)
	+ \varphi_2 U^i_{2t} \big(\vec{l}^{i}_t - \vec{x}^{i}_t\big) \;\;\;\text{for $i = 1, \dots, n$},\\
	\label{eq:psoPosition}
	\vec{x}^{i}_{t+1} &= \vec{x}^{i}_t + \vec{v}^{i}_{t+1},
\end{align}
where $n$ is the number of particles. 

The position of the particles (vectors $\vec{x}$), which represent candidate solutions to the optimization problem, are updated in every iteration $t$ of the algorithm 
by computing a new velocity vector (\Eq~\ref{eq:psoVelocity}) that is added
to their current positions (\Eq~\ref{eq:psoPosition}).
The computation of a particle's new velocity makes use of two random diagonal matrices, $U^i_{1t}$ and $U^i_{2t}$, to introduce diversity to the particle's movement, and of three real parameters, \PsoParam, that control, respectively, the influence of the previous velocity (also called the particle's inertia), of the cognitive component and of the social component.
The role of vectors $\vec{p}_t$ and $\vec{l}_t$ in the velocity update rule is to combine the knowledge acquired by each particle during the search with the knowledge of the best-informed individual in the neighborhood of the particle.


\subsubsection{\CMAESall}\label{sec:CMAESall}
The evolution strategy (\ES) with covariance matrix adaptation (\CMAES) \cite{Han1997,HanOst2001ec} is a \ES in which the complete covariance matrix of the normal mutation distribution is adapted at execution time.
The main idea in \ESs is to simulate the process by which a population of $\mu$ parents (solutions) undergoes recombination and mutation to generate $\lambda$ offspring (new solutions). Then, a selection operator is applied to choose a subset of these solutions to form the population for the next iteration.
\CMAES is similar to the Quasi-Newton method \cite{NocWri2006} in the sense that it is a second-order estimator that iteratively estimates a positive definite matrix, specifically the covariance matrix. However, unlike the Quasi-Newton method, \CMAES does not use approximate gradients, nor does it assume their existence.
The standard \CMAES implementation \cite{Hansen2006cma,Hansen2016:cma-tutorial} is composed of three main steps.

The first step, \textit{random sampling}, consist in sampling a population of $\lambda$ solutions from a multivariate normal distribution $\mathcal{N}$ with mean $\vec{m} \in \mathbb{R}^d$ and covariance $\mathbf{C} \in \mathbb{R}^{d\times d}$, as shown in the following equation:
\begin{equation}
	\label{eq:CMAES_sampling}
	\vec{x}^{i}_{t+1} = \vec{m}_{t} + \sigma_{t} \times \mathcal{N}(0,\mathbf{C}_{t}) \;\;\;\text{for $i = 1, \dots, \lambda$},
\end{equation}
where $\vec{x}^{i}_{t+1}$ is a vector representing the $i$-th individual in the population, and $\sigma_{t} > 0 $ is the standard deviation that controls the sampling radius or step-size.

In the second step, \textit{weighted intermediate recombination}, the $\lambda$ individuals are ranked in ascending order and the $\mu$ best ones are selected to update the mean of the sampling distribution. The equation describing this process is the following:
\begin{equation}
	\label{eq:CMAES_recombination}
	\vec{m}_{t+1} = \sum_{i=1}^{\mu} w^{i} \vec{x}^{(i|\lambda)}_{t+1},
\end{equation}
where
$w^{1} \geq w^{2} \geq \dots w^{\mu} > 0$,
$\sum_{i=1}^{\mu} w^{i} = 1$,
and $\vec{x}^{(i|\lambda}_{t+1})$ denotes the $i$-th ranked individual of the $\lambda$ solutions sampled using \Eq~\ref{eq:CMAES_sampling}. The weights $w^{i}$ are decreased logarithmically using:
\begin{equation}
	\label{eq:CMAES_weighting}
	w^{i} = \log\bigg(\frac{\lambda-1}{2} + 1\bigg) - \log(i) 
	\;\;\;\text{for $i = 1, \dots, \mu$}.
\end{equation}

The last step is the \textit{covariance matrix adaptation}. The process of adapting the covariance matrix for the next iteration uses a combination of rank-one update (i.e., the mean of the estimated covariance matrices using a single selected step, namely the "evolution path") and the rank-$\mu$ update (i.e., the mean of the estimated covariance matrices from all previous iterations). The update of the covariance matrix is done as follows:
\begin{equation}
	\label{eq:CMAES_cma}
	\mathbf{C}_{t+1} = (1 - c_{\mathrm{cov}}) \mathbf{C}_{t} +
	\frac{c_{\mathrm{cov}}}{\mu_{\mathrm{cov}}}
	\mathrlap{\underbrace{
			\phantom{\;\vec{p}^c_{t+1} \Big({\vec{p}_{t+1}^c}\Big)^T}}_{\text{rank-one update}
	}}
	\;\vec{p}^c_{t+1} \Big({\vec{p}_{t+1}^c}\Big)^T +
	c_{\mathrm{cov}} \bigg( 1 - \frac{1}{\mu_{\mathrm{cov}}}\bigg)
	\mathrlap{\underbrace{
			\phantom{\sum_{i=1}^{\mu} w^{i} \vec{y}^{(i|\lambda)}_{t+1} \Big(\vec{y}^{(i|\lambda)}_{t+1}\Big)^T}}_{\text{rank-$\mu$ update}
	}}
	\sum_{i=1}^{\mu} w^{i} \vec{y}^{(i|\lambda)}_{t+1} \Big(\vec{y}^{(i|\lambda)}_{t+1}\Big)^T
	,
\end{equation}
where 
$c_{\mathrm{cov}} \in [0, 1]$ is the learning rate for the covariance matrix update, 
$\mu_{\mathrm{cov}}$ is used to determine the weighting between the rank-one and rank-$\mu$ update, 
$\vec{p}^c_{t+1}$ 
is the evolution path (i.e., the search path the algorithm takes over a number of iterations and it is expressed as a sum of consecutive steps of $\vec{m}$), and
$\vec{y}^{(i|\lambda)}_{t+1} = (\vec{x}^{(i|\lambda)}_{t+1} - \vec{m}_{t})/\sigma_{t}$.
For a detailed explanation of how to compute the evolution path $\vec{p}^c_{t+1}$ and the new step-size $\sigma_{t+1}$, we refer the reader to \cite{HanArnAug2015evolutionstrategies,Hansen2016:cma-tutorial}.

\subsubsection{\DEall}\label{sec:DEall}
\DEall (\DE) \cite{StoPri1997:de,PriStoLam2005:book} is another evolutionary approach, though it is often regarded as distinct because it incorporates concepts that are similar to those found in \SSII methods.
\DE implements a mutation operator, called \textit{differential mutation}, that is similar to the moves in the Nelder-Mead simplex search method \cite{NelMea1965:tcj-simplex}.
The differential mutation operator consists of selecting three solutions from the population,
computing the difference of the first two solutions and multiplying it by a scaling factor, and then adding the scaled vector difference to the third solution.
More formally, in \DE, the mutation operator is defined as follows:
\begin{equation}
	\label{eq:DE_mutation}
	\vec{m}^{i} = \vec{x}^{a} + \beta \cdot(\vec{x}^{b}-\vec{x}^{c}),
\end{equation}
where $i = 1, \dots, n$ denotes the $i$th individual in a population of $n$ solutions, $\beta$ is the scaling factor and vectors $\vec{x}^{a} \neq \vec{x}^{b} \neq \vec{x}^{c}$ are the three solutions randomly chosen from the population.
Vector $\vec{x}^{a}$, which is referred to as \textit{base vector} and can be selected in many ways, has to be different from the solution in the population for which it is targeted, that is, vector $\vec{x}^{i}$, which is referred to as \textit{target vector}. 
The result of applying \Eq~\ref{eq:DE_mutation} is a vector called \textit{mutant vector}, indicated as $\vec{m}^{i}$.

The creation of the mutant vector is followed by recombination, in which the mutant vector ($\vec{m}^{i}$)
is recombined with target vector ($\vec{x}^{i}$)
to create a \textit{trial vector} $\vec{u}^{i}$.
The equation to apply recombination and obtain the trial vector is the following:
\begin{equation}
	\label{eq:DE_recombination}
	{u}^{i,k} = 
	\begin{cases}
		{m}^{i,k}, & \text{if  $(\mathcal{U}[0,1] \geq p_a) \vee (k = k^{i}_{rand})$}\\
		{x}^{i,k}, & \text{otherwise}
	\end{cases}, \, \text{$\forall k$, $\forall i$},
\end{equation}
where $k = 1, \dots, d$ allows to iterate between the values of the vectors, $\mathcal{U}[0,1]$ is a random number sampled from a uniform distribution, $d$ is the number of dimensions of \cop, $p_a$ is a real parameter in the range $[0,1]$ that controls the fraction of values copied from the mutant vector into the trial vector, and $k^{i}_{rand}$ is a randomly chosen dimension that ensures that the trial vector is not a duplicate of the target vector.  
The newly generated trial vector $\vec{u}^{i}$ only replaces the target vector $\vec{x}^{i}$ in the population if it has better quality, otherwise is discarded. 
Also, as indicated in \Eq~\ref{eq:DE_recombination}, the mutation and recombination operators are iteratively applied for every solution in the population. 

\subsection{Local search strategies for \cops}\label{sec:Local search strategies for cops}

Among the best-known \ls algorithms proposed to tackle \cops are the Nelder-Mead Simplex algorithm \cite{NelMea1965:tcj-simplex}, the Powell’s conjugate directions set algorithm \cite{Powell1964}, and the BOBYQA algorithm \cite{Powell2006} (also by Powell).
In this paper, we consider the more recent \mtsls (\MTSLS) algorithm \cite{TseChe08:ieee-ec-ci}. 
The choice of \MTSLS was motivated by the excellent results obtained by various hybrid algorithms for continuous optimization that integrate this local search \cite{LiaStu2013cec,LaTMuePen11:soco}.

The \MTSLS algorithm starts by generating a candidate solution $\vec{s}=(s_1,s_2, \ldots, s_d)$ uniformly at random inside the search range. 
The initial step size $ss$ is set to $ss= 0.5\times(ub_j-lb_j)$, where 0.5 is a default value \cite{TseChe08:ieee-ec-ci} and $ub$ and $lb$ are, respectively, the upper and lower bounds of dimension $j$.
\MTSLS visits the dimensions of the problem in a fixed order, searching one dimension at a time. 
For $j= 1\dots d$, \MTSLS proceeds as follows. First, the value $s_j$ is modified as $s'_j = s_j-ss$ and the resulting solution $s'$ is evaluated. 
If $f(s') < f(s)$, then $s_j= s'_j$ and the search continues in dimension $j+1$; otherwise, $s''_j = s_j+0.5\times ss$ and the candidate solution $s''$ is evaluated. 
If $f(s'') < f(s)$, then $s_j = s''_j$ and the search continues in dimension $j+1$. 
However, if both $s'_j$ and $s''_j$ do not improve over $s_j$, then $s_j$ remains unchanged and the same process is applied to dimension $j+1$. 
If no improvement is found in any of the dimensions during one iteration of \MTSLS, the next iteration uses only half the value of the step size, i.e., $ss_{t+1} = ss_t/2$. In our implementation of \MTSLS, the maximum number of iterations in one execution of \MTSLS is determined by the parameter \MTSLSiterations.

\subsection{Automatic design of \mh implementations}

\Ad~\cite{CamStuDor2023:IC:disNewMH} has been proposed as an alternative to manual design, with the goal of reducing the burden on algorithm designers who have to deal with the problem of creating metaheuristic implementations that meet certain performance requirements. 
In \ad ~\cite{StuLop2019hb,CamStuDor2023:IC:disNewMH}, generating effective algorithm designs is treated as an optimization problem, focused on discovering high-performing combinations of algorithm components and parameter settings.
By using an \ad approach, algorithm designers do not have to deal themselves with the tasks of composing many different metaheuristic designs, assessing their performance and selecting the best ones, since these tasks are delegated to an \aact (\AACT). 
The \AACT can systematically explore the design space of a metaheuristic until it finds one that meets the user's needs or until a maximum computational budget is exhausted.


The \AACT used in this paper is \irace~\cite{LopDubStu2011irace,LopDubPerStuBir2016irace}, which implements an \textit{iterated racing} approach.
The iterated racing approach~\cite{MarMoo1997air,BirYuaBal2010:emaoa} is based on the idea of performing sequential statistical testing 
in order to create a sampling model that can be refined by iteratively ``racing'' candidate configurations and discarding those that perform poorly.
The way in which \irace works can be summarized as follows. 
First, it samples candidate configurations from the parameter space and evaluates these configurations on a set of instances using a racing procedure, where each configuration is tested on one instance at a time. 
Then, \irace eliminates the statistically inferior configurations based on Friedman's non-parametric two-way analysis of variance by ranks.
Throughout the configuration process, which proceeds sequentially within a given computational budget, \irace adjusts the sampling distribution to favor the best configurations identified so far.
This process is repeated iteratively until a computational budget is exhausted and \irace returns the configuration that performed best on the training instances.

\section{Previous works on hybridization and \ad}
\label{sec:Previous works on hybridization}
The goal of creating hybrid \mhs is to exploit the strengths that different optimization techniques can offer to solve optimization problems
~\cite{GroAbr07:hea,Tal2013hm,BluRai2016:book,StuLop2019hb,CalArmMas2017:Learnheuristics}.
In the following, we discuss the main ideas proposed in hybrids of \algInPaper, as well as some modular \msfs that have been proposed to study these metaheuristics and their hybrids in an \ad context.
Note, however, that this review of the literature is by no means exhaustive, as we focus on papers proposing ideas that are particularly amenable to be implemented in \MetafoR, such as single-objective hybrids and not overly complicated hybrids.

\subsection{\PSO and \DE hybrids}
Hendtlass proposed an algorithm called swarm differential evolution algorithm (SDEA) \cite{Hendtlass01:iea-aie}. SDEA works mainly as the standard \PSO algorithm, but intermittently applies the \DE mutation operator to the particles' current solutions to avoid local minima. 
Zhang and Xie proposed an algorithm called DEPSO, in which the \stanPSO is applied during even iterations of the algorithm, and the \DE mutation operator is applied to the particles' personal best during odd iterations~\cite{ZhaXie03:ieee-smc}.
Das \etal proposed a hybrid algorithm in which the personal component of the velocity update rule of the \stanPSO is modified based on the mutation operator of \DE~\cite{DasKonCha05:gecco}.
In \cite{OmrEng07:ieee-sis}, Omran \etal introduced a modified version of Hendtlass’ SDEA that uses a probabilistic approach based on the "barebones" \PSO. 
In their algorithm, with probability $p_r$, particles update their position using the barebones \PSO position update rule and add a vector difference, and with probability $1-p_r$, they update their position to the personal best of a randomly selected particle.
Pant \etal introduced a two-phase hybrid version of \DE that uses \PSO as perturbation mechanism~\cite{PanThaGro08:ieee-dim}. Pant \etal's algorithm follows the usual \DE procedure up to the point where the trial vector is generated; if the trial vector is rejected, the algorithm applies the \PSO velocity and position update rules to generate a new solution.
Epitropakis \etal proposed to evolve the social and cognitive components of a swarm by applying the three usual \DE operators (i.e., mutation, recombination and selection) to the personal best positions particles (i.e., vectors $\vec{p}^{i}$ in \Eq~\ref{eq:psoVelocity})~\cite{EpiPlaVra12:is}. 


\subsection{\PSO and \CMAES hybrids}
One of the earliest \PSO--\ESs hybrids is the "particle swarm guided evolution strategy" proposed by Hsieh \etal~\cite{HsiCheChe07:gecco}.
Although this approach is not properly based on \CMAES, the authors introduced the idea that \ESs can be used to focus on exploiting good quality solutions, while \PSO can be used to focus on performing effective search space exploration.
%
Müller \etal proposed to run multiple \CMAES instances in parallel considering each instance as an individual particle in \PSO~\cite{MulBauSba09:ieee-cec}.
The proposed algorithm is divided into two phases: a \CMAES phase, which follows the usual \CMAES algorithm, and a \PSO phase, where the best solutions found by each \CMAES instance applies the \stanPSO velocity and position update rules.
The equation to adapt the covariance matrix of \CMAES was modified to combine the \textit{local} information gathered by a \CMAES with the \textit{global} information of \PSO.
Müller \etal also added a bias to the mean of the sampling distribution of \CMAES in the following cases: (i) when the instance has already converged to a local minimum located far from the the global best solution, and (ii) when the instance is different from the one that produced the global best solution and the step-size has fallen below a certain threshold.
%
In \cite{XuLuoLin19:ieee-cec}, Peilan \etal introduced a hybrid, three-phase algorithm that uses multiple populations and two different versions of \PSO, the \stanPSO and a \PSO with time windows (PSOtw).
In PSOtw, particles can only access their personal best if it is within a certain time window $tw$ given in number of iterations.
The first phase of Peilan \etal's algorithm consists in applying PSOtw by each population for a number of iterations; in the second phase, the best and second best solutions in each population are selected to create a temporary population $P_t$; in the third phase and last, the \stanPSO and \CMAES algorithms are applied one after the other to $P_t$ for a number of function evaluations and the best solutions in $P_t$ are selected for the next iteration.
%

\subsection{\CMAES and \DE hybrids}
Kämpf and Robinson proposed to execute \CMAES and \DE in sequential order, with \CMAES followed by \DE~\cite{KamJerRob09:asoc}.
The elite solutions found by \CMAES are input to \DE, but since \DE uses a larger population size compared to \CMAES, the \DE population is completed with randomly created solutions.
Ghosh \etal introduced a hybrid algorithm that incorporates the operators of \DE into the structure of \CMAES~\cite{GhoDasRoy12:is}.
The standard mechanism of \CMAES is used in each iteration to sample new solutions from a multivariate normal distributions, after which the population is handled as in \DE.
At each iteration, Ghosh \etal' algorithm performs the following steps: (i) update the mean of the sampling distribution; (ii) adapt the covariance matrix;
(iii) create a population of mutants using components from the eigen decomposition of the covariance matrix; and (iv) apply the crossover and selection operators of the usual \DE algorithm.
In \cite{GuoYan14:tevc}, Guo and Yang proposed a crossover operator that allows individuals to exchange information between the target vector and the mutant vector using the eigenvector basis instead of the natural basis, thus rotating the coordinate system and making the algorithm rotation invariant.
In \cite{HeZho18:asoc}, He and Zhou presented a hybrid algorithm based on a new mutation operator and a simplified step-size control rule.
The mutation operator used the information of previously rejected solutions and a Gaussian distribution to sample probabilistically "better" solutions.
The standard covariance matrix update of \CMAES was modified to include only the rank-$\mu$ update, and not the rank-one update.
He and Zhou also added a local search that takes place only if there are still at least 10\% of the total number of function evaluations available by the time the algorithm has converged. 
Chen and Liu proposed a hybrid algorithm in which \DE is used to break the stagnation state of \CMAES~\cite{CheLiu22:nature-scirep}.
The algorithm measures the average fitness improvement of the solutions after each iteration of \CMAES and, when the average falls below a predefined threshold, \DE is executed to regenerate the population. 
To make sure that the new solutions obtained by \DE are redistributed in the search space, only the offspring are selected for survival.


\subsection{Modular \msfs for \algInPaper}
\Msfs (\MSFs) that can be used to automatically generate implementations of \algInPaper have already been proposed in the literature: \PSOX~\cite{CamDorStu2022:tec}, \aCMAES~\cite{deNVerWan2021:gecco-CMAESframework}, \aDE~\cite{VerCarKon2023:autoDE:gecco} and \modPSODE~\cite{BokWanBac2020:gecco-PSO-DE}. 
Both \PSOX, \aCMAES and \aDE are parameterized \MSFs that allow users to create many different implementations by simply changing the parameters used to execute the framework. 
This aspect makes them particularly suitable for an \ad context, as they can be easily coupled with an \AACT tool such as \irace.
On the other hand, the main drawback of these \MSFs is that they only contain components of a specific metaheuristic, \PSO, \CMAES and \DE, respectively.

The modular \modPSODE framework is, to the best of our knowledge, the only \MSF specifically developed for creating \PSO and \DE hybrids in an automatic design context~\cite{BokWanBac2020:gecco-PSO-DE}.
Unfortunately, it has two important downsides: (i) a limited number of components of \PSO and \DE compared to \PSOX and \aDE; and (ii) lack of flexibility to combine components at a fine-grained level.
The \modPSODE framework is based on the idea of simultaneously generating two populations, one using \PSO components and the other using \DE components, and combining them into a single population that is then reduced to a predefined population size using a selection operator.
This type of hybridization is not particularly useful to explore the interaction of \PSO components in \DE and vice-versa.

\section{The \texorpdfstring{\MetafoR}~software framework}\label{sec:Metafor}
\MetafoR is composed of three main modules: 
a \PSO module (\PSOmod), which includes algorithm components of \PSO to update solutions by adding a velocity vector; 
a \DE module (\DEmod), which includes algorithm components of \DE to update solutions using differential mutation, recombination and selection;
and a \CMAES module (\CMAESmod), which includes algorithm components of \CMAES to update solutions by applying random sampling, recombination and covariance matrix adaptation.
\MetafoR also includes a \ls module (\LSmod), which allows the execution of \MetafoRModules to be interleaved with a local search (\MTSLS or \CMAES)
for a number of function evaluations (FEs) and then resume with the main algorithm execution. 

\MetafoRModules can be used standalone or in combination with other modules by selecting specific components from them.
As an example, with \MetafoR it is possible to create a hybrid \DE--\PSO implementation that combines the differential mutation of \DE with the velocity update rule of \PSO, and runs \CMAES as local search for a number of FEs.
The way in which \MetafoRModules interact with each other is controlled by the algorithm component \Execution.
\Execution operates at a high level in \MetafoR and contains the necessary options for creating different types of hybrid implementations (e.g., component-based, multiple phases, etc.). 
We will now provide a detailed description of the modules that comprise \MetafoR. We use a \textsf{sans-serif} font to indicate both the algorithm components implemented in the framework and their available options.

\subsection{\PSOmod}\label{sec:The PSOX module}
\PSOmod includes all the algorithm components and implementation options as the original \PSOX framework~\cite{CamDorStu2022:tec}.\footnote{For specific details about the implementation of the algorithm components included in the \PSOX framework, we refer the reader to the original paper~\cite{CamDorStu2022:tec} and supplementary material~\cite{CamDorStu2021:posx-supp}.}
The two top-level algorithm components of \PSOmod are \Pop and \Topo.
\Pop handles the number of solutions (particles) in the implementation and has three options available: \PopConst, \PopIncre and \PopTV. 
When \PopIncre or \PopTV are used, the user has to specify the way new solutions added to the population are initialized, which is done via the algorithm component \Init using option \InitHorizontal or \InitRandom.
The \Topo algorithm component determines the way particles connect to each other and it can be implemented in seven different ways: \TopFC, \TopHie, \TopNeum, \TopRing, \TopRandom, \TopTVFI and \TopWheel; each option providing different connectivity and information transfer speed to the swarm.

In \PSOmod, the components involved in the computation of a particle's new velocity vector ($\vec{v}^{\,i}_{t+1}$) are combined using a generalized velocity update rule, which is defined as follows:
\begin{equation} 
\label{eq:GeneralizedVUR}
\vec{v}^{\,i}_{t+1} = \omega_1\,\vec{v}^{\,i}_t + \omega_2\,\DNPP(i,t) + \omega_3\,\PertRnd(i,t),
\end{equation}
where $\omega_1$, $\omega_2$ and $\omega_3$ are three real parameters in the range $[0,1]$ used to control the influence of each term in the equation, $\vec{v}^{\,i}_t$ is the velocity of the particle $i$ at iteration $t$, \DNPP is an algorithm component that determines the type of mapping from a particle's current position to the next one, and \PertRnd is an optional algorithm component to add a perturbation vector.
\PSOmod provides nine options for computing the value of $\omega_1$, most commonly known as the inertia weight (\IWcomp). Three of these options update their value at regular intervals of the algorithm execution based on predefined schedules (\OmegaLinDec, \OmegaLinInc and \OmegaRnd), whereas the other six options update their value adaptively, based on information gathered from the option process (\OmegaSelfReg, \OmegaAdapVel, \OmegaDoubExp, \OmegaRnkBsd, \OmegaSuccBsd and \OmegaConvBsd).
The parameters $\omega_{2}$ and $\omega_{3}$, which regulate the influence of \DNPP and \PertRnd, respectively, can either be set equal to $\omega_{1}$ or determined using the \OmegaRnd and \OmegaCons options.

The options for implementing the \DNPP algorithm component are \DNPPRect, \DNPPSphe, \OperatorS, \OperatorD, \OperatorN and \OperatorCG.
The \DNPPRect option is the most commonly used in \PSO variants, including the \textit{standard} \PSO \cite{KenEbeShi01}, the \textit{fully-informed} \PSO \cite{MenKenNev2004fullyInformed_pso} and the \textit{locally convergent rotationally invariant} \PSO \cite{BonMic2014:swarm}, and it was used as the basis for the \DNPPSphe option, proposed for the \textit{standard} \PSO--\textit{2011} \cite{Clerc2011:pso,ZamCleRoj2013:cec2013}. 
In the case of \OperatorS, \OperatorD, \OperatorN and \OperatorCG, they were proposed for the so-called simple dynamic \PSO algorithms, whose main characteristic is that they do not use the previous velocity vector or the random diagonal matrices.
The algorithm component \VectBasis, which is not part of the original \PSOX framework, works in combination with \DNPP.
\VectBasis uses the eigenvector information of the population covariance matrix to rotate the coordinate system, with the goal of making the implementation invariant to rotated search spaces.
\VectBasis can be implemented using the option \VBnatural, where no changes are made to the coordinate system, or the option \VBeigen, where the vectors involved in the computation of the \DNPP are rotated in the direction of the eigenvector basis.

When \DNPP is implemented as \DNPPRect or \DNPPSphe, the user has to indicate the options for the \textsf{Random Matrices} (\Mtx), \Moi (\textsf{MoI}) and \textsf{Acceleration Coefficients} (\ACcomp) components.
\Mtx refers to the different ways in which the random matrices can be constructed, such as \MtxDiagonal, \MtxLinear, \MtxExponential, \MtxEuclideanOne (rotation in plane), \MtxEuclideanAll (rotation in all possible planes) and \MtxIncreasingGroupBased.
\Moi specifies which neighbor solutions influence a particle movement. The available options for this component are \MoiBoN (only the best neighbor), \MoiFI (all of its the neighbors and all with same weight) and \MoiRFI  (all of its neighbors, but the higher its quality, the more weight it has).
\ACcomp manages the computation of the parameters $\varphi_1$ and $\varphi_2$ (see \Eq~\ref{eq:psoVelocity}) and offers four options for this purpose: \PhiConstant, \PhiRandom, \PhiTV and \PhiExtra.

\PSOmod has five optional algorithm components:
\PertInf, \PertRnd, \VelClp, \StagDet and \IgnPbest.
Both \PertInf and \PertRnd apply a perturbation to the components of the generalized velocity update rule (\Eq~\ref{eq:GeneralizedVUR}) --- \PertInf does it to the vectors involved in the computation of the selected \DNPP option, whereas \PertRnd generates a random vector that is added directly to a particle's new velocity.
The options available for implementing \PertInf are \PertGau, \PertLev and \PertUni, while the options for implementing \PertRnd are \PertRect and \PertNoi.
When \PertInf and \PertRnd are used, the user has to select the strategy to compute the magnitude of the perturbation (variable $pm$), which is done via the algorithm component \textsf{Perturbation Magnitude} (\PMcomp) using options \MagCons, \MagEucli, \MagOFd and \MagSucc.

The algorithm components \VelClp and \StagDet limit the size of the velocity vectors of the particles.
\VelClp halves the magnitude of the velocity vector when its value exceeds the bounds of the search space, while \StagDet re-initializes the velocity vector when $||\vec{v}^{\,i}_{t}|| + ||\vec{l}^{i}_t - \vec{x}^{\,i}_{t}|| \leq 10^{-3}$, where $||\cdot||$ denotes the L$^2$ norm of a vector.
The component \IgnPbest allows particles to ignore their personal best vectors and to use instead their current positions in the computation of the velocity and position update rules.
Despite \IgnPbest is not a common design choice in most \PSO implementations, 
the motivation to include it in \MetafoR is the large number of 
"novel" metaphor-based \mh 
proposed in recent years (e.g., \GwoAll, \MfAll and \WoaAll) that seem to be based on \PSO~\cite{AraCamCam-etal2022:openletter:si,CamDorStu2022:exposing:itor}.
Our goal is that, with \MetafoR, users can also replicate some of these "novel" \mh and evaluate their performance in a systematic manner.

\subsection{\CMAESmod}
\label{sec:The CMAES module}
\CMAESmod was developed based on the implementation of \CMAES 
publicly available from its creator's website.\footnote{http://www.cmap.polytechnique.fr/\~nikolaus.hansen/cmaes\_inmatlab.html\#code} 
\CMAESmod allows to implement the algorithm components proposed in some of the best-known variants of \CMAES, including the \textit{standard}-\CMAES \textit{with intermediate recombination}~\cite{HanMulKou03:ec}, the \textit{separable}-\CMAES \cite{RosHan2008:sepCMAES} and the \textit{restart}-\CMAES \textit{with increasing population size} \cite{AugHan2005cec}.
\CMAESmod is composed of four main algorithm components:
\CMAESPop, which handles the number of solutions in the implementation;
\CMAESMtx, which specifies the way the covariance matrix is adapted;
\CMAESRes, which restarts the algorithm based on criteria related to the range of improvement of the solutions found by the population;
and \CMAESRW, which specifies the weighing mechanism used by recombination.

In \CMAESmod, the size of the population is controlled using the \CMAESPop algorithm component. The two options available for implementing this component are \CMAESPopConst, where the size of the population remains the same throughout the algorithm execution, and \CMAESPopInc, where it increases according to a multiplication factor.
The initial population size (denoted by $\lambda_{t=0}$) and the number of parents selected for recombination (denoted by $\mu_t$) are computed using $\lambda_{t=0} = 4 + \lfloor \CMAESpara \ln(d)\rfloor$ and $\mu_t = \lambda_t / \CMAESparb$, where \CMAESpara and \CMAESparb are two real parameters in the range $[1, 10]$ and $[1, 5]$, respectively.
To sample the initial population (\Eq~\ref{eq:CMAES_sampling}), the initial step size is computed as $\sigma_{t=0} = \CMAESparc [lb_j, ub_j]$, where \CMAESparc is a real parameter in the range $(0,1)$ and $lb_j$ and $ub_j$ are the lower and upper bounds of dimension $j$.
When \CMAESPopInc is used, the size of the population is updated according to the following equation:
\begin{equation} 
\label{eq:CMAES_pop_increment}
\lambda_{t+1} = 
\begin{cases}
	\CMAESpard * \lambda_t, & \text{if \CMAESRes = \textsf{true}}\\
	\lambda_t, & \text{otherwise}
\end{cases},
\end{equation}
where \CMAESpard is a real parameter in the range $[1,4]$ that controls the velocity at which the population grows, and  \CMAESRes (described below) determines the moment at which the increments take place.

The \CMAESMtx algorithm component allows to adapt the covariance matrix in three different ways, \CMAESMtxCov, \CMAESMtxDia and \CMAESMtxCovDia. 
The \CMAESMtxCov option is the one described in \Eq~\ref{eq:CMAES_cma}, where the adaptation of the covariance matrix is done using a combination of the rank-one update 
and the rank-$\mu$ update.
\CMAESMtxDia is a low complexity alternative for \CMAESMtx intended for separable functions, where the covariances are assumed to be zeros; therefore, instead of the full covariance matrix, \CMAESMtxDia uses a diagonal matrix similar to matrices $U_1$ and $U_2$ in \PSO (see \Eq~\ref{eq:psoVelocity}). 
The \CMAESMtxCovDia option allows to start the implementation using \CMAESMtxCov, and then, after $2+100 \times \frac{d}{\sqrt{\lambda}}$ FEs to switch to \CMAESMtxDia.

When used in the implementation, the algorithm component \CMAESRes performs two actions: first, it re-initializes the mean of the random sampling with a randomly generated solution; second, if \CMAESPopInc is used, it increases the population size according to \Eq~\ref{eq:CMAES_pop_increment}. 
\CMAESRes is a boolean algorithm component that becomes \textsf{true} if either the range of improvement of (i) all function values of the most recent generation, (ii) the best solution found in the last $[10 + \texttt{round}(30d/\lambda)]$ generations, or (iii) the standard deviation of the normal distribution falls below a certain threshold.
\CMAESmod keeps the information related to (i), (ii) and (iii) in variables \texttt{stopTolFun}, \texttt{stopTolFunHist} and \texttt{stopTolX}, respectively, and 
iteratively verifies the conditions that make \CMAESRes $= \textsf{true}$, namely:
$\texttt{stopTolFun} \leq 10^{\CMAESpare}$,
$\texttt{stopTolFunHist} \leq 10^{\CMAESparf}$, or
$\texttt{stopTolX} \leq 10^{\CMAESparg}$,
where \CMAESpare, \CMAESparf and \CMAESparg are real parameters in the range $[-20, -6]$.

The last algorithm component in \CMAESmod is \CMAESRW. 
This component is used to specify the weighting scheme used during recombination and it can be implemented using options \CMAESRWlog, \CMAESRWlindec and \CMAESRWeq.
The \CMAESRWlog option is the one shown in \Eq~\ref{eq:CMAES_weighting}, where weights decrease in a logarithmic fashion.
Differently, \CMAESRWlindec is defined as $w^{i} = (\lambda - i-1)/\sum_{k=0}^{\mu-1}\big(\lambda-k\big)$ for $i=1\dots\mu$, and \CMAESRWeq is defined as $w^{i} = 1/\mu $ for all $i$.


\subsection{\DEmod}\label{sec:The De module}
\DEmod was developed considering some of the most popular variants of \DE published in the literature~\cite{DasSug2011:tec,VerCarKon2023:autoDE:gecco}.
\DE implementations are typically referred to using a mnemonic of the form \DE/\textit{term\_2}/\textit{term\_3}/\textit{term\_4}, where \textit{term\_2} indicates the way in which the base vector is chosen, \textit{term\_3} indicates the number of vector differences added to the base vector, and \textit{term\_4} indicates the number of values donated by the mutant vector (see \Sect~\ref{sec:DEall} for details). 
Some popular variants of \DE indicated using their mnemonics are \textit{DE/rand/1/bin} (classic \DE), 
\textit{DE/target-to-best/1/bin}, and \textit{DE/rand/1/exp}.
The implementations created with \DEmod can be referred to in a similar way using the following extended mnemonic:
\begin{equation} 
\label{eq:GeneralizedDEmnemonic}
\textrm{\DE/\DEBaseVect/\DEVectDifferences/\DERcb/\DEVect/\VectBasis}
\end{equation}
where \DEBaseVect, \DEVectDifferences and \DERcb refer to the same concepts as \textit{term\_2}, \textit{term\_3} and \textit{term\_4} in the original mnemonic, \DEVect indicates the solution vectors 
used to compute the vector differences, and \VectBasis indicates the vector basis used in the implementation.
In \DEmod, the population is handled using the algorithm component \Pop (already described in \Sect~\ref{sec:The PSOX module}) with the exact same options available (i.e., \PopConst, \PopIncre and \PopTV).

The algorithm component \DEBaseVect determines the way the base vector (vector $\vec{x}^{a}$ in \Eq~\ref{eq:DE_mutation}) is chosen.
\DEBaseVect provides five options for its implementation: \DEBVrand, \DEBVbest, \DEBVttbest, \DEBVdirRand and \DEBVdirBest. 
The first three options are the most commonly used in popular \DE variants. In \DEBVrand the base vector is chosen at random; in \DEBVbest the base vector is the best-so-far solution; and in \DEBVttbest the base vector lies between the target vector and the best-so-far solution. 
The implementation options \DEBVdirRand and \DEBVdirBest seek to incorporate information from the objective function into the creation of the mutant vector.
When options \DEBVdirRand and \DEBVdirBest are used, the creation of the mutant vector is done as follows:
\begin{equation}
\label{eq:DE_mutation_directed}
\vec{m}^{i} = \vec{x}^{a} + \frac{\beta}{2} \cdot(\vec{x}^{a}-\vec{x}^{b}-\vec{x}^{c}),
\end{equation}
where $\vec{x}^{a}$, $\vec{x}^{b}$ and $\vec{x}^{c}$ are chosen at random, with $f(\vec{x}^{a}) \leq \{ f(\vec{x}^{b}),f(\vec{x}^{c})\}$.
The only difference between \DEBVdirBest and \DEBVdirRand is that, in \DEBVdirBest, $\vec{x}^{a}$ is the best solution found so far.
The numerical parameter $\beta \in (0,1]$ scales the vector differences added to the base vector and is divided by the number of vector differences when there are more than one, as shown in \Eq~\ref{eq:DE_mutation_directed}.

The number of vector differences added to a base vector is computed by the algorithm component \DEVectDifferences.
Although most implementations of \DE only add one or two vector differences, \DEVectDifferences allows to add up to a quarter of the population size as vector differences. 
In practice, \DEVectDifferences takes into account the option implemented for \DEBaseVect and assigns to each individual in the population the set of solutions that will be used to create its mutant vector.
Also, when options \PopIncre or \PopTV are used, \DEVectDifferences adjusts the number of vector differences according to the current population size.

The \DERcb component specifies the type of recombination (\DERcbcomp) used in the implementation.
The options available for implementing \DERcb are \DERcbBinomial and \DERcbExponential.
The option \DERcbBinomial, a.k.a. uniform random, is the one shown in \Eq~\ref{eq:DE_recombination} in \Sect~\ref{sec:DEall}, where the number of values donated by the mutant vector follows a binomial distribution.
Differently, in the \DERcbExponential (or two-point modulo), the number of values donated by the mutant vector follows an exponentially distributed random variable.
The fraction of values copied from the mutant vector into the trial vector during recombination is controlled by the 
parameter $p_a$ (see \Eq~\ref{eq:DE_recombination}), whose value is a real number in the range $[0, 1]$. 

In addition to the typical algorithm components used in \DE implementations, \DEmod includes components \DEVect and \VectBasis.
The algorithm component \DEVect allows to use different types of solutions when creating a mutant vector.
The options available to implement \DEVect are \DEVectPos, which uses the current solutions (most \DE implementations used this option); \DEVectPbest, where individuals keep track of their the personal best solutions and use them to compute the mutant vector (an idea borrowed from \PSO); and \DEVectMix, which uses a combination of the current and personal best solutions.
On the other hand, the algorithm component \VectBasis allows to rotate the coordinate system of the vectors involved in \DERcb.
The two options available for the \VectBasis component are \VBnatural, which uses the natural basis (i.e., no rotation is performed), and \VBeigen, which computes the eigenvector basis of the population and uses it to perform the rotation.
When \VBeigen is used, the mutant and target vectors are rotated in the direction of the eigenvector before applying \DERcb, and then, after \DERcb takes place, they are rotated back to the natural basis to apply selection.

\DEmod has two algorithm components specifically designed to facilitate the creation of component-based \PSO and \DE hybrids.
The first one, \DERecompVel specifies the type of update performed to the velocity vector ($\vec{v}^{\,i}_t$ ) when \DE, which has precedence over \PSO in component-based \PSO and \DE hybrids, finds a better solution.
In this case, the velocity vector of the particle does not correspond anymore to its position, which can negatively impact the application of \PSO.
To fix this issue, \DERecompVel recomputes the velocity vector of the particle according to one of the following strategies: \DERVgoBack (proposed by Boks \etal \cite{BokWanBac2020:gecco-PSO-DE}), where $\vec{v}^{\,i}_t$ is recomputed as the difference between the new solution found by the application of \DE ($\vec{x}^{\,i,\text{DE-new}}_{t}$) and the previous current solution ($\vec{x}^{\,i}_{t}$), that is, $\vec{v}^{\,i}_t = \vec{x}^{\,i,\text{DE-new}}_{t} - \vec{x}^{\,i}_{t}$; \DERVrandom, where $\vec{v}^{\,i}_t$ is regenerated at random; \DERVposition, where $\vec{v}^{\,i}_t = \vec{x}^{\,i}_{t}$; and \DERVnone, where no update is done to $\vec{v}^{\,i}_t$.
The second algorithm component, called \PSOonFail, is the one proposed in the \DE-\PSO hybrid by Pant \etal \cite{PanThaGro08:ieee-dim}.
As its name suggests, \PSOonFail is used to specify that \PSO is applied only when \DE failed to produce a new better solution.
Since the number of FEs used per iteration by \textsf{components-based} hybrids of \DE and \PSO is twice the size of the population, \PSOonFail can lower this number by preventing the application of \PSO after $\vec{x}^{\,i}_{t}$ has already been improved by \DE.


\subsection{\LSmod}\label{sec:The Ls module}
\LSmod allows to interleave the execution of \MetafoRModules with a subordinate local search.
The execution of the \ls algorithm is controlled by two parameters, \LSbudget and \LSdivide.
\LSbudget, which is a real number in the range $(0,1]$, indicates what percentage of the total number of FEs is allocated to \ls, whereas \LSdivide, which is an integer in the range $[1,100]$, specifies the number of independent runs of the \ls in the implementation.
For example, setting $\LSbudget =0.25$ and $\LSdivide =10$ produces 10 independent runs of the \ls, where each run has a maximum budget of $(0.25 \cdot \mathrm{total\,number\,of\,function\,evaluations})/10$.
If the \ls algorithm reaches the value of \LSdivide without finishing its budget, \LSmod adds extra runs, one at a time per iteration, until there is no more budget available.

The two \ls algorithms that can be implemented using \LSmod are \CMAES and \MTSLS.
Although the implementation of \CMAES as \ls is also done via \CMAESmod with the exact same available options and parameters (see \Sect~\ref{sec:The CMAES module}), each \CMAES instance is treated independently in \MetafoR.
In other words, in a single \MetafoR implementation, it is possible to have two instances of \CMAES, one as (or part of) the main optimization algorithm, and another as \ls, each with its own algorithm components and parameter values.
Whatever the type of implementation, \LSmod uses as input for the \ls algorithm the best solution found so far ($\vec{x}^\text{best}$), which in the case of \CMAES is used as mean for the random sampling, and in the case of \MTSLS as starting solution.

The implementation of \MTSLS works as described in \Sect~\ref{sec:Local search strategies for cops} and has three parameters associated: \MTSLSinitss, which is the initial step size, \MTSLSiterations, which is an integer in the range $[1,3]$ that  determines the number of iterations per run of \MTSLS, and \MTSLSbias, which is a real number in the range $[-1,1]$ that controls how close a randomly generated solution is to $\vec{x}^\text{best}$.
In our implementation of \MTSLS, solutions that exceed the boundaries of the search space are penalized by adding the sum of the squares of their offsets to the function evaluation. This ensures that the farther a solution is from the search space, the greater the penalty it incurs.
At the end of each iteration, \MTSLS tests for convergence using $\vert f(\vec{s}) - f(\pvec{s}') \vert \leq 10^{-20}$, where $\vec{s}$ is the initial solution and $\pvec{s}'$ the improved solution.
If the convergence test is positive, the value of \MTSLSinitss is reinitialized to a random number sampled from $\mathcal{U}[0.3,0.6]$.
After completing one iteration, if no improvement was made or if the improvement was smaller than $10^{-20}$, \MTSLS generates a random starting solution for the next iteration as follows: 
$s_j = \mathcal{U}[lb_j,ub_j] + \big( (1-\MTSLSbias)\cdot\mathcal{U}[0,1] + \MTSLSbias \big) \cdot (x_j^{\text{best}}-\mathcal{U}[lb_j,ub_j])$ for $j=1\dots d$.

\subsection{The \Execution algorithm component}\label{sec:The Execution module}
Based on the literature review in \Sect~\ref{sec:Previous works on hybridization}, we developed an algorithm component called \Execution that allows to replicate the way in which hybrids of \algInPaper are created.
The three options available to implement \Execution are \ExecutionCB, \ExecutionPB and \ExecutionMP.
The \ExecutionCB option (available for each module when used standalone and for \PSOmod and \DEmod when used together) allows users to compose an algorithm by selecting individual algorithm components from different modules.

Differently, in the \ExecutionPB option (available only for \PSOmod and \DEmod), each module is applied probabilistically based on parameter $pr \in [0, 1]$, so that the probability of updating a solution using the first module is $pr$ and the probability of updating it using the second module is $1-pr$.
\ExecutionPB provides three options to sample random numbers to compare with the value of $pr$, \ExecutionPBuni, which uses a uniform distribution $\mathcal{U}(0,1)$, \ExecutionPBnor, which uses a normal distribution $\mathcal{N}(0,\,\ExecutionParStd)$, and \ExecutionPBlev, which uses a Lévy distribution $\mathcal{L}(0,\, \ExecutionParStd,\, \gamma_t)$.
In both \ExecutionPBnor and \ExecutionPBlev, the mean of the distribution is 0 and the standard deviation is controlled by the parameter \ExecutionParStd.
In the case of \ExecutionPBlev, $\gamma_t$ is a real parameter that controls the sharpness of the distribution and its value is sampled from a discrete uniform distribution $\mathcal{U}\lbrace10, 20\rbrace$, so that the probability of generating a random value in the tail of the distribution varies iteration by iteration---see \cite{RicBla2006levyPSO} for details.

\ExecutionMP (available for all four modules) is similar to \ExecutionPB, but instead of a probability, the user gives a computational budget $xb$ to each module that determines the number of consecutive FEs the module can use. \ExecutionMP divides the total number of FEs among the different modules based on their assigned budget $xb$ and applies each module until $xb$ is over, one module after the other, e.g., \PSOmod, followed by \DEmod, followed by \CMAESmod. 

\subsection{Algorithm template used by \texorpdfstring{\MetafoR}~}\label{sec:Algorithm template used by MetafoR}

\begin{algorithm}
{\small 
	\caption{High-level structure of the algorithm template used by \MetafoR}\label{alg:genMetafor}
	\begin{algorithmic}[1]
		\Require Set of \textsf{parameters ($\Pi$)} 
		\State \texttt{config} $\gets$ \Call{Validate}{}($\Pi$, \MetafoRModulesAlg) \label{alg:line:validate}
		\State \texttt{exec} $\gets$ \Call{SetExecutionMode}{}(\texttt{config.get(\Execution})) \label{alg:line:initExec}
		\State \texttt{pop} $\gets$ \Call{Initialize}{}(\texttt{config.get(\Pop)}, \texttt{config.getDynamic}) \label{alg:line:initPop}
		\Repeat \label{alg:line:main_loop_start}
		\For{$i \gets 1$ \textbf{to} size(\texttt{pop})} \label{alg:line:updatePopLoop:begin}
		\State \texttt{pop}$^{\,i}$ $\gets$ \Call{UpdateSolution}{}(\texttt{pop}$^{\,i}$, \texttt{exec}) \label{alg:line:updatePop}
		\EndFor \label{alg:line:updatePopLoop:end}
		
		\State \texttt{best} $\gets$ \Call{GetBestSolution}{}(\texttt{pop})\label{alg:line:getBest}
		
		\If{\LSmod is \texttt{enabled}} \label{alg:line:isLocalSearchEnabled}
		\State \Call{PerformLocalSearch}{\texttt{best}, \texttt{config.getLS}}\label{alg:line:applyLocalSearch}
		\EndIf
		
		\State apply \Reinit \Comment{optional} \label{alg:line:reinitializeSolutions}
		\State \texttt{pop} $\gets$ \Call{UpdatePopulationParameters}{\texttt{config.get(\Pop)}, \texttt{config.getDynamic}} \label{alg:line:popoParamUpdate}
		\State \texttt{exec} $\gets$ \Call{UpdateExecutionParameters}{}(\texttt{exec})\label{alg:line:execVariablesUpdate}
		\Until{termination criterion is met} \label{alg:line:main_loop_end}
		\State \textbf{return} \texttt{best}
	\end{algorithmic}
}
\end{algorithm}

In \MetafoR, the algorithm components that make up a \mh implementation are combined using \Alg~\ref{alg:genMetafor}.
As shown in line~\ref{alg:line:validate} of \Alg~\ref{alg:genMetafor}, after receiving the set of \textsf{parameters} $\Pi$, the first action performed by \MetafoR is to validate their values and dependence.
The procedure \textsc{Validate()} checks that the set of parameters input by the user are within the defined limits and that there are no conflicting or missing options. 
For example, if the user executes \MetafoR indicating the use of \CMAESmod and \DEmod, but the parameters needed to create the implementation are incomplete or their values are out range, \textsc{Validate()} stops the execution and prints out the missing, conflicting and/or out-of-range values.
If \textsc{Validate()} ends successfully, it creates an object called \texttt{config} that stores the options with which \MetafoR is executed.
\MetafoR has implemented different ways to access \texttt{config}, for example, \texttt{config.get($\cdot$)} retrieves the algorithm component option or parameter value indicated as argument, \texttt{config.getLS} retrieves the set of parameters used by \LSmod, and \texttt{config.getDynamic} retrieves the set of algorithm components and parameters whose value change over time, such as \TopTVFI, 
\CMAESPopInc, \CMAESMtxCovDia, etc.

In line~\ref{alg:line:initExec}, the procedure \textsc{SetExecutionMode}() is performed to initialize the variables required by the algorithm component \Execution, such as the probability $pr$ and the execution budget $xb$.
Then, in line~\ref{alg:line:initPop}, the procedure \textsc{Initialize}() creates the initial population (\texttt{pop}) and initializes the variables and structures needed to handle the population throughout the execution. 
\textsc{Initialize}() also takes care of the structures and variables required by dynamic algorithm components, such as variable \texttt{t\_schedule} which is used to determine when connections are removed from \TopTVFI.

The main optimization process takes place inside the loop that goes from line~\ref{alg:line:main_loop_start} to line~\ref{alg:line:main_loop_end}. 
In line~\ref{alg:line:updatePop}, the \textsc{UpdateSolution}() procedure is applied to each individual in the population based on the type of \Execution implemented and the modules selected by the user.
When \texttt{exec} is set to \ExecutionCB, the same set of components is applied to every individual until the termination criterion is met. 
When \texttt{exec} is set to \ExecutionPB, the specific set of components applied to an individual alternates probabilistically based on the parameter $pr$. Finally, when \texttt{exec} is set to \ExecutionMP, two or three different modules are applied in a pipeline, where each module is assigned its own budget $xb$, and the execution of a module begins only after the budget of the previous module is exhausted.
The \textsc{UpdateSolution}() procedure also handles the update of the variable associated to each solution (e.g., the personal best solution vectors) and the algorithm components options that have to update solution-wise (e.g., \OmegaSelfReg, \MagSucc and \DEVectDifferences).
In line~\ref{alg:line:getBest}, after updating the population, the best solution is kept in the variable \texttt{best}. 

In line~\ref{alg:line:isLocalSearchEnabled}, \MetafoR checks whether \LSmod is enabled and, if this is the case, the procedure \textsc{PerformLocalSearch}() is executed in line~\ref{alg:line:applyLocalSearch} using the \texttt{best} solution as input.
If the \ls algorithm finds an improved solution, \textsc{PerformLocalSearch}() updates \texttt{best} with the new solution.
In line~\ref{alg:line:reinitializeSolutions}, the optional component \Reinit takes place.
There are two options for implementing \Reinit: \RIchange and \RIsimilarity.
\RIchange re-initializes the population if either of the following conditions is verified: (i) if the standard deviation of the solutions falls below $10^{-3}$, or (ii) if the total change in the objective function over the last $(10\cdot d)/\text{size}(\texttt{pop})$ iterations is less than $10^{-8}$.
The \RIsimilarity option computes the Euclidean distance between a solution and $\vec{x}^\text{best}$ and re-initializes those solutions whose distance to $\vec{x}^\text{best}$ is lower than $10^{-3}$.
Finally, in lines~\ref{alg:line:popoParamUpdate}--\ref{alg:line:execVariablesUpdate}, \MetafoR updates the variables and structures related to the population (\texttt{pop}) and the execution (\texttt{exec}).
The procedure \textsc{UpdatePopulationParameters}() also takes care of the dynamic algorithm components whose values have to be updated iteration-wise (e.g., \PopIncre, \TopTVFI and \CMAESMtxCovDia).

\MetafoR has a total of 104 parameters, of which 53 belong to \PSOmod, 5 to \DEmod, 9 to \CMAESmod, 15 to \LSmod, 7 to the algorithm component \Execution and 15 are shared by several modules (e.g. \Pop, \VectBasis, \Reinit, etc.).
As shown below in \Tbl~\ref{table:ParameterSettingsALL}, the actual number of parameters used in our automatically created implementations vary from as few as 10 parameters (e.g., \CMAESldo) to 32 parameters (e.g., \PCHybtfo).
In \MetafoR, optional algorithm components can simply be omitted from the set of execution parameters when they are not going to be used in the implementation.


\section{Experimental Study}\label{sec:ExperimentalStudy}

We conducted experiments on a set of 50 continuous functions belonging to the CEC'05 and CEC'14 "Special Session on Single Objective Real-Parameter Optimization" \cite{SugHanLia2005cec,LiaQu2013problem} and to the Soft Computing (SOCO'10) "Test Suite on Scalability of Evolutionary Algorithms and other Metaheuristics for Large Scale Continuous Optimization Problems" \cite{HerLozMol2010test}.\footnote{Due to space limitation, we refer the reader to the provided references for a complete description of the benchmark functions.}
The goal of using a benchmark set of functions composed of various test suites is to include as many as possible of the characteristics that make \cops difficult, 
such as mathematical transformations (e.g., translations and rotations), multiple local optima and non-separable objective functions.
As shown in \Tbl~\ref{table:benchmark functions}, our benchmark set is composed of 12 unimodal functions ($f_{1-12}$), 14 multimodal functions ($f_{13-26}$), 14 hybrid functions ($f_{27-40}$), and 10 hybrid composition functions ($f_{41-50}$).
With the exception of $f_{41}$, none of the hybrid composition functions is separable, and functions $f_{42-50}$ include an additional rotation in the objective function.

\subsection{Benchmark functions}
\begin{table*}[ht!]
\centering
{\scriptsize
	\caption{Benchmark functions}
	\label{table:benchmark functions}
	\begin{tabular}{ c m{2.9cm} C{1.4cm} m{0.8cm} | C{0.3cm} m{4.7cm} C{1.2cm} m{0.8cm}}
		\toprule 
		\textbf{$f_{\#}$} & \textbf{Name} & \textbf{Search Range} & \textbf{Suite} & \textbf{$f_{\#}$} & \textbf{Name}  & \textbf{Search range} & \textbf{Suite}\\ 
		\midrule 
		$f_1$ & Shifted Sphere & [-100,100] & SOCO'10 & 
		$f_{26}$ & Shifted Rotated HGBat & [-100,100] & CEC’14 \\
		$f_2$ & Shifted Rotated High Conditioned Elliptic & [-100,100] & CEC'14 & 
		$f_{27}$ & Hybrid Function 1 (N = 2) & [-100,100] & SOCO'10 \\
		$f_3$ & Shifted Rotated Bent Cigar  & [-100,100] & CEC’14 & 
		$f_{28}$ & Hybrid Function 2 (N = 2) & [-100,100] & SOCO'10 \\
		$f_4$ & Shifted Rotated Discus & [-100,100] & CEC’14 & 
		$f_{29}$ & Hybrid Function 3 (N = 2) & [-5,5] & SOCO'10 \\
		$f_5$ & Shifted Schwefel 22.1 & [-100,100] & SOCO'10 & 
		$f_{30}$ & Hybrid Function 4 (N = 2) & [-10,10] & SOCO'10 \\
		$f_6$ & Shifted Schwefel 1.2 & [-65.536,65.536] & SOCO'10 & 
		$f_{31}$ & Hybrid Function 7 (N = 2) & [-100,100] & SOCO'10 \\
		$f_7$ & Shifted Scfewels12 noise in fitness & [-100,100] & CEC’05 & 
		$f_{32}$ & Hybrid Function 8 (N = 2) & [-100,100] & SOCO'10 \\
		$f_8$ & Shifted Schwefel 2.22 & [-10,10] & SOCO'10 & 
		$f_{33}$ & Hybrid Function 9 (N = 2) & [-5,5] & SOCO'10 \\
		$f_9$ & Shifted Extended $f_{10}$ & [-100,100] & SOCO'10 & 
		$f_{34}$ & Hybrid Function 10 (N = 2) &  [-10,10] & SOCO'10 \\
		$f_{10}$ & Shifted Bohachevsky & [-100,100] & SOCO'10 & 
		$f_{35}$ & Hybrid Function 1 (N = 3) & [-100,100] & CEC’14 \\
		$f_{11}$ & Shifted Schaffer & [-100,100] & SOCO'10 & 
		$f_{36}$ & Hybrid Function 2 (N = 3) & [-100,100] & CEC’14 \\
		$f_{12}$ & Shchwefel 2.6 Global Optimum on Bounds & [-100,100] & CEC’05 & 
		$f_{37}$ & Hybrid Function 3 (N = 4) & [-100,100] & CEC’14 \\
		$f_{13}$ & Shifted Ackley & [-32,32] & SOCO'10 & 
		$f_{38}$ & Hybrid Function 4 (N = 4) & [-100,100] & CEC’14 \\
		$f_{14}$ & Shifted Rotated Ackley & [-100,100] & CEC’14 & 
		$f_{39}$ & Hybrid Function 5 (N = 5) & [-100,100] & CEC’14 \\
		$f_{15}$ & Shifted Rosenbrock & [-100,100] & SOCO'10 & 
		$f_{40}$ & Hybrid Function 6 (N = 5) & [-100,100] & CEC’14 \\
		$f_{16}$ & Shifted Rotated Rosenbrock & [-100,100] & CEC’14 & 
		$f_{41}$ & Hybrid Composition Function & [-5,5] & CEC’05 \\
		$f_{17}$ & Shifted Griewank & [-600,600] & SOCO'10 & 
		$f_{42}$ & Rotated Hybrid Composition Function & [-5,5] & CEC’05 \\
		$f_{18}$ & Shifted Rotated Griewank & [-100,100] & CEC’14 & 
		$f_{43}$ & Rotated H. Composition F. with Noise in Fitness & [-5,5] & CEC’05 \\
		$f_{19}$ & Shifted Rastrigin & [-100,100] & SOCO'10 & 
		$f_{44}$ & Rotated Hybrid Composition F. & [-5,5] & CEC’05 \\
		$f_{20}$ & Shifted Rotated Rastrigin & [-100,100] & CEC’14 & 
		$f_{45}$ & Rotated H. Composition F. with a Narrow Basin for the Global Opt. & [-5,5] & CEC’05 \\
		$f_{21}$ & Shifted Schwefel & [-100,100] & CEC’14 & 
		$f_{46}$ & Rotated H. Comp. F. with the Gbl. Opt. On the Bounds & [-5,5] & CEC’05 \\
		$f_{22}$ & Shifted Rotated Schwefel & [-100,100] & CEC’14 & 
		$f_{47}$ & Rotated Hybrid Composition Function & [-5,5] & CEC’05 \\
		$f_{23}$ & Shifted Rotated WeierStrass & [-100,100] & CEC’05 & 
		$f_{48}$ & Rotated H. Comp. F. with High Condition Num. Matrix & [-5,5] & CEC’05 \\
		$f_{24}$ & Shifted Rotated Katsuura & [-100,100] & CEC’14 & 
		$f_{49}$ & Non-Continuous Rotated Hybrid Composition Function & [-5,5] & CEC’05 \\
		$f_{25}$ & Shifted Rotated HappyCat & [-100,100] & CEC’14 & 
		$f_{50}$ & Rotated Hybrid Composition Function & [-5,5] & CEC’05 \\
		\bottomrule
	\end{tabular} 
}
\end{table*}

In addition to evaluating the performance of the algorithms on the characteristics aforementioned, we are interested in their scalability, especially for problems with more than 100 dimensions.
In order to test how well the algorithms scale to large dimensional (LD) problems, we use the functions of the SOCO'10 test suite ($f_{1,5,6,8-11,13,15,17,19,27-34}$) with 200, 500 and 1$\,$000 dimensions.
In contrast to the CEC'2005 and CEC'2014 test suites, which are primarily designed to test the performance of the algorithms on highly complex, multimodal functions, the SOCO'10 test suite contains problems whose complexity is mainly due to their large size.

\subsection{Experimental setup}\label{sec:experimental setup}
Our experimental study involves two phases: a \textit{training phase}, where we use \irace with different training instances to create and configure a number of algorithms; and a \textit{testing phase}, where we run the algorithms on different sets functions 
and measure aspects related to their performance.
To guarantee that the training set and the testing sets are different enough, we excluded the CEC'14 functions from the training set, and use only the CEC'05 and SOCO'10 test suites to define two different training scenarios.
For the first scenario, we created a training set of 88 functions by randomly selecting 75\% of the functions in the CEC'05 test suite and 75\% of the functions in the SOCO'10 test suite, where the selected CEC'05 functions use $d=\{10, 30, 50\}$ and the selected SOCO'10 functions use $d=\{500, 1\,000\}$.
We refer to this first scenario as \TFOUT, since "25\%" of the total number of instances were left "out" from the training set.
For the second scenario, we consider a leave-one-class-out cross-validation approach by creating a training set of 107 functions where all the functions from both test suites are included but only with a low dimensional number. In other words, we left out the large dimensional class and included the 25 functions of the CEC'05 suite with $d=\{10, 30\}$ and the 19 functions of the SOCO'10 suite with $d=\{50, 100, 200\}$.
We refer to this second scenario as \LDO, where \texttt{LDO} stands for "largest dimensions out". 
In \Tbl~\ref{table:instance distrubution per scenarios}, we show the percentage of functions of each class included in each training scenario and in the CEC'05, CEC'14 and SOCO'10 test suites.

\begin{table}
{\fontsize{6.5}{6.5}\selectfont
	\caption{Percentage of functions of each class in scenarios \TFOUT, \LDO and in the CEC'05, CEC'14 and SOCO'10 test suites.}
	\label{table:instance distrubution per scenarios}
	\begin{tabular}{c c c c c c c c l}
		\toprule
		\textbf{Scenario/Suite} &\textbf{\# of Functions} & \textbf{UNI (\%)} & \textbf{MUL (\%)} & \textbf{HYB and HCP (\%)} & \textbf{ROT (\%)} & \textbf{LD (\%)} & \textbf{Separable (\%)} & \textbf{$d$ Range} \\
		\toprule
		\multirow{2}{*}{\TFOUT}  & \multirow{2}{*}{88} & \multirow{2}{*}{24.75} & \multirow{2}{*}{32.75} & \multirow{2}{*}{42.5} & \multirow{2}{*}{36} & \multirow{2}{*}{31} & \multirow{2}{*}{17} & \fCECFIVE $d=\{10,30,50\}$,\\
		& & & & & & & & \fSOCO $d=\{500,1000\}$\\
		\multirow{2}{*}{\LDO} & \multirow{2}{*}{107} & \multirow{2}{*}{28.5} & \multirow{2}{*}{28.5} & \multirow{2}{*}{43} & \multirow{2}{*}{36} & \multirow{2}{*}{0}  & \multirow{2}{*}{16} & \fCECFIVE $d=\{10,30\}$,\\
		& & & & & & & & \fSOCO $d=\{50,100,500\}$\\ 
		CEC'05  & 25 per $d$ & 20    & 36    & 44   & 16 & 0  & 12 &  $d=\{2, 10,30,50\}$ \\
		CEC'14  & 22 per $d$ & 14    & 59    & 27   & 63 & 0  & 9  &  $d=\{2, 10,30,50, 100\}$\\
		SOCO'10 & 19 per $d$ & 37    & 21    & 42   & 0  & 100 & 21 &  $d=\{10, \dots, \infty\}$\\
		\bottomrule
	\end{tabular}
}
\end{table}

To design and configure the algorithms in our comparison, we used \irace with different computational budgets. 
In the case of \CMAES and \DE, where the number of parameters is relatively small (less than 15 parameters), we used \irace with a budget of $10\,000$ executions, and in the case of \PSO and \MetafoR, which have 59 and 104 parameters, respectively, we used \irace with a budget of $30\,000$ and $50\,000$ executions.
In addition to these algorithms, we used \irace with a budget of $30\,000$ executions to create three hybrid implementations, a \PDHyb, a \DCHyb and a \PCHyb.
Our rationale was that, rather than manually selecting specific hybrids of \algInPaper for our comparison, it would be far more insightful to evaluate the performance of automatically generated hybrid implementations using the algorithm components and execution modes already implemented in \MetafoR.
By combining the seven parameter spaces (\MetafoR, \CMAES, \DE, \PSO, \PDHyb, \DCHyb and \PCHyb) and the two training scenarios (\TFOUT and \LDO), we produce a total of \textbf{14 algorithms}.
In the following sections, we study the performance of the 14 automatically created algorithms and \textbf{3 default variants} of \algInPaper using parameter settings.

In the reminder of this paper, we use the following abbreviations: "\MTF" for the implementations created using the entire search space of \MetafoR, "\StdPSO" for the \StanPSO algorithm, "DFT" for implementations using default parameter values, "UNI" for the unimodal functions ($f_{1-12}$), "MUL" for the multimodal functions ($f_{13-26}$), "HYB" for the hybrid functions ($f_{27-40}$), "HCP" for the hybrid composition functions ($f_{41-50}$), "ROT" for the rotated functions ($f_{2-4,6,14,16,18,20,22-26,42-50}$), "SHU" for the shifted unimodal functions in the SOCO'10 test suite ($f_{1,5,6,8-11}$).
We also use "\fCECFIVE", "\fCECFOURTEEN" and "\fSOCO" to refer, respectively, to the CEC'05, CEC'14 and SOCO'10 test suite, and we use "\fBENCHMARK" to refer to the benchmark set of 50 function shown in \Tbl~\ref{table:benchmark functions}.
The algorithm components and parameter settings of the \textbf{17 algorithms} included in our comparison 
are shown in \Tbl~\ref{table:ParameterSettingsALL}.

\begin{ThreePartTable}
\begin{TableNotes}
	\scriptsize
	\item[*] Conditional parameters and variables are shown inside brackets, preceded by the algorithm components on which they depend.
\end{TableNotes}

{\fontsize{6.5}{6.5}\selectfont
	\begin{longtable}{C{1cm} C{0.01cm} m{12.7cm}} 
		\caption{Parameter settings of the algorithms included in the comparison.}
		\label{table:ParameterSettingsALL}\\
		\toprule
		\multicolumn{2}{c}{\textbf{Algorithm}} & \multicolumn{1}{c}{\textbf{Settings}} \\
		\midrule
		\endfirsthead
		
		\caption*{ Table \ref{table:ParameterSettingsALL} (Continued.)}\\
		\toprule
		\multicolumn{2}{c}{\textbf{Algorithm}} & \multicolumn{1}{c}{\textbf{Settings}} \\
		\midrule
		\endhead
		
		\bottomrule
		\endfoot
		
		\bottomrule
		\insertTableNotes  
		\endlastfoot
		
		
		\multicolumn{2}{c}\MTFtfo & \ExecutionMP (\CMAESmod, $xb_1 =0.8265$, \DEmod, $xb_2 =0.3535$), 
		$\CMAESpara=4.2552$, $\CMAESparb=2.2402$, $\CMAESparc=0.6247$, $\CMAESpard=3.6447$, $\CMAESpare=-18.3449$, $\CMAESparf=-19.0782$, $\CMAESparg=-14.4971$, \CMAESMtxDia, \CMAESRWlog, 
		\DEBVbest, $\DEVectDifferences=0.1198$, \DERcbBinomial, $p_a=0.4783$, $\beta=0.5197$, \VBeigen, \DEVectPos,
		\PopConst ($\texttt{pop}_{ini}=324$). \\
		\midrule
		\multicolumn{2}{c}\MTFldo & \ExecutionMP (\CMAESmod, $xb_1 =0.9154$, \DEmod, $xb_2 =0.1573$, \LSmod, $\LSbudget =0.579$ and $\LSdivide =3$), 
		\MTSLS ($\MTSLSbias =-0.0549$, $\MTSLSinitss = 0.3892$, $\MTSLSiterations =1.6483$), $\CMAESpara=7.7473$, $\CMAESparb=1.9539$, $\CMAESparc=0.6037$, $\CMAESpard=2.3443$, $\CMAESpare=-19.7597$, $\CMAESparf=-17.1491$, $\CMAESparg=-12.1016$, \CMAESMtxCovDia, \CMAESRWlog, 
		\DEBVdirRand, \DEVectDifferences$=0.0488$, \DERcbBinomial, $p_a=0.8117$, $\beta=0.3273$, \VBnatural, \DEVectPos, \PopConst ($\texttt{pop}_{ini}=252$),
		\RIsimilarity. \\
		
		\midrule
		\multicolumn{2}{c}\CMAESdft & \ExecutionCB (\CMAESmod), $\CMAESpara=3$, $\CMAESparb=2$, $\CMAESparc=0.5$, $\CMAESpard=2$, $\CMAESpare=-12$, $\CMAESparf=-20$, $\CMAESparg=-12$, \CMAESMtxCov, \CMAESRWlog. \\
		\midrule
		\multicolumn{2}{c}\CMAEStfo & \ExecutionCB (\CMAESmod),
		$\CMAESpara=9.3314$, $\CMAESparb=3.9293$, $\CMAESparc=0.1237$, $\CMAESpard=3.7189$, $\CMAESpare=-14.5256$, $\CMAESparf=-15.7226$, $\CMAESparg=-16.3546$, \CMAESMtxDia, \CMAESRWlog. \\
		\midrule
		\multicolumn{2}{c}\CMAESldo & \ExecutionCB (\CMAESmod), $\CMAESpara=3.6811$, $\CMAESparb=2.8753$, $\CMAESparc=0.1583$, $\CMAESpard=3.8432$, $\CMAESpare=-19.9498$, $\CMAESparf=-19.7584$, $\CMAESparg=-14.4405$, \CMAESMtxDia, \CMAESRWlog. \\
		
		\midrule
		\multicolumn{2}{c}\DEdft & \ExecutionCB (\DEmod), \VBnatural, \DEBVbest, $\DEVectDifferences=1$, \DERcbBinomial, $p_a=0.5$, $\beta=0.9$, \PopConst ($\texttt{pop}_{ini}=d$). \\
		\midrule
		\multicolumn{2}{c}\DEtfo & \ExecutionCB (\DEmod), \VBnatural, \DEBVbest, $\DEVectDifferences=0.0151$, \DERcbExponential, $p_a=0.3566$, $\beta=0.2614$, 
		\PopTV ($\texttt{pop}_{ini}=39$, $\texttt{pop}_{min}=39$, $\texttt{pop}_{max}=132$, $\texttt{pop}_{sch}=26$, \InitRandom). \\
		\midrule
		\multicolumn{2}{c}\DEldo & \ExecutionCB (\DEmod), \VBnatural, \DEBVbest, $\DEVectDifferences=0.1711$, \DERcbExponential, $p_a=0.2105$, $\beta=0.197$, \PopIncre ($\texttt{pop}_{min}=46$, $\texttt{pop}_{max}=69$, $\texttt{pop}_{add}=5$, \InitHorizontal). \\
		
		\midrule
		\multicolumn{2}{c}\StdPSOdft & \ExecutionCB (\PSOmod), \DNPPRect, \TopFC, \MoiBoN, \MtxDiagonal, \OmegaCons ($\omega_{1} =0.749$), $\omega_{2} = 1$, \PopConst ($\texttt{pop}_{ini}=40$), $\varphi_1 = 1.496180$, $\varphi_2 = 1.496180$, \VelClp. \\
		\midrule
		\multicolumn{2}{c}\PSOtfo & \ExecutionCB (\PSOmod), 
		\VBnatural, \OperatorS, \TopWheel, \MoiRFI, $\varphi_1 = 0.0967$, $\varphi_2 = 0.9359$, \OmegaRnkBsd ($\omega_{1,min} = 0.6134$, $\omega_{1,max} = 0.944$), $\omega_{2} = \omega_{1}$, $\omega_{3} = 1$, \PertGau (\MagSucc, $pm_{1}= 0.211$, $\textsf{par\_sc}_1 = 46$, $\textsf{par\_fc}_1 = 24$), \PertRect (\MagSucc $pm_{2}= 0.498$, $\textsf{par\_sc}_2 = 45$, $\textsf{par\_fc}_2 = 42$), 
		\PopIncre ($\texttt{pop}_{min}=5$, $\texttt{pop}_{max}=20$, $\texttt{pop}_{add}=2$, \InitHorizontal). \\
		\midrule
		\multicolumn{2}{c}\PSOldo & \ExecutionCB (\PSOmod), \VBnatural, \DNPPRect, \MoiBoN, \TopTVFI ($\textsf{par\_tSch} =8$), \PhiExtra, \OmegaLinDec ($\omega_{1,min} = 0.1156$, $\omega_{1,max} =0.4517$, $\omega_{sch} =5$), $\omega_{2} = \OmegaRnd$, \PertLev (\MagSucc, $\PMcomp_{1}= 0.8248$, $\textsf{par\_sc}_1 = 2$, $\textsf{par\_fc}_1 = 32$), \MtxEuclideanOne ($\textsf{par\_angle}=7$), \PopIncre ($\texttt{pop}_{min}=2$, $\texttt{pop}_{max}=34$, $add(\texttt{pop})=8$, \InitRandom),
		\VelClp, \StagDet, \RIchange. \\
		
		\midrule
		\multicolumn{2}{c}\PDHybtfo & \ExecutionMP (\DEmod, $xb_1 =0.7385$, \PSOmod, $xb_2 =0.9477$, \ExecutionPBuni, $\ExecutionParStd =0.7304$) 
		\VBnatural, \OmegaCons ($\omega_{1} =0.0205$), $\omega_{2} = \OmegaRnd$, \OperatorN, \MoiBoN, \TopTVFI ($\textsf{par\_tSch}= 9$), \DEBVttbest, $\DEVectDifferences=0.0477$, \DERcbExponential, $p_a=0.8266$, $\beta=0.7779$, \DEVectMix, 
		\PopConst ($\texttt{pop}_{ini}=24$). \\
		\midrule
		\multicolumn{2}{c}\PDHybldo & \ExecutionPB (\DEmod, $xb_1 =0.6355$, \PSOmod, $xb_2 =0.6838$), \ExecutionPBlev ($\ExecutionParStd =0.4762$), \VBnatural, \OperatorN, \TopNeum, \MoiBoN, \OmegaSuccBsd ($\omega_{1,min} = 0.1178$, $\omega_{1,max} =0.6264$), $\omega_{2} =1$, \DEBVbest, $\DEVectDifferences=0.0801$, \DERcbExponential, $p_a=0.3857$, $\beta=0.3817$, \DEVectPos, \PopTV ($\texttt{pop}_{ini}=29$, $\texttt{pop}_{min}=18$, $\texttt{pop}_{max}=253$, $\texttt{pop}_{sch}=71$, \InitRandom). \\
		
		\midrule
		\multicolumn{2}{c}\DCHybtfo & \ExecutionMP (\CMAESmod, $xb_1 =0.8695$, \DEmod, $xb_2 =0.2598$),
		$\CMAESpara=7.7115$, $\CMAESparb=1.4923$, $\CMAESparc=0.4212$, $\CMAESpard=3.4793$, $\CMAESpare=-18.5851$, $\CMAESparf=-18.072$, $\CMAESparg=-14.2609$, \CMAESMtxCovDia, \CMAESRWlog, \DEBVrand, $\DEVectDifferences=0.1683$, \DERcbExponential, $p_a=0.9742$, $\beta=0.339$, \DEVectPos, \VBnatural,
		\PopConst ($\texttt{pop}_{ini}=66$),
		\RIsimilarity. \\
		\midrule
		\multicolumn{2}{c}\DCHybldo & \ExecutionMP (\CMAESmod, $xb_1 =0.8677$, \DEmod, $xb_2 =0.7543$), 
		$\CMAESpara=4.2166$, $\CMAESparb=1.7168$, $\CMAESparc=0.1898$, $\CMAESpard=3.4845$, $\CMAESpare=-19.8045$, $\CMAESparf=-19.7732$, $\CMAESparg=-18.6365$, \CMAESMtxDia, \CMAESRWlog,
		\VBeigen, \DEBVdirRand, $\DEVectDifferences=0.1658$, \DERcbBinomial, $p_a=0.96$, $\beta=0.5775$, \DEVectPos, \PopConst ($\texttt{pop}_{ini}=459$). \\
		
		\midrule
		\multicolumn{2}{c}\PCHybtfo & \ExecutionMP (\CMAESmod, $xb_1 =0.9695$, \PSOmod, $xb_2 =0.2238$),
		\VBnatural, \OperatorCG (\OperatorQrandNeigh, $\textsf{parm\_r} = 0.4139$), \TopNeum, \MoiFI, \OmegaDoubExp ($\omega_{1,max} = 0.2985$), $\omega_{2} = 0.2569$, \PertLev (\MagCons, $pm_{1} = 0.2303$), 
		$\CMAESpara=9.0134$, $\CMAESparb=1.2005$, $\CMAESparc=0.7269$, $\CMAESpard=3.1025$, $\CMAESpare=-19.9248$, $\CMAESparf=-19.8913$, $\CMAESparg=-17.4091$, \CMAESMtxCovDia, \CMAESRWlog, 
		\PopTV ($\texttt{pop}_{ini}=181$, $\texttt{pop}_{min}=38$, $\texttt{pop}_{max}=367$, $\texttt{pop}_{sch}=23$, \InitHorizontal),
		\RIchange. \\
		\midrule
		\multicolumn{2}{c}\PCHybldo & \ExecutionMP (\CMAESmod, $xb_1 =0.9543$, \PSOmod, $xb_2 =0.2452$), \VBnatural, \OperatorCG (\OperatorQrandNeigh, $\textsf{parm\_r}= 0.7225$), \TopTVFI, \MoiFI, $\textsf{par\_tSch} =10$), \OmegaCons ($\omega_{1} =0.9485$), $\omega_{2} = \omega_{1}$, 
		$\CMAESpara=8.3477$, $\CMAESparb=2.3073$, $\CMAESparc=0.8218$, $\CMAESpard=2.9854$, $\CMAESpare=-19.3423$, $\CMAESparf=-19.4916$, $\CMAESparg=-19.1069$, \CMAESMtxCovDia, \CMAESRWlog, \PopIncre ($\texttt{pop}_{min}=22$, $\texttt{pop}_{max}=489$, $\texttt{pop}_{add}=1$, \InitHorizontal),
		\RIsimilarity. \\
	\end{longtable}
}
\end{ThreePartTable}

For the testing phase, we perform 50 independent runs of each of the 17 algorithms on each function and report the median (MED) solution produced by the algorithms, the median solution error (MEDerr) with respect to the best solution found by any of them, the median absolute deviation (MAD) and the wall-clock execution time in seconds.
In all cases, the algorithms were stopped after reaching $5\,000 \times d$ objective FEs or a wall-clock time of 600 seconds for $d \leq 50$, $1\,800$ seconds for $50 < d \leq 100$, $6\,000$ seconds for $100 < d \leq 500$, and $9\,000$ seconds for $500 < d \leq 1\,000$.
Both the tuning and the testing of the algorithms were done on dual core AMD Epyc Rome 7452 running at 2.2 GHz with 512 Kb cache size under \texttt{Rocks Mamba GNU/Linux CentOS 6.3}.
The version of \irace is 3.5.
\MetafoR is coded in \texttt{C++}, was compiled using \texttt{gcc} 9.4.0 and its source code is publicly available from: \url{https://github.com/clcamachov/METAFOR}.

\subsection{Analysis of Results}
We divide our experiments into three parts.
In the first part, we investigate the advantages that hybrid designs have over default and configured single-approach implementations.
Our goal is to identify both the approaches that work best for each class of functions and the main algorithm components responsible for their performance. 
To do this, we compare the result obtained by nine variants of \algInPaper with the six automatically generated hybrids of these approaches on \fBENCHMARK with $d=50$. 

In the second part of our experiments, we compare the configured and hybrid implementations of \algInPaper
with the automatically generated \MTF implementations.
The purpose of this analysis is to know the extent of the benefits of exploring the entire parameter space of \MetafoR, which has a larger number of parameters compared to the rest of the algorithms and requires a larger computational budget, versus fine-tuning implementations whose main design choices were manually selected by their algorithm designers. 
For this analysis, we consider the functions $f_{1-40}$ in \fBENCHMARK with $d=100$ and $f_{41-50}$ in \fBENCHMARK with $d=50$.

In the third part of our experimental analysis, we study the algorithms based on how well they generalize to other kinds of problems involving different combinations of characteristics from those used to train them. 
Since our configured and hybrid implementations were trained using different sets of function instances, we expect the implementations to exhibit different degrees of generalization to functions that were not included in their training sets.
Therefore, by comparing the performance of the algorithms on different function classes, we intend to highlight the primary advantages and limitations of the two training strategies.
For this analysis, we also consider the six configured single-approach implementations, the six \algInPaper hybrids, and the two \MTF implementations on the 22 functions of \fCECFOURTEEN with $d=\{50, 100\}$ and the 19 functions of \fSOCO with $d=\{750, 1250\}$.

To facilitate the calculation of the MED, MEDerr and MAD metrics (reported in the tables below) and to have suitable scales to visualize the results in box-plots (reported in the supplementary material of this article), we 
limit the maximum (worst) value returned by any of the algorithms to $1.00e+{10}$.
In particular, in functions $f_{3,8,15,28,30,32,36}$, not all algorithms were able to improve their initial solution, which is on the order of $1.00e+{308}$, resulting in extremely large numbers in some cases.
The limit of $1.00e+{10}$ was set based on the results of the worst performing configured algorithms and it is at least one order of magnitude higher than any of the worst returned  values. 
In the supplementary material of the article, we include both raw data, processed data with a limit value of $1.00e+{10}$, and the scripts we used for post-processing for replicability purposes.

\Tbls~\ref{table:MED-MEDerr-MAD-dftVStuned-50} to \ref{table:MED-MEDerr-MAD-LD-1250} show the averages of the MED, MEDerr, MAD and the ranking of the algorithms grouped by function classes. 
The correspondence between classes of functions and the functions listed in \Tbl~\ref{table:benchmark functions} is the following:
in \Tbls~\ref{table:MED-MEDerr-MAD-dftVStuned-50} and \ref{table:MED-MEDerr-MAD-mtfVSHyb-100-50}, UNI ($f_{1-12}$), MUL ($f_{13-26}$), HYB ($f_{27-40}$), HCP ($f_{41-50}$), ROT ($f_{2-4,6,14,16,18,20,22-26,42-50}$); 
and in \Tbls~\ref{table:MED-MEDerr-MAD-LD-750} and \ref{table:MED-MEDerr-MAD-LD-1250}, SHU ($f_{1,5,6,8-11}$), MUL ($f_{13,15,17,19}$) and HYB ($f_{27-34}$).
In the case of \Tbls~\ref{table:MED-MEDerr-MAD-cec14-50} and \ref{table:MED-MEDerr-MAD-cec14-100}, the functions listed in the tables are those defined for \fCECFOURTEEN~\cite{LiaQu2013problem} (listed only in the supplementary material of the article) and are grouped as follows: UNI ($f_{1-3}$), MUL ($f_{4-16}$), HYB ($f_{17-22}$), ROT ($f_{1-7,9-16}$).
The grouping ALL corresponds to all functions listed in the table, except for \Tbl~\ref{table:MED-MEDerr-MAD-mtfVSHyb-100-50}, where ALL corresponds to functions $f_{1-40}$ with $d=100$.

The limits for the scalability of the functions vary across test suites: for those belonging to \fCECFIVE, the limit is $d=50$, for those belonging to \fCECFOURTEEN, the limit is $d=100$, and for those belonging to \fSOCO, the limit is arbitrarily large.
In the row ``Wilcoxon test'', we use the symbol $\approx$ to indicate those cases in which the Wilcoxon test with confidence at 0.95 using Bonferroni's correction did not return a p-value lower than $\alpha = 0.05$, and the symbol $+$ to indicate those cases in which the difference was significant (i.e., $p< \alpha$).
The statistical tests were conducted to the complete data sets of MED, MEDerr and MAD values.
Finally, in the row "Wins", we show the number of times each algorithm found the best solution. 
Even though we include the number of "wins" for competitive comparison, our analysis primarily relies on more robust performance metrics, such as rankings and statistical tests.

\subsubsection{To hybridize, or not to hybridize --- which approach works best?}
\label{sec:results_exp1}
One of the main questions we want to explore in this work is how hybrid algorithms compare to single-approach implementations.
To answer this question, we compare the results obtained by the configured and default variants of \algInPaper with that of their automatically generated \algInPaperHyb on \fBENCHMARK with $d=50$ (\Tbl~\ref{table:MED-MEDerr-MAD-dftVStuned-50}). 
We focus on the functions with $d=50$ because this is the largest dimension available for all 50 functions in \fBENCHMARK.

\input{tables/tbl_dftVstuned}

The first thing to note in \Tbl~\ref{table:MED-MEDerr-MAD-dftVStuned-50} is that, while two of the three hybrids (namely \DCHyb and \PCHyb) ranked better that most single-approach implementations across all functions, in the case of the three variants of \CMAES and \DEldo, the differences in performance were not statistically significant.
Based on the MED and MEDerr metrics, the algorithm with the highest ranking across all functions is \PCHybldo, the second-best is \CMAESldo, and the third-best is \PCHybtfo.
However, when we split the result by classes of functions, we observe that \PCHybldo ranks either first or second in the UNI, MUL and ROT functions, but it is among the worst ranked in the HYB and HCP functions, where it was outperformed by the two \PDHybs, the two \DCHybs and the three \CMAES variants.

In both the HYB and HCP functions, the best ranked algorithm was \CMAESdft.
Based on the Av.MEDerr obtained by \PCHybldo, which is similar to that of \CMAESdft in the HYB functions and one order of magnitude higher in the HCP functions, \PCHybldo seems to be able to get close to the best solution, but fails to converge to it.
The lack of convergence of \PCHybldo to the best solution is caused by the use of algorithm components heavily biased toward exploration, such as \OperatorQrandNeigh, a large value of $\omega_1 = 0.9485$ and \RIsimilarity (see \Tbl~\ref{table:ParameterSettingsALL}).
In particular, \RIsimilarity, which re-initializes solutions whose Euclidean distance to $\vec{x}^\text{best}$ is lower than $10^{-3}$, seems counterproductive at latest stages of the optimization process.
In the case of \PCHybtfo, the implementation of \PSO also uses components with a strong exploration bias (e.g., \PertLev and \RIchange), which hinder the ability of the algorithm to converge to the global optimum.

An interesting finding is that all three \CMAES variants performed as well as any of the automatically generated hybrids, including the implementation of \CMAES using default parameter settings.
Although both \CMAEStfo and \CMAESldo produce better MED solutions and have smaller MEDerrs than \CMAESdft across all functions, \CMAESdft ranks first in both metrics in the HYB and HCP functions. 
There are two main reasons that explain these results.
The first one is that our implementation of \CMAESdft is not the basic $(\mu,\lambda)$-\CMAES, but the restart-\CMAES with increasing population size, most commonly known as IPOP-\CMAES~\cite{AugHan2005cec}. 
In the original IPOP-\CMAES, the parameter values were determined based on factors such as the dimensionality of the problem (e.g., $\lambda_{t=0} = 4 + \lfloor 3 \ln(d)\rfloor $) and detailed empirical analyses conducted on a limited set of simple test functions (e.g., $\texttt{stopTolFun} = 10^{-12}$).
Our method for creating and configuring \CMAEStfo and \CMAESldo builds upon the parameterized formulation of IPOP-\CMAES introduced by Liao \etal~\cite{LiaMolStu2015}. It involves employing \irace to identify combinations of algorithm components and parameter settings that achieve the best generalization across the training instances.
This explains why the automatically generated \CMAEStfo and \CMAESldo produce on average better MED and MEDerrs values than \CMAESdft.

However, the main reason why \CMAESdft performs better than any of the compared algorithms on the HYB and HCP functions is that, unlike \CMAEStfo and \CMAESldo, which use the less computationally expensive \CMAESMtxDia component, \CMAESdft uses the more computationally expensive (but also more accurate) \CMAESMtxCov. 
As discussed in \cite{RosHan2008:sepCMAES}, the algorithm component \CMAESMtxDia provides a similar performance to \CMAESMtxCov on separable functions; however, on non-separable functions, it can negatively impact the quality of the solutions found by the algorithm, especially for functions with highly complicated landscapes, such as HYB and HCP functions.
Although it is not completely clear why none of the automatically generated hybrids or configured implementations make use of \CMAESMtxCov in their design, we believe this is due to the constraint we put on the execution time of the algorithms.
By constraining the execution of the algorithms using wall-clock time, the more computationally intensive implementations generated by \irace were less likely to complete their execution within the available function evaluations (FEs). This approach inherently favored the use of less computationally demanding components, such as \CMAESMtxDia instead \CMAESMtxCovDia, in the algorithms designs.


Several notable algorithmic differences exist among \CMAESdft, \CMAEStfo and \CMAESldo, which merit detailed discussion.
Two minor differences are the initial population size ($\lambda_{t=0}$), which is approx. 2.5 times larger in \CMAEStfo than in \CMAESdft and \CMAESldo, and the number of parents selected for recombination, which is half of the population size in \CMAESdft and about a quarter of the population size in \CMAEStfo and \CMAESldo.
\CMAESdft uses an initial step size ($\sigma_{t=0} = 0.5$) that is almost four times larger than the one used in \CMAEStfo and \CMAESldo ($\sigma_{t=0} \approx 0.14$), resulting in a more diverse initial population in \CMAESdft compared to \CMAEStfo and \CMAESldo.
Another difference is that both \CMAEStfo and \CMAESldo have higher thresholds for triggering a restart than \CMAESdft. However, after restarting, the size of their populations increases by a factor of $\approx 4$, whereas in \CMAESdft, it grows by a factor of 2.

Among the six hybrid implementations, the combination that produced the best overall results was \PSO and \CMAES, followed by the combination of \DE and \CMAES, and finally, the combination of \PSO and \DE.
Both the type of hybridization (i.e., the option used for the algorithm component \Execution) and the order in which the modules are applied in the implementations (see \Tbl~\ref{table:ParameterSettingsALL} for details) were selected by \irace.
In the case of \PDHybtfo, the algorithm includes a \DE phase (allotted 46\% of the FEs) followed by a \PSO phase (allotted 56\% of the FEs), and in the case of \PDHybldo, the algorithm iteratively alternates between \DE and \PSO with a 50-50\% chance.
In the case of \PCHybtfo and \PCHybldo, which use the algorithm component \ExecutionMP, applying \CMAES first and \DE second turned out to be a better strategy than applying \DE first and \CMAES second.
This was also the case for \DCHybtfo and \DCHybldo, where \CMAES is applied before \DE.
As discussed in \Sect~\ref{sec:Previous works on hybridization}, although applying \CMAES before \PSO or \DE is not unusual in \CMAES hybrids, it seems a little counterintuitive to us because \DE and \PSO are often regarded as techniques with enhanced exploration capabilities than exploitation.

The algorithms ranked best according to the MAD metric are the two \PCHybs, followed by the two \DCHybs, followed by \CMAESldo.
As opposed to the MED and MEDerr, which provide information about the quality of the solutions, the MAD measures the variability of the results obtained by the algorithms across several independent runs.
In other words, the MAD is a measure of the robustness of the algorithms.
According to this metric, the overall performance of the \CMAES implementations was less competitive. Specifically, \CMAESdft was outperformed by all six hybrid algorithms, while both \CMAEStfo and \CMAESldo demonstrated performance comparable to that of \PDHybldo.
Nevertheless, the ranking of the algorithms per classes of function according to the MAD metric are consistent with the MED and MEDerr rankings per function classes, that is, \PCHybldo ranked first in the UNI, ROT and ALL functions, second in the MUL functions and among the last ones in the HYB and HCP functions, whereas \CMAESdft ranked second and first on the HYB and HCP functions, respectively.

The primary conclusion from this first part of our study is that, although single-approach algorithms such as \CMAES can achieve excellent results across a diverse set of problems---particularly with complex objective functions---hybrid implementations incorporating \CMAES are more robust and yield superior overall results compared to using \CMAES alone.
Allocating between $55\%$ and $75\%$ of the computational budget to \CMAES and the rest to either a \PSO or \DE seems to be a good strategy for hybridization.
Another conclusion of our study is that single-approach implementations of \PSO and \DE are less performing than their hybrid counterparts, in particular those that use default parameter values, which ranked consistently among the last places.
Note, however, that, in the case of \DEldo (similarly to \CMAES), we did not find any significant differences in its performance compared to the best algorithm (\PCHybldo) across the entire set of functions, and \DEldo showed to be as competitive as any of the hybrid implementations on the MUL, HYB and HCP functions.
In \Sect~\ref{sec:results_exp3}, we study the algorithmic design of \DEldo in more detail and create a link between its performance on HYB and HCP functions and on functions with large dimensionality.
Finally, although it can only be observed in \Tbls 4 and 5 in the supplementary material of the article, we found an inverse relationship between the number of dimensions in the functions and the performance of \CMAESdft.
According to the MED and MEDerr metrics, on average across all function classes, \CMAESdft ranked behind the other two \CMAES variants and all hybrids in \fBENCHMARK with $d=100$, and second to last in \fSOCO with $d=200$.



\subsubsection{Exploring new hybrid designs versus exploiting implementations with a fixed design --- what is the trade-off?}\label{sec:results_exp2}

To understand how exploring the entire search space of \MetafoR compares to fine-tuning implementations with predefined design choices, we create two implementations, \MTFtfo and \MTFldo, considering the 104 parameters of the framework.
In this section, we report the results obtained by the two \MTF implementations, the six configured \CMAES, \DE and \PSO implementations, and the six \algInPaperHyb on functions $f_{1-40}$ with $d=100$ and $f_{41-50}$ with $d=50$.
We excluded the default variants from the compared algorithms in this section to focus our analysis on comparing the performance of automatically created algorithms using different design spaces; however, in the supplementary material, we report the complete set of results obtained by the 17 algorithms on \fBENCHMARK with $d=\{50, 100, 200, 500, 1\,000\}$. 

\input{tables/tbl_mtfVSHyb}

As shown in \Tbl~\ref{table:MED-MEDerr-MAD-mtfVSHyb-100-50}, according to the MED and MEDerr metrics, the algorithm with the best overall performance was \MTFldo, followed by \PCHybldo, followed by \CMAESldo; and according to the MAD metric, the best algorithm was \MTFldo, followed by \DCHybtfo and finally \PCHybtfo.
For all three performance metrics, the second- and third-best ranked algorithms exhibited slightly superior performance than \MTFldo on the UNI and ROT functions, whereas \MTFldo was the clear winner in the MUL and HYB functions.
The results shown in \Tbl~\ref{table:MED-MEDerr-MAD-mtfVSHyb-100-50} also suggest that single-approach and hybrids involving \DE and \PSO are less competitive than single-approach and hybrids involving \CMAES.
This finding is supported by the average rankings across all functions and the Wilcoxon test using Bonferroni's correction, which returned a p-value lower than $\alpha = 0.05$ for \DEtfo, \DEldo, \PSOtfo, \PSOldo and the two \PDHybtfo and \PCHybldo.
Another important finding is that the algorithms trained on the \LDO scenario produce better results in the UNI, MOD and ROT functions, whereas those trained on the \TFOUT scenario were better in the HYB and HCP functions.
In \Sect~\ref{sec:results_exp3}, we study in detail the impact that different training strategies have on the performance of the algorithms. 

In the case of \PCHybldo and \PCHybtfo, which were among the top three algorithms in \fBENCHMARK with $d=50$, we observe that their inability to converge to the best solution on the HYB and HCP functions becomes increasingly problematic as the number of dimensions increases.
To try to understand why, we analyze the MED and MEDerr values obtained by \PCHybldo and \PCHybtfo on the HYB and HCP functions. 
Based on their MED and MEDerr, both per function and on average,  \PCHybldo and \PCHybtfo yield results comparable to those of the top-ranked algorithms in most cases. However, as noted in \Sect~\ref{sec:results_exp1}, while \PCHybldo and \PCHybtfo appear effective in identifying the search region containing the optimal solution, they struggle to converge to the solution itself.
In our opinion, this is due to the use of components with a strong exploratory bias, such as \Reinit, \PertInf and a large value for $\omega_1$, as well as the use of \CMAESMtxCovDia, which forces the algorithms to switch from the more accurate \CMAESMtxCov to the less accurate \CMAESMtxDia after $2+100 \times \frac{d}{\sqrt{\lambda}}$ FEs.
Although the algorithmic design of \PCHybldo and \PCHybtfo is well suited to UNI, MUL and ROT functions, it lacks the right exploration-exploitation balance to solve the most complicated HYB and HCP functions, which, based on our experiments, require an algorithmic design that allows intensive exploitation of the search space in the final stages of the optimization process.

We also study the main designs aspects of \MTFtfo and \MTFldo.
As shown in \Tbl~\ref{table:ParameterSettingsALL}, the two \MTF implementations are essentially \CMAES--\DE hybrids, but they have a number of design choices that make them unique.
In \MTFtfo, the implementation of \CMAES (allotted $\approx 62\%$ of the FEs) is quite similar to \CMAESdft, except for the choice of the algorithm component \CMAESMtx, which is implemented using \CMAESMtxDia, resulting in a faster, although less precise variant of \CMAES compared to \CMAESdft.
To compensate for any possible lack of exploitation capabilities, the implementation of \DE in \MTFtfo uses components \DEBVbest and \VBeigen, where \DEBVbest bias the search towards the best solution found so far, and \VBeigen (which is similar to the use of \CMAESMtxCov in \CMAES) provides \MTFtfo with a mechanism to effectively tackle rotated search spaces.

Of the two \MTF implementations, only \MTFldo includes a local search.
\MTFldo uses three modules, \CMAESmod, \DEmod and \LSmod, where each module is allocated, respectively, about 34\%, 6\% and 60\% of the total number of FEs.
The implementation of \CMAES in \MTFldo is also similar to \CMAESdft, except for two aspects. The first one is the use of the \CMAESMtxCovDia option, which allows the algorithm to take advantage of the information provided by the covariance matrix for a number of FEs and then switch to the faster diagonal matrix.
The second aspect is a higher tolerance for the range of improvement of all function values of the most recent generation (i.e., \texttt{stopTolFun}), which means that this implementation of \CMAES allows solution to remain for longer in the population before restarting.

The most salient aspect of \MTFldo is the use of the local search algorithm \MTSLS, which is executed after \CMAES and is assigned more than half of the computational budget. 
The parameter settings of \MTSLS are roughly the same used by its authors in~\cite{TseChe08:ieee-ec-ci}. 
However, the way in which \DE and \MTSLS interact with each other is quite interesting and has two main distinctive features.
The first feature is the obvious bias toward exploitation that results from allocating a much larger number of FEs to \MTSLS than to \DE in each iteration.
The second feature is the use of \DEBVdirRand, \RIsimilarity and a large crossover rate ($p_a = 0.81$) in the implementation of \DE, which allows it to generate solutions far away from the region where \MTSLS is locally searching.
By combining these two features, \MTFldo can perform an intensive exploitation of the best solution found so far, while keeping some degree of exploration of the search space up until the end of the optimization process.

\subsubsection{Generalizing to unseen problems --- which training strategy is the best?}\label{sec:results_exp3}

Our final analysis focuses on comparing the two training scenarios, \TFOUT and \LDO, and identifying their main advantages and disadvantages.
As mentioned in \Sect~\ref{sec:experimental setup}, the scenarios \TFOUT and \LDO differ in the set of functions available for \irace to train the algorithms.
We use a "machine learning"-type approach to select the functions in each scenario: in the case of \TFOUT, we randomly selected a fixed percentage of 75\% of the test functions of each class to form a stratified training set, and in the case of \LDO, we used a leave-one-class-out cross-validation approach to train the algorithms excluding the "large dimensional" (LD) class from the training set.
In practice, the training set of \TFOUT contains 88 functions that were selected using $\big(f_{1,3-7,9-12,14-17,19-23,25} \subset$ \fCECFIVE $\times\, d=\{10,30,50\}\big) \bigcup \big(f_{1, 3-6, 9-11, 13-15, 17-19} \subset$ \fSOCO $\times\, d=\{500, 1\,000\}\big)$, whereas the training set of \LDO contains 107 functions that were selected using $\big(f_{1-25} \subset$ \fCECFIVE $\times\, d=\{10,30\}\big) \bigcup \big(f_{1-19} \subset$ \fSOCO $\times \,d=\{50,100,200\}\big)$. 
In \Tbl~\ref{table:instance distrubution per scenarios}, we show the composition of each scenario in terms of functions classes.

Although no single instance separation strategy is optimal for training metaheuristic algorithm, each strategy represents a trade-off between generalization and specialization.
When we defined scenarios \TFOUT and \LDO, our intuition was that, since all classes of functions are represented in \TFOUT, from UNI to LD, \TFOUT should produce algorithms that generalize better at the expense of a lower solution quality.
In contrast, for \LDO, we hypothesized that it would produce algorithms  specialized in complex low-dimensional functions, albeit with limited  scalability.
In the previous sections, we identified two trends associated with the use of different training instances.
The first trend, observed in the aggregated data for ALL functions in \Tbls~\ref{table:MED-MEDerr-MAD-dftVStuned-50} and \ref{table:MED-MEDerr-MAD-mtfVSHyb-100-50}, shows that, on average, algorithms trained on \LDO achieve superior MED, MEDerr and MAD values compared to those trained on \TFOUT.
The second trend, which is the opposite of the first, applies specifically to the results for the HYB and HCP functions, where  algorithms trained on \TFOUT consistently outperform those trained on \LDO.

To determine whether the performance differences in the HYB functions increases with the number of dimension---suggesting a connection between the second trend and the fact that \TFOUT includes dimensional functions---we compare the distance in Av.Ranking obtained by each pair of \TFOUT-\LDO algorithms in the HYB functions.
The aggregated values obtained for all pairs were $0.88$, in the functions with $d=50$, and $0.92$, in the functions with $d=100$. 
The fact that these two values are similar indicates that the differences in performance between the algorithms trained on \TFOUT and those trained on \LDO remain constant despite the increase in the number of dimensions.
Based on the fact that (i) the main difference between \TFOUT and \LDO is that \TFOUT includes functions with large dimensionality and (ii) each pair of \TFOUT-\LDO algorithms performs similarly well independently of the dimensionality of the functions, we conclude that the set of components that are useful for solving LD functions are also useful for solving HYB and HCP functions.
In the following analysis of the algorithm's performance on LD functions, we provide further evidence supporting this thesis.

\paragraph{Generalization of the algorithms trained on \TFOUT and \LDO to \fCECFOURTEEN}

To measure the ability of the algorithms to generalize to other types of functions, we compare their performance on the 22 functions of the \fCECFOURTEEN with $d=\{50, 100\}$.
The \fCECFOURTEEN has 22 functions, of which 3 are UNI, 13 are MUL, 6 are HYB, and 20 are ROT (see \Tbl~\ref{table:instance distrubution per scenarios}).
As indicated in \Tbl~\ref{table:MED-MEDerr-MAD-cec14-50}, based on the MED and MEDerr metrics, the top-performing algorithms across all functions in \fCECFOURTEEN with $d=50$ are \MTFldo and \PCHybldo (tied for first place), followed by \PCHybtfo and \CMAESldo.
Although both \MTFldo and \PSOldo ranked first, we found statistically significant differences in favor of \MTFldo in five cases (\DEtfo, \DEldo, \PSOtfo, \PSOldo and \PDHybtfo), while in the case of \PCHybldo the statistical test showed differences in only two cases, \DEtfo and \PSOtfo. 
Based on the data per function class in \Tbl~\ref{table:MED-MEDerr-MAD-cec14-50}, \MTFldo performs quite well in the MUL, HYB and ROT functions, but is outperformed by \CMAESldo and the two \PCHyb in the UNI functions.
In the case of \PCHyb, its performance is particularly good in the ROT functions, but again poor in the HYB functions, where it is the worst ranked algorithm.

\input{tables/tbl_cec14_1}

Regarding the robustness of the algorithms, measured by their MAD value, the results are consistent with those obtained by the algorithms on the MED and MEDerr metrics. 
In particular, \PCHybldo showed greater robustness in the UNI, ROT and ALL functions, while \MTFldo showed greater robustness in the MUL functions. 
It should be noted that although \fCECFOURTEEN was reserved for testing only, the functions $f_{1,4-10,15,16} \in$ \fCECFOURTEEN are the same as the functions $f_{3,6-14} \in $\fCECFIVE.
This means that 7 of the 88 functions in the \TFOUT training set are also present in the \fCECFOURTEEN test set, with 6 functions belonging to the MUL class and 1 function belonging to the UNI class.
In the cases of the training set \LDO and the test set \fCECFOURTEEN, there is no overlap because \LDO excludes dimensions 50 and 100, and these are the dimensions we consider for the analysis using \fCECFOURTEEN.
Despite the overlap between \TFOUT and \fCECFOURTEEN, we found no evidence that the algorithms trained on \TFOUT had an advantage on the MUL functions over those trained on \LDO.

\input{tables/tbl_cec14_2}

Now we focus on the results in \fCECFOURTEEN with $d=100$ (\Tbl~\ref{table:MED-MEDerr-MAD-cec14-100}).
Based on the MED and MEDerr metrics, the best performing algorithm is \CMAESldo, followed by \PCHybtfo and \PCHybldo (both tied for second place), and finally \DCHybldo.
The analysis by function class shows that \CMAESldo is the best algorithm for the UNI, MUL, and ROT functions, and fourth for the HYB functions, where it is outperformed by the two \MTF hybrids and \CMAEStfo. 
The performance of \CMAESldo is clearly superior to most of the compared algorithms based on MED and MEDerr, but has a higher variability in its results.
According to the MAD metric, the most robust algorithm is \PCHybldo, followed by \PCHybtfo, and finally \DCHybtfo, while \CMAESldo ranks seventh.

After comparing the results obtained by the algorithms on the \fBENCHMARK (\Tbls~\ref{table:MED-MEDerr-MAD-dftVStuned-50} and \ref{table:MED-MEDerr-MAD-mtfVSHyb-100-50}) and \fCECFOURTEEN (\Tbls~\ref{table:MED-MEDerr-MAD-cec14-50} and \ref{table:MED-MEDerr-MAD-cec14-100}) with $d=50$ and $d=100$, we found consistency in their rankings for median solution quality (MED), median solution error (MEDerr), and median absolute deviation (MAD) on both dimensions.
In the two benchmarks, the best results over all function sets are obtained by \MTF, \PCHyb, and \CMAES, and the worst results are obtained by \PSO, \DE, and \PDHyb.
Also, for both \fBENCHMARK and \fCECFOURTEEN, we found that (i) on average, the algorithms trained on \LDO outperform those trained on \TFOUT, and (ii) for the HYB functions, the algorithms trained on \TFOUT outperform those trained on \LDO. 
These results suggest that both training scenarios produce algorithms with similar generalization capabilities when it comes to functions with features such as multimodality, mathematical transformations, and function hybridization.

\paragraph{Generalization of the algorithms trained on \TFOUT and \LDO to \fSOCO}

So far, we have analyzed the performance of the algorithms only on functions with low to medium dimensionality. 
The goal of this final analysis is to compare the performance of the algorithms using functions with a much larger scale, namely $750$ and $1\,250$ dimensions.
For this purpose, we performed experiments on \fSOCO, which consists of 19 freely scalable functions, of which 7 are shifted unimodal (SHU) functions, 4 are MUL functions, and 8 are HYB functions.

\input{tables/tbl_LD_3}

In \Tbl~\ref{table:MED-MEDerr-MAD-LD-750} we report the results of the algorithms on \fSOCO with dimension $750$. 
Based on MED and MEDerr performance metrics, the best performing algorithms across all functions are 
\PCHybtfo, followed by \DCHybtfo and \PCHybldo.
The analysis of results per function class shows that \PCHybldo and \DCHybtfo are tied for the first place for the SHU functions, \PCHybtfo is ranked first in the MUL functions and \MTFldo is ranked first in the HYB functions.
It is also interesting to note that the trend observed earlier, where the algorithms trained on \LDO consistently outperform those trained on \TFOUT across all functions, is partially reversed in this set of benchmark functions.
In fact, with the exception of \MTF and \PDHybs, the algorithms trained on \TFOUT obtained better MED and MEDerr values than those trained on \LDO; and this effect is particularly notable for \DE.

Although the statistical test did not find a significant difference in the performance of the algorithms based on the MED and MEDerr values (with the exception of \PSOldo), the number of "wins" of the algorithm gives us an indication of how competitive they are.
The number of "wins" refers to the number of times the algorithm obtained the lowest MED, MEDerr or MAD value.
As shown in \Tbl~\ref{table:MED-MEDerr-MAD-LD-750}, \PCHybtfo obtained the best value 8 times, \DEtfo 6 times, the two \CMAES and \PCHybtfo 2 times and \PSOtfo and \DCHybldo 1 time each.
The number of "wins", while useful for competitive testing, is not necessarily a reliable performance metric on its own due to its sensitivity to the precision of algorithms in objective functions with mathematical transformations.
This means that if no precision is defined, only the algorithm that comes closest to the optimum will be awarded a "win", even if the differences in solution with other algorithms are negligible. 

Another noteworthy finding from the experiments on \fSOCO with LD is the difference in performance between \DEtfo and \DEldo.
While \DEtfo is competitive with the best-ranked algorithms on the MUL, HYB and ALL functions, ranking either third or fourth, \DEldo is among the worst-ranked algorithms on all function classes.
However, the two \DE implementations differ only in a few design choices.
Both \DE implementations use the \VBnatural and \DEBVbest algorithm components and have similar values for the crossover rate ($p_a$) and the scaling factor ($\beta$), although these are slightly higher in the case of \DEtfo (see \Tbl~\ref{table:ParameterSettingsALL}).
The main differences between \DEtfo and \DEldo are the algorithm components \Pop and \DEVectDifferences.
In the case of \DEtfo, the population is handled by \PopTV, which is updated every 26 iterations and varies between 39 and 132 solutions.
Since the percentage of \DEVectDifferences in \DEtfo is 0.0151, the algorithm adds only one vector difference when the population size is less than 100, and two vector differences when it is greater than or equal to 100.
Compared to \DEtfo, the implementation of \DEldo uses a larger \DEVectDifferences value (0.1711) and a smaller population size, starting with 46 solutions and ending with 69 solutions.
Therefore, from the fifth iteration, when \DEtfo reaches its maximum population size, each individual in the population will add 12 vector differences to compute its trail vector.

Based on the results of \DEtfo and \DEldo on \fBENCHMARK (\Tbls~\ref{table:MED-MEDerr-MAD-dftVStuned-50} to \ref{table:MED-MEDerr-MAD-mtfVSHyb-100-50}) and on \fCECFOURTEEN (\Tbls~\ref{table:MED-MEDerr-MAD-cec14-50} to \ref{table:MED-MEDerr-MAD-cec14-100}) with $d=\{50, 100\}$, where both implementations gave similar results, with \DEldo slightly better than \DEtfo on average, the performance of \DE does not seem to be particularly sensitive to the options used for the \Pop and \DEVectDifferences components.
However, in large dimensional search spaces, as in the case of \fSOCO with $d=750$, the \Pop and \DEVectDifferences components have a strong impact on the performance of \DE.
In practice, on complex multimodal search spaces with less than 100 dimensions, a good strategy is to implement \DE using a population of about 75 individuals and adding a maximum of 15\% of the population as vector differences.
However, when the number of dimensions is greater than 100, the performance of \DE can be improved by slightly increasing the population size and reducing the number of vector differences to one or at most two.

Finally, in \Tbl~\ref{table:MED-MEDerr-MAD-LD-1250}, we show the results obtained by the algorithms on \fSOCO with $1\,250$ dimensions.
The best-ranked algorithms across the entire set of functions, based on the MED and MEDerr performance metrics, are \DCHybtfo, first place, \PCHybtfo, second place, and \MTFldo, third place.
Similarly to \fSOCO with $750$ dimensions, \DCHybtfo and \PCHybtfo obtained the best results overall and \PCHybtfo also in the MUL functions; however, in the SHU and HYB functions, \MTFtfo was clear winner, outperforming the first and second place by a significant margin.
In terms of scoring the higher number of "wins" for the MED and MEDerr metrics, the best performing algorithm is \PCHybtfo, 8 times, followed by \DCHybtfo, 4 times, the two \MTF, 3 times, \CMAEStfo and \DEtfo, 2 times, and finally \CMAESldo, \PSOtfo and \PCHybldo, 1 times each.
We found a statistical significant difference in favor of \PCHybtfo with respect to \PDHybtfo, \PDHybldo and \DCHybldo, in the MED and MEDerr results. 
The results obtained by the algorithms on the MAD metric indicate that \MTF, \CMAES, \DCHyb, \PCHyb, \DEldo and, \PSOldo exhibit comparable robustness, with \CMAESldo and \PCHybldo outperforming the others. In contrast, \DEtfo, \PSOtfo and \PDHyb display variability in their outcomes.

\input{tables/tbl_LD_4}

Based on the results shown in \Tbls~\ref{table:MED-MEDerr-MAD-LD-750} and \ref{table:MED-MEDerr-MAD-LD-1250}, we observe strong differences in the generalization capabilities of the algorithms when transitioning from low-dimensional to high-dimensional search spaces. 
In particular, \CMAESldo and \MTFldo, which obtained the best results on functions with dimensions $d=\{50, 100\}$, were outperformed by \DCHybtfo and \PCHybtfo on functions with dimensions $d=\{750, 1250\}$.
The implementations with better generalization in terms of scalability are \PCHybldo and \PCHybtfo, which perform similarly on average on \fBENCHMARK, \fCECFOURTEEN, and \fSOCO irrespective of problem dimensionality. However, they consistently underperformed on HYB functions across all cases.
Finally, we found that the algorithms trained on \TFOUT performed better than those trained on \LDO on both HYB and LD functions, suggesting that the algorithm components useful for tackling HYB functions are also useful for tackling LD functions.

\section{Conclusions}\label{sec:conclusions}
In this article, we have proposed \MetafoR, a metaheuristic software framework for \algInPaperLS applicable to single-objective \cops.
The main advantage of \MetafoR is that it enables the creation of high-performance \mh implementations using an \ad approach, i.e., the selection of algorithm components and parameter settings in the implementation is done using an \aact, such as \irace, instead of relying on the expertise and intuition of a human algorithm designer.
To develop \MetafoR, we considered a component-based architecture and four main modules (\MetafoRModules), resulting in a total of 104 parameters, each representing either an algorithm component or a numerical parameter related to the use of a component.
\MetafoR allows three different types of hybridization, \ExecutionCB, \ExecutionPB and \ExecutionMP, which were selected based on a brief literature review of prominent hybrids of \algInPaper.
The three main \mh approaches included in \MetafoR (\algInPaper) can be used alone or in combination, allowing users not only to replicate hundreds of single-approach and hybrid algorithms already proposed in the literature, but also to generate many new implementations never proposed before.

We have shown experimentally that \MetafoR can be used to create implementations that outperform the state of the art on various metrics, as well as to create implementations that are tailored to specific problem classes. 
We compared a total of 17 algorithms and studied two popular "machine-learning"-type instance separation strategies often used in \ad, namely the creation of a stratified training set using a fixed percentage (\TFOUT) and leave-one-class-out cross-validation (\LDO). 
Although neither instance separation strategy can be considered better than the other, we found that \TFOUT works well for HYB and LD functions, while \LDO is a good strategy for low-dimensional functions with smoother search spaces, such as UNI and MUL functions. 
Also, based on the design of the best performing algorithms in our experimental study, we have identified strategies that can be used to guide the design of hybrid \mhs. 

Of course, our work also has limitations and could be improved in a number of ways. 
One limitation is the fact that the number of components available in the literature is enormous, and it is impossible to imagine including all of them in \MetafoR.
Another limitation is the type of hybridization currently available in \MetafoR. In particular, to keep things manageable, we have omitted some of the most complex types of hybridization and a number of components that can potentially provide better results. 
For future improvements to our work, we are considering the following: (i) incorporating other \mh and \ls approaches, such as \abc \cite{AydYavOzyYasStu2017} and Powell's BOBYQA and NEWUOA \cite{Powell2002,Powell2006}; (ii) exploring hybridization with ML techniques, including deep learning and reinforcement learning hybrids; and (iii) integrating components from other metaheuristic software frameworks that have demonstrated their utility in experimental studies \cite{KosVerKor2024:ev}.

\begin{acks}
Christian L. Camacho-Villalón acknowledges support from the SMASH postdoctoral fellowship
program of Slovenia from which he is a MSCA fellow.
Marco Dorigo and Thomas St\"utzle acknowledge support from the Belgian F.R.S.-FNRS, of
which they are Research Directors. 
\end{acks}


\providecommand{\MaxMinAntSystem}{{$\cal MAX$--$\cal MIN$} {A}nt {S}ystem} \providecommand{\Rpackage}[1]{#1} \providecommand{\SoftwarePackage}[1]{#1} \providecommand{\proglang}[1]{#1}





\end{document}

%% file: tables/tbl_dftVstuned.tex
\begin{ThreePartTable} 
\setlength{\tabcolsep}{0.5pt} 

        {\fontsize{6.5}{6.2}\selectfont
        \begin{longtable}{m{0.1cm} C{1.25cm} m{0.88cm} m{0.88cm} m{0.88cm} m{0.88cm} m{0.88cm} m{0.88cm} m{0.88cm} m{0.88cm} m{0.88cm} m{0.88cm} m{0.88cm} m{0.88cm}m{0.88cm} m{0.88cm} m{0.88cm}} 
        \caption{Average median (Av.MED), average median error (Av.MEDerr) and average median absolute deviation (Av.MAD) obtained by the algorithms on functions $f_{1-50}$ with $\mathbf{d = 50}$.} \label{table:MED-MEDerr-MAD-dftVStuned-50}\\
        \toprule
        & $f\#$ &  
        \textbf{\CMAESdftTBL} & \textbf{\CMAEStfoTBL} & \textbf{\CMAESldoTBL} & 
        \textbf{\DEdftTBL} & \textbf{\DEtfoTBL} & \textbf{\DEldoTBL} & 
        \textbf{\StdPSOdftTBL} & \textbf{\PSOtfoTBL} & \textbf{\PSOldoTBL} & 
        \textbf{\PDHybtfoTBL} & \textbf{\PDHybldoTBL} & 
        \textbf{\DCHybtfoTBL} & \textbf{\DCHybldoTBL} &
        \textbf{\PCHybtfoTBL} & \textbf{\PCHybldoTBL} \\
        \midrule
        \endfirsthead
        
        \caption*{ Table \ref{table:MED-MEDerr-MAD-dftVStuned-50} (Continued.)}\\
        \toprule
        & $f\#$ & 
        \textbf{\CMAESdftTBL} & \textbf{\CMAEStfoTBL} & \textbf{\CMAESldoTBL} & 
        \textbf{\DEdftTBL} & \textbf{\DEtfoTBL} & \textbf{\DEldoTBL} & 
        \textbf{\StdPSOdftTBL} & \textbf{\PSOtfoTBL} & \textbf{\PSOldoTBL} & 
        \textbf{\PDHybtfoTBL} & \textbf{\PDHybldoTBL} & 
        \textbf{\DCHybtfoTBL} & \textbf{\DCHybldoTBL} &
        \textbf{\PCHybtfoTBL} & \textbf{\PCHybldoTBL} \\
        \midrule
        \endhead
        
        \bottomrule
        \endfoot
        
        \bottomrule
        \endlastfoot

& {Wins \newline (MED) } & \multicolumn{1}{c}{13} & \multicolumn{1}{c}{7} & \multicolumn{1}{c}{6} & \multicolumn{1}{c}{0} & \multicolumn{1}{c}{11} & \multicolumn{1}{c}{10} & \multicolumn{1}{c}{0} & \multicolumn{1}{c}{2} & \multicolumn{1}{c}{3} & \multicolumn{1}{c}{11} & \multicolumn{1}{c}{18} & \multicolumn{1}{c}{6} & \multicolumn{1}{c}{8} & \multicolumn{1}{c}{12} & \multicolumn{1}{c}{9} \\
\midrule
\multirow{2}{*}{\rotatebox[origin=c]{90}{UNI}} & Av.MED  & \textbf{0.00e+00} & 2.26e+00 & 9.08e+00 & 1.83e+09 & 2.91e+06 & 3.13e+06 & 8.44e+08 & 1.06e+05 & 2.10e+05 & 1.71e+06 & 2.18e+06 & 1.72e+01 & 5.60e+02 & 1.42e+02 & \textbf{0.00e+00} \\
& Av.Ranking  & 7.75 & 5.58 & \textbf{3.83} & 14.92 & 9.08 & 8.75 & 13.33 & 10.33 & 10.42 & 6.75 & 8.58 & 6.83 & 5.33 & 6.42 & 4.67 \\
\midrule
\multirow{2}{*}{\rotatebox[origin=c]{90}{MUL}} & Av.MED  & 3.15e+03 & 3.19e+03 & 3.13e+03 & 6.67e+08 & 6.08e+03 & 5.86e+03 & 2.63e+08 & 4.03e+03 & 3.69e+03 & 3.29e+03 & 3.36e+03 & 3.05e+03 & 3.12e+03 & 3.06e+03 & \textbf{3.02e+03} \\
& Av.Ranking  & 7.31 & 7.58 & 5.96 & 14.92 & 9.23 & 8.08 & 13.00 & 11.92 & 10.15 & 6.81 & 6.12 & 6.23 & 6.35 & 6.12 & \textbf{5.77} \\
\midrule
\multirow{2}{*}{\rotatebox[origin=c]{90}{HYB}} & Av.MED  & 8.31e+02 & 2.40e+03 & 8.11e+02 & 6.19e+08 & 1.43e+06 & 6.64e+05 & 1.09e+08 & 1.54e+04 & 1.55e+04 & 2.77e+04 & 2.49e+04 & 8.30e+02 & 8.11e+02 & \textbf{8.05e+02} & 8.30e+02 \\
& Av.Ranking  & \textbf{4.30} & 4.60 & 5.00 & 10.18 & 7.53 & 7.90 & 11.03 & 10.72 & 9.47 & 8.40 & 8.65 & 8.10 & 9.15 & 9.07 & 9.40 \\
\midrule
\multirow{2}{*}{\rotatebox[origin=c]{90}{HCP}} & Av.MED  & 5.09e+02 & 5.57e+02 & 5.62e+02 & 1.07e+03 & 6.24e+02 & 6.46e+02 & 8.54e+02 & 7.75e+02 & 6.86e+02 & 6.62e+02 & 5.85e+02 & \textbf{4.62e+02} & 4.71e+02 & 4.63e+02 & 5.50e+02 \\
& Av.Ranking  & \textbf{1.78} & 2.66 & 3.78 & 5.88 & 6.06 & 7.12 & 8.44 & 8.92 & 9.14 & 9.82 & 10.60 & 10.62 & 11.76 & 12.36 & 13.12 \\
\midrule
\multirow{2}{*}{\rotatebox[origin=c]{90}{ROT}} & Av.MED  & 1.34e+03 & 1.39e+03 & 1.38e+03 & 5.44e+08 & 1.58e+06 & 1.71e+06 & 4.60e+08 & 5.87e+04 & 1.16e+05 & 9.34e+05 & 1.19e+06 & \textbf{1.29e+03} & 1.33e+03 & 1.30e+03 & 1.32e+03 \\
& Av.Ranking  & 5.55 & 6.05 & 5.64 & 14.27 & 10.82 & 11.32 & 13.41 & 11.09 & 9.41 & 8.82 & 9.55 & 5.59 & 5.59 & 5.36 & \textbf{5.27} \\
\midrule
\multirow{2}{*}{\rotatebox[origin=c]{90}{ALL}} & Av.MED  & 1.30e+03 & 1.80e+03 & 1.31e+03 & 8.25e+08 & 1.13e+06 & 9.52e+05 & 3.14e+08 & 3.15e+04 & 5.63e+04 & 4.20e+05 & 5.32e+05 & \textbf{1.27e+03} & 1.42e+03 & 1.30e+03 & 1.28e+03 \\
& Av.Ranking  & 6.84 & 6.48 & 5.70 & 14.62 & 9.52 & 9.04 & 13.36 & 11.70 & 10.00 & 7.18 & 7.42 & 6.08 & 6.16 & 6.00 & \textbf{5.48} \\
\midrule
& \multicolumn{1}{c}{Wilcoxon test} & \multicolumn{1}{c}{$\approx$} & \multicolumn{1}{c}{$\approx$} & \multicolumn{1}{c}{$\approx$} & \multicolumn{1}{c}{$+$} & \multicolumn{1}{c}{$+$} & \multicolumn{1}{c}{$\approx$} & \multicolumn{1}{c}{$+$} & \multicolumn{1}{c}{$+$} & \multicolumn{1}{c}{$+$} & \multicolumn{1}{c}{$+$} & \multicolumn{1}{c}{$\approx$} & \multicolumn{1}{c}{$\approx$} & \multicolumn{1}{c}{$\approx$} & \multicolumn{1}{c}{$\approx$} & \multicolumn{1}{c}{} \\

\toprule
& {Wins \newline (MEDerr) } & \multicolumn{1}{c}{13} & \multicolumn{1}{c}{7} & \multicolumn{1}{c}{6} & \multicolumn{1}{c}{0} & \multicolumn{1}{c}{11} & \multicolumn{1}{c}{10} & \multicolumn{1}{c}{0} & \multicolumn{1}{c}{2} & \multicolumn{1}{c}{3} & \multicolumn{1}{c}{11} & \multicolumn{1}{c}{18} & \multicolumn{1}{c}{6} & \multicolumn{1}{c}{8} & \multicolumn{1}{c}{12} & \multicolumn{1}{c}{9} \\
\midrule
\multirow{2}{*}{\rotatebox[origin=c]{90}{UNI}} & Av.MEDerr  & \textbf{0.00e+00} & 2.26e+00 & 9.08e+00 & 1.83e+09 & 2.91e+06 & 3.13e+06 & 8.44e+08 & 1.06e+05 & 2.10e+05 & 1.71e+06 & 2.18e+06 & 1.72e+01 & 5.60e+02 & 1.42e+02 & \textbf{0.00e+00} \\
& Av.Ranking  & 7.75 & 5.58 & \textbf{3.83} & 14.92 & 9.08 & 8.75 & 13.33 & 10.33 & 10.42 & 6.75 & 8.58 & 6.83 & 5.33 & 6.42 & 4.67 \\
\midrule
\multirow{2}{*}{\rotatebox[origin=c]{90}{MUL}} & Av.MEDerr  & 2.17e+02 & 2.59e+02 & 1.99e+02 & 6.67e+08 & 3.15e+03 & 2.93e+03 & 2.63e+08 & 1.10e+03 & 7.59e+02 & 3.60e+02 & 4.31e+02 & 1.18e+02 & 1.92e+02 & 1.31e+02 & \textbf{9.13e+01} \\
& Av.Ranking  & 7.31 & 7.58 & 5.96 & 14.92 & 9.23 & 8.08 & 13.00 & 11.92 & 10.15 & 6.81 & 6.12 & 6.23 & 6.35 & 6.12 & \textbf{5.77} \\
\midrule
\multirow{2}{*}{\rotatebox[origin=c]{90}{HYB}} & Av.MEDerr  & 5.66e+01 & 1.63e+03 & 3.72e+01 & 6.19e+08 & 1.43e+06 & 6.63e+05 & 1.09e+08 & 1.46e+04 & 1.48e+04 & 2.69e+04 & 2.41e+04 & 5.61e+01 & 3.68e+01 & \textbf{3.10e+01} & 5.53e+01 \\
& Av.Ranking  & \textbf{4.30} & 4.60 & 5.00 & 10.18 & 7.53 & 7.90 & 11.03 & 10.72 & 9.47 & 8.40 & 8.65 & 8.10 & 9.15 & 9.07 & 9.40 \\
\midrule
\multirow{2}{*}{\rotatebox[origin=c]{90}{HCP}} & Av.MEDerr  & 8.61e+01 & 1.34e+02 & 1.39e+02 & 6.50e+02 & 2.01e+02 & 2.24e+02 & 4.31e+02 & 3.53e+02 & 2.63e+02 & 2.39e+02 & 1.62e+02 & \textbf{3.88e+01} & 4.79e+01 & 4.03e+01 & 1.28e+02 \\
& Av.Ranking  & \textbf{1.78} & 2.66 & 3.78 & 5.88 & 6.06 & 7.12 & 8.44 & 8.92 & 9.14 & 9.82 & 10.60 & 10.62 & 11.76 & 12.36 & 13.12 \\
\midrule
\multirow{2}{*}{\rotatebox[origin=c]{90}{ROT}} & Av.MEDerr  & 1.11e+02 & 1.60e+02 & 1.47e+02 & 5.44e+08 & 1.58e+06 & 1.71e+06 & 4.60e+08 & 5.75e+04 & 1.14e+05 & 9.32e+05 & 1.19e+06 & \textbf{6.49e+01} & 9.70e+01 & 7.47e+01 & 9.56e+01 \\
& Av.Ranking  & 5.55 & 6.05 & 5.64 & 14.27 & 10.82 & 11.32 & 13.41 & 11.09 & 9.41 & 8.82 & 9.55 & 5.59 & 5.59 & 5.36 & \textbf{5.27} \\
\midrule
\multirow{2}{*}{\rotatebox[origin=c]{90}{ALL}} & Av.MEDerr  & 1.01e+02 & 5.97e+02 & 1.04e+02 & 8.25e+08 & 1.13e+06 & 9.51e+05 & 3.14e+08 & 3.03e+04 & 5.51e+04 & 4.18e+05 & 5.31e+05 & \textbf{6.49e+01} & 2.14e+02 & 9.18e+01 & 7.20e+01 \\
& Av.Ranking  & 6.84 & 6.48 & 5.70 & 14.62 & 9.52 & 9.04 & 13.36 & 11.70 & 10.00 & 7.18 & 7.42 & 6.08 & 6.16 & 6.00 & \textbf{5.48} \\
\midrule
& \multicolumn{1}{c}{Wilcoxon test} & \multicolumn{1}{c}{$\approx$} & \multicolumn{1}{c}{$\approx$} & \multicolumn{1}{c}{$\approx$} & \multicolumn{1}{c}{$+$} & \multicolumn{1}{c}{$+$} & \multicolumn{1}{c}{$\approx$} & \multicolumn{1}{c}{$+$} & \multicolumn{1}{c}{$+$} & \multicolumn{1}{c}{$+$} & \multicolumn{1}{c}{$+$} & \multicolumn{1}{c}{$\approx$} & \multicolumn{1}{c}{$\approx$} & \multicolumn{1}{c}{$\approx$} & \multicolumn{1}{c}{$\approx$} & \multicolumn{1}{c}{} \\

\toprule
& {Wins \newline (MAD) } & \multicolumn{1}{c}{5} & \multicolumn{1}{c}{7} & \multicolumn{1}{c}{7} & \multicolumn{1}{c}{3} & \multicolumn{1}{c}{9} & \multicolumn{1}{c}{9} & \multicolumn{1}{c}{1} & \multicolumn{1}{c}{2} & \multicolumn{1}{c}{1} & \multicolumn{1}{c}{9} & \multicolumn{1}{c}{16} & \multicolumn{1}{c}{6} & \multicolumn{1}{c}{10} & \multicolumn{1}{c}{10} & \multicolumn{1}{c}{12} \\
\midrule
\multirow{2}{*}{\rotatebox[origin=c]{90}{UNI}} & Av.MAD  & \textbf{0.00e+00} & 1.98e+00 & 6.80e+02 & 2.97e+07 & 2.59e+06 & 9.11e+05 & 3.97e+06 & 1.92e+04 & 1.77e+04 & 3.64e+05 & 2.26e+05 & 1.72e+01 & 1.87e+02 & 1.42e+02 & \textbf{0.00e+00} \\
& Av.Ranking  & 7.67 & 7.25 & 5.42 & 12.58 & 9.17 & 8.92 & 12.92 & 10.50 & 11.08 & 7.00 & 8.33 & 6.58 & 5.08 & 5.67 & \textbf{4.92} \\
\midrule
\multirow{2}{*}{\rotatebox[origin=c]{90}{MUL}} & Av.MAD  & 4.01e+01 & 4.19e+01 & 4.11e+01 & 7.70e+02 & 2.35e+02 & 2.01e+02 & 1.25e+08 & 5.52e+01 & 5.86e+01 & 3.14e+01 & 3.68e+01 & 3.38e+01 & 4.96e+01 & \textbf{2.60e+01} & 4.14e+01 \\
& Av.Ranking  & 8.15 & 7.04 & 6.85 & 12.81 & 8.88 & 7.92 & 12.96 & 10.69 & 9.88 & 7.81 & 6.27 & 5.92 & 6.96 & \textbf{5.19} & 5.73 \\
\midrule
\multirow{2}{*}{\rotatebox[origin=c]{90}{HYB}} & Av.MAD  & 1.43e+01 & 1.82e+02 & 4.09e+01 & 2.34e+08 & 1.33e+06 & 2.96e+05 & 1.29e+07 & 5.51e+03 & 3.04e+03 & 2.59e+03 & 4.06e+03 & 2.24e+01 & 1.59e+01 & \textbf{3.73e+00} & 2.31e+01 \\
& Av.Ranking  & 5.38 & \textbf{4.75} & 5.72 & 9.15 & 7.03 & 6.80 & 10.93 & 10.28 & 9.45 & 9.10 & 8.07 & 7.85 & 9.12 & 8.80 & 9.12 \\
\midrule
\multirow{2}{*}{\rotatebox[origin=c]{90}{HCP}} & Av.MAD  & 8.05e+01 & 3.92e+01 & 7.50e+01 & 5.58e+01 & 2.02e+01 & 3.08e+01 & 9.60e+01 & 8.14e+01 & 7.07e+01 & 1.18e+01 & 2.48e+01 & \textbf{3.13e+00} & 7.80e+00 & 2.05e+01 & 4.71e+01 \\
& Av.Ranking  & \textbf{2.28} & 2.96 & 3.92 & 5.36 & 5.78 & 6.52 & 8.30 & 8.94 & 9.06 & 9.90 & 10.06 & 10.62 & 11.82 & 12.36 & 13.08 \\
\midrule
\multirow{2}{*}{\rotatebox[origin=c]{90}{ROT}} & Av.MAD  & 5.31e+01 & 3.16e+01 & 5.01e+01 & 1.62e+07 & 1.41e+06 & 4.97e+05 & 2.16e+06 & 1.03e+04 & 9.55e+03 & 1.98e+05 & 1.23e+05 & 1.86e+01 & 2.13e+01 & \textbf{1.71e+01} & 4.00e+01 \\
& Av.Ranking  & 7.18 & 6.59 & 6.55 & 11.55 & 10.23 & 9.59 & 12.55 & 10.82 & 9.27 & 9.45 & 7.55 & 5.91 & 6.32 & 5.45 & \textbf{4.86} \\
\midrule
\multirow{2}{*}{\rotatebox[origin=c]{90}{ALL}} & Av.MAD  & 3.40e+01 & 7.62e+01 & 2.04e+02 & 7.74e+07 & 1.02e+06 & 3.08e+05 & 4.22e+07 & 6.28e+03 & 5.19e+03 & 8.81e+04 & 5.54e+04 & \textbf{2.17e+01} & 6.61e+01 & 4.76e+01 & 2.97e+01 \\
& Av.Ranking  & 7.98 & 6.96 & 6.72 & 12.44 & 9.06 & 8.26 & 13.10 & 11.00 & 10.14 & 7.88 & 6.76 & 5.80 & 6.40 & 5.50 & \textbf{5.32} \\
\midrule
& \multicolumn{1}{c}{Wilcoxon test} & \multicolumn{1}{c}{$\approx$} & \multicolumn{1}{c}{$\approx$} & \multicolumn{1}{c}{$\approx$} & \multicolumn{1}{c}{$+$} & \multicolumn{1}{c}{$+$} & \multicolumn{1}{c}{$\approx$} & \multicolumn{1}{c}{$+$} & \multicolumn{1}{c}{$+$} & \multicolumn{1}{c}{$+$} & \multicolumn{1}{c}{$\approx$} & \multicolumn{1}{c}{$\approx$} & \multicolumn{1}{c}{$\approx$} & \multicolumn{1}{c}{$\approx$} & \multicolumn{1}{c}{$\approx$} & \multicolumn{1}{c}{} \\
    \end{longtable} }
\end{ThreePartTable}

%% file: tables/tbl_mtfVSHyb.tex
\begin{ThreePartTable} 
\setlength{\tabcolsep}{0.5pt} 

        {\fontsize{6.5}{6.2}\selectfont
        \begin{longtable}{m{0.1cm} C{1.25cm} m{0.88cm} m{0.88cm} m{0.88cm} m{0.88cm} m{0.88cm} m{0.88cm} m{0.88cm} m{0.88cm} m{0.88cm} m{0.88cm} m{0.88cm} m{0.88cm} m{0.88cm} m{0.88cm}} 
        \caption{Average median (Av.MED), average median error (Av.MEDerr) and average median absolute deviation (Av.MAD) obtained by the algorithms on functions $f_{1-40}$ with $\mathbf{d = 100}$ and $f_{41-50}$ (HCP) with $\mathbf{d = 50}$.} \label{table:MED-MEDerr-MAD-mtfVSHyb-100-50}\\
        \toprule
        & $f\#$ &  
        \textbf{\MTFtfoTBL} & \textbf{\MTFldoTBL} & 
        \textbf{\CMAEStfoTBL} & \textbf{\CMAESldoTBL} & 
        \textbf{\DEtfoTBL} & \textbf{\DEldoTBL} & 
        \textbf{\PSOtfoTBL} & \textbf{\PSOldoTBL} & 
        \textbf{\PDHybtfoTBL} & \textbf{\PDHybldoTBL} & 
        \textbf{\DCHybtfoTBL} & \textbf{\DCHybldoTBL} &
        \textbf{\PCHybtfoTBL} & \textbf{\PCHybldoTBL} \\
        \midrule
        \endfirsthead
        
        \caption*{ Table \ref{table:MED-MEDerr-MAD-mtfVSHyb-100-50} (Continued.)}\\
        \toprule
        & $f\#$ &  
        \textbf{\MTFtfoTBL} & \textbf{\MTFldoTBL} & 
        \textbf{\CMAEStfoTBL} & \textbf{\CMAESldoTBL} & 
        \textbf{\DEtfoTBL} & \textbf{\DEldoTBL} & 
        \textbf{\PSOtfoTBL} & \textbf{\PSOldoTBL} & 
        \textbf{\PDHybtfoTBL} & \textbf{\PDHybldoTBL} & 
        \textbf{\DCHybtfoTBL} & \textbf{\DCHybldoTBL} &
        \textbf{\PCHybtfoTBL} & \textbf{\PCHybldoTBL} \\
        \midrule
        \endhead
        
        \bottomrule
        \endfoot
        
        \bottomrule
        \endlastfoot

& {Wins \newline (MED) } & \multicolumn{1}{c}{4} & \multicolumn{1}{c}{11} & \multicolumn{1}{c}{4} & \multicolumn{1}{c}{9} & \multicolumn{1}{c}{8} & \multicolumn{1}{c}{7} & \multicolumn{1}{c}{0} & \multicolumn{1}{c}{1} & \multicolumn{1}{c}{7} & \multicolumn{1}{c}{12} & \multicolumn{1}{c}{3} & \multicolumn{1}{c}{7} & \multicolumn{1}{c}{9} & \multicolumn{1}{c}{5} \\
\midrule
\multirow{2}{*}{\rotatebox[origin=c]{90}{UNI}} & Av.MED  & 4.44e+03 & 2.37e+04 & 1.02e+04 & 3.45e+03 & 3.52e+07 & 2.49e+07 & 3.28e+05 & 7.80e+05 & 1.22e+07 & 1.16e+07 & 3.28e+03 & 3.68e+03 & \textbf{1.75e+03} & 2.42e+03 \\
& Av.Ranking  & 6.92 & 6.25 & 7.42 & \textbf{4.50} & 9.25 & 9.25 & 10.92 & 11.58 & 9.00 & 8.75 & 7.42 & 5.17 & 6.42 & 5.00 \\
\midrule
\multirow{2}{*}{\rotatebox[origin=c]{90}{MUL}} & Av.MED  & 6.30e+03 & \textbf{6.13e+03} & 6.58e+03 & 6.55e+03 & 1.11e+04 & 1.25e+04 & 9.84e+03 & 7.55e+03 & 7.49e+03 & 7.95e+03 & 6.19e+03 & 6.24e+03 & 6.23e+03 & 6.21e+03 \\
& Av.Ranking  & 7.62 & \textbf{5.15} & 8.65 & 6.54 & 9.65 & 7.96 & 12.23 & 10.81 & 7.81 & 6.65 & 6.85 & 6.69 & 6.69 & 5.92 \\
\midrule
\multirow{2}{*}{\rotatebox[origin=c]{90}{HYB}} & Av.MED  & 5.07e+03 & 1.31e+04 & 8.22e+03 & 4.34e+03 & 1.18e+07 & 6.15e+06 & 2.63e+04 & 3.32e+04 & 6.22e+04 & 7.20e+04 & 4.77e+03 & 3.75e+03 & \textbf{3.44e+03} & 4.48e+03 \\
& Av.Ranking  & 3.52 & \textbf{2.95} & 4.97 & 5.25 & 6.32 & 6.42 & 9.07 & 9.07 & 8.75 & 8.60 & 9.53 & 10.07 & 10.55 & 11.12 \\
\midrule
\multirow{2}{*}{\rotatebox[origin=c]{90}{HCP}} & Av.MED  & 5.54e+02 & \textbf{4.62e+02} & 5.57e+02 & 5.62e+02 & 6.24e+02 & 6.46e+02 & 7.75e+02 & 6.86e+02 & 6.62e+02 & 5.85e+02 & \textbf{4.62e+02} & 4.71e+02 & 4.63e+02 & 5.50e+02 \\
& Av.Ranking  & \textbf{2.40} & 2.56 & 3.56 & 4.72 & 6.18 & 7.24 & 8.24 & 8.50 & 9.24 & 9.92 & 9.90 & 11.14 & 11.70 & 12.42 \\\midrule
\multirow{2}{*}{\rotatebox[origin=c]{90}{ROT}} & Av.MED  & 5.73e+03 & 2.34e+04 & 1.32e+04 & 4.46e+03 & 3.25e+07 & 2.29e+07 & 2.96e+05 & 7.19e+05 & 1.12e+07 & 1.07e+07 & 5.48e+03 & 4.14e+03 & \textbf{4.11e+03} & 5.01e+03 \\
& Av.Ranking  & 6.08 & 5.92 & 6.92 & \textbf{4.46} & 12.00 & 11.62 & 10.08 & 9.77 & 9.31 & 10.31 & 6.00 & 4.92 & 5.62 & 4.77 \\
\midrule
\multirow{2}{*}{\rotatebox[origin=c]{90}{ALL}} & Av.MED  & 5.60e+03 & 1.43e+04 & 8.60e+03 & 5.12e+03 & 1.50e+07 & 9.76e+06 & 1.11e+05 & 2.49e+05 & 3.68e+06 & 3.51e+06 & 5.09e+03 & 4.85e+03 & \textbf{4.15e+03} & 4.74e+03 \\
& Av.Ranking  & 7.40 & \textbf{5.53} & 8.03 & 5.95 & 9.68 & 8.55 & 11.90 & 11.07 & 8.22 & 7.47 & 6.82 & 6.12 & 6.38 & 5.58 \\
\midrule
& \multicolumn{1}{c}{Wilcoxon test} & \multicolumn{1}{c}{$\approx$} & \multicolumn{1}{c}{} & \multicolumn{1}{c}{$\approx$} & \multicolumn{1}{c}{$\approx$} & \multicolumn{1}{c}{$+$} & \multicolumn{1}{c}{$+$} & \multicolumn{1}{c}{$+$} & \multicolumn{1}{c}{$+$} & \multicolumn{1}{c}{$+$} & \multicolumn{1}{c}{$+$} & \multicolumn{1}{c}{$\approx$} & \multicolumn{1}{c}{$\approx$} & \multicolumn{1}{c}{$\approx$} & \multicolumn{1}{c}{$\approx$} \\

\toprule
& {Wins \newline (MEDerr) } & \multicolumn{1}{c}{4} & \multicolumn{1}{c}{11} & \multicolumn{1}{c}{4} & \multicolumn{1}{c}{9} & \multicolumn{1}{c}{8} & \multicolumn{1}{c}{7} & \multicolumn{1}{c}{0} & \multicolumn{1}{c}{1} & \multicolumn{1}{c}{7} & \multicolumn{1}{c}{12} & \multicolumn{1}{c}{3} & \multicolumn{1}{c}{7} & \multicolumn{1}{c}{9} & \multicolumn{1}{c}{5} \\
\midrule
\multirow{2}{*}{\rotatebox[origin=c]{90}{UNI}} & Av.MEDerr  & 4.40e+03 & 2.37e+04 & 1.01e+04 & 3.41e+03 & 3.52e+07 & 2.49e+07 & 3.28e+05 & 7.80e+05 & 1.22e+07 & 1.16e+07 & 3.23e+03 & 3.64e+03 & \textbf{1.70e+03} & 2.38e+03 \\
& Av.Ranking  & 6.92 & 6.25 & 7.42 & \textbf{4.50} & 9.25 & 9.25 & 10.92 & 11.58 & 9.00 & 8.75 & 7.42 & 5.17 & 6.42 & 5.00 \\
\midrule
\multirow{2}{*}{\rotatebox[origin=c]{90}{MUL}} & Av.MEDerr  & 4.15e+02 & \textbf{2.41e+02} & 6.88e+02 & 6.59e+02 & 5.24e+03 & 6.57e+03 & 3.96e+03 & 1.66e+03 & 1.60e+03 & 2.06e+03 & 3.00e+02 & 3.55e+02 & 3.46e+02 & 3.21e+02 \\
& Av.Ranking  & 7.62 & \textbf{5.15} & 8.65 & 6.54 & 9.65 & 7.96 & 12.23 & 10.81 & 7.81 & 6.65 & 6.85 & 6.69 & 6.69 & 5.92 \\
\midrule
\multirow{2}{*}{\rotatebox[origin=c]{90}{HYB}} & Av.MEDerr  & 3.06e+03 & 1.11e+04 & 6.21e+03 & 2.33e+03 & 1.18e+07 & 6.15e+06 & 2.43e+04 & 3.12e+04 & 6.02e+04 & 7.00e+04 & 2.76e+03 & 1.74e+03 & \textbf{1.43e+03} & 2.47e+03 \\
& Av.Ranking  & 3.52 & \textbf{2.95} & 4.97 & 5.25 & 6.32 & 6.42 & 9.07 & 9.07 & 8.75 & 8.60 & 9.53 & 10.07 & 10.55 & 11.12 \\
\midrule
\multirow{2}{*}{\rotatebox[origin=c]{90}{HCP}} & Av.MEDerr  & 1.37e+02 & 4.61e+01 & 1.41e+02 & 1.46e+02 & 2.08e+02 & 2.30e+02 & 3.59e+02 & 2.69e+02 & 2.45e+02 & 1.68e+02 & \textbf{4.52e+01} & 5.43e+01 & 4.67e+01 & 1.34e+02 \\
& Av.Ranking  & \textbf{2.40} & 2.56 & 3.56 & 4.72 & 6.18 & 7.24 & 8.24 & 8.50 & 9.24 & 9.92 & 9.90 & 11.14 & 11.70 & 12.42 \\\midrule
\multirow{2}{*}{\rotatebox[origin=c]{90}{ROT}} & Av.MEDerr  & 2.23e+03 & 1.99e+04 & 9.67e+03 & 9.56e+02 & 3.25e+07 & 2.29e+07 & 2.92e+05 & 7.16e+05 & 1.12e+07 & 1.07e+07 & 1.98e+03 & 6.33e+02 & \textbf{6.04e+02} & 1.51e+03 \\
& Av.Ranking  & 6.08 & 5.92 & 6.92 & \textbf{4.46} & 12.00 & 11.62 & 10.08 & 9.77 & 9.31 & 10.31 & 6.00 & 4.92 & 5.62 & 4.77 \\
\midrule
\multirow{2}{*}{\rotatebox[origin=c]{90}{ALL}} & Av.MEDerr  & 2.62e+03 & 1.13e+04 & 5.61e+03 & 2.14e+03 & 1.50e+07 & 9.76e+06 & 1.08e+05 & 2.46e+05 & 3.67e+06 & 3.51e+06 & 2.12e+03 & 1.88e+03 & \textbf{1.18e+03} & 1.76e+03 \\
& Av.Ranking  & 7.40 & \textbf{5.53} & 8.03 & 5.95 & 9.68 & 8.55 & 11.90 & 11.07 & 8.22 & 7.47 & 6.82 & 6.12 & 6.38 & 5.58 \\
\midrule
& \multicolumn{1}{c}{Wilcoxon test} & \multicolumn{1}{c}{$\approx$} & \multicolumn{1}{c}{} & \multicolumn{1}{c}{$\approx$} & \multicolumn{1}{c}{$\approx$} & \multicolumn{1}{c}{$+$} & \multicolumn{1}{c}{$+$} & \multicolumn{1}{c}{$+$} & \multicolumn{1}{c}{$+$} & \multicolumn{1}{c}{$+$} & \multicolumn{1}{c}{$+$} & \multicolumn{1}{c}{$\approx$} & \multicolumn{1}{c}{$\approx$} & \multicolumn{1}{c}{$\approx$} & \multicolumn{1}{c}{$\approx$} \\

\toprule
& {Wins \newline (MAD) } & \multicolumn{1}{c}{3} & \multicolumn{1}{c}{11} & \multicolumn{1}{c}{1} & \multicolumn{1}{c}{6} & \multicolumn{1}{c}{9} & \multicolumn{1}{c}{6} & \multicolumn{1}{c}{0} & \multicolumn{1}{c}{1} & \multicolumn{1}{c}{9} & \multicolumn{1}{c}{10} & \multicolumn{1}{c}{2} & \multicolumn{1}{c}{3} & \multicolumn{1}{c}{6} & \multicolumn{1}{c}{6} \\
\midrule
\multirow{2}{*}{\rotatebox[origin=c]{90}{UNI}} & Av.MAD  & \textbf{4.86e+02} & 2.66e+03 & 4.90e+03 & 5.34e+02 & 1.69e+07 & 2.25e+06 & 1.79e+04 & 3.38e+04 & 3.44e+06 & 3.17e+05 & 1.60e+03 & 5.45e+02 & 1.54e+03 & 2.00e+03 \\
& Av.Ranking  & 6.42 & 6.42 & 7.58 & \textbf{4.67} & 8.92 & 8.50 & 11.42 & 11.58 & 8.67 & 8.58 & 7.00 & 5.42 & 5.92 & 6.33 \\
\midrule
\multirow{2}{*}{\rotatebox[origin=c]{90}{MUL}} & Av.MAD  & 1.66e+02 & 3.95e+01 & 1.02e+02 & 2.33e+02 & 4.13e+02 & 3.58e+02 & 5.51e+02 & 1.36e+02 & 1.55e+02 & 1.45e+02 & 5.33e+01 & 2.48e+02 & \textbf{3.77e+01} & 6.12e+01 \\
& Av.Ranking  & 6.77 & \textbf{4.62} & 7.27 & 7.81 & 8.35 & 8.88 & 11.92 & 10.54 & 8.12 & 7.58 & 5.42 & 7.54 & 5.92 & 6.23 \\
\midrule
\multirow{2}{*}{\rotatebox[origin=c]{90}{HYB}} & Av.MAD  & 1.49e+03 & 1.33e+03 & 1.27e+03 & 2.25e+03 & 9.74e+06 & 1.62e+06 & 2.94e+03 & 2.91e+03 & 5.87e+03 & 1.71e+04 & 1.13e+03 & 1.04e+03 & \textbf{2.43e+02} & 1.47e+03 \\
& Av.Ranking  & 3.28 & \textbf{2.70} & 4.50 & 5.25 & 6.12 & 6.88 & 8.85 & 8.50 & 8.85 & 9.20 & 8.88 & 10.45 & 10.68 & 11.22 \\
\midrule
\multirow{2}{*}{\rotatebox[origin=c]{90}{HCP}} & Av.MAD  & 1.28e+02 & \textbf{1.95e+00} & 3.92e+01 & 7.50e+01 & 2.02e+01 & 3.08e+01 & 8.14e+01 & 7.07e+01 & 1.18e+01 & 2.48e+01 & 3.13e+00 & 7.80e+00 & 2.05e+01 & 4.71e+01 \\
& Av.Ranking  & 2.54 & \textbf{2.42} & 3.84 & 4.76 & 5.88 & 6.58 & 8.10 & 8.34 & 9.20 & 9.36 & 9.98 & 11.16 & 11.66 & 12.36 \\\midrule
\multirow{2}{*}{\rotatebox[origin=c]{90}{ROT}} & Av.MAD  & 3.84e+02 & 1.57e+03 & 3.44e+03 & 4.37e+02 & 1.56e+07 & 2.08e+06 & 1.33e+04 & 2.98e+04 & 3.18e+06 & 2.90e+05 & 1.34e+03 & 5.31e+02 & 1.21e+03 & \textbf{1.76e+02} \\
& Av.Ranking  & 5.08 & 5.15 & 6.38 & 6.08 & 10.31 & 11.38 & 10.46 & 10.77 & 9.08 & 10.15 & 5.54 & 6.08 & 5.38 & \textbf{4.00} \\
\midrule
\multirow{2}{*}{\rotatebox[origin=c]{90}{ALL}} & Av.MAD  & 7.68e+02 & 1.31e+03 & 1.98e+03 & 1.09e+03 & 8.73e+06 & 1.28e+06 & 6.55e+03 & 1.13e+04 & 1.03e+06 & 1.01e+05 & 9.24e+02 & 6.47e+02 & \textbf{5.67e+02} & 1.18e+03 \\
& Av.Ranking  & 6.70 & \textbf{5.25} & 7.15 & 6.95 & 8.60 & 8.97 & 11.88 & 10.68 & 8.12 & 8.00 & 5.82 & 7.00 & 5.78 & 6.10 \\
\midrule
& \multicolumn{1}{c}{Wilcoxon test} & \multicolumn{1}{c}{$\approx$} & \multicolumn{1}{c}{} & \multicolumn{1}{c}{$\approx$} & \multicolumn{1}{c}{$\approx$} & \multicolumn{1}{c}{$+$} & \multicolumn{1}{c}{$+$} & \multicolumn{1}{c}{$+$} & \multicolumn{1}{c}{$+$} & \multicolumn{1}{c}{$\approx$} & \multicolumn{1}{c}{$+$} & \multicolumn{1}{c}{$\approx$} & \multicolumn{1}{c}{$\approx$} & \multicolumn{1}{c}{$\approx$} & \multicolumn{1}{c}{$\approx$} \\

    \end{longtable} }
\end{ThreePartTable}

%% file: tables/tbl_cec14_1.tex
\begin{ThreePartTable}
    \setlength{\tabcolsep}{0.5pt} 
    {\fontsize{6.5}{6.2}\selectfont
    \begin{longtable}{m{0.1cm} C{1.25cm} m{0.88cm} m{0.88cm} m{0.88cm} m{0.88cm} m{0.88cm} m{0.88cm} m{0.88cm} m{0.88cm} m{0.88cm} m{0.88cm} m{0.88cm}m{0.88cm} m{0.88cm} m{0.88cm}} 
    \caption{Average median (Av.MED), average median error (Av.MEDerr) and average median absolute deviation (Av.MAD) obtained by the algorithms on the 22 functions of the CEC'14 test suite with $\mathbf{d = 50}$.} \label{table:MED-MEDerr-MAD-cec14-50}\\
    \toprule
    & $f\#$ &  
    \textbf{\MTFtfoTBL} & \textbf{\MTFldoTBL} & 
    \textbf{\CMAEStfoTBL} & \textbf{\CMAESldoTBL} & 
    \textbf{\DEtfoTBL} & \textbf{\DEldoTBL} & 
    \textbf{\PSOtfoTBL} & \textbf{\PSOldoTBL} & 
    \textbf{\PDHybtfoTBL} & \textbf{\PDHybldoTBL} & 
    \textbf{\DCHybtfoTBL} & \textbf{\DCHybldoTBL} &
    \textbf{\PCHybtfoTBL} & \textbf{\PCHybldoTBL} \\
    \midrule
    \endfirsthead
    
    \caption*{ Table \ref{table:MED-MEDerr-MAD-cec14-50} (Continued.)}\\
    \toprule
    & $f\#$ &  
    \textbf{\MTFtfoTBL} & \textbf{\MTFldoTBL} & 
    \textbf{\CMAEStfoTBL} & \textbf{\CMAESldoTBL} & 
    \textbf{\DEtfoTBL} & \textbf{\DEldoTBL} & 
    \textbf{\PSOtfoTBL} & \textbf{\PSOldoTBL} & 
    \textbf{\PDHybtfoTBL} & \textbf{\PDHybldoTBL} & 
    \textbf{\DCHybtfoTBL} & \textbf{\DCHybldoTBL} &
    \textbf{\PCHybtfoTBL} & \textbf{\PCHybldoTBL} \\
    \midrule
    \endhead

    \bottomrule
    \endfoot
    
    \bottomrule
    \endlastfoot

& {Wins \newline (MED) } & \multicolumn{1}{c}{1} & \multicolumn{1}{c}{5} & \multicolumn{1}{c}{2} & \multicolumn{1}{c}{7} & \multicolumn{1}{c}{1} & \multicolumn{1}{c}{0} & \multicolumn{1}{c}{2} & \multicolumn{1}{c}{2} & \multicolumn{1}{c}{1} & \multicolumn{1}{c}{4} & \multicolumn{1}{c}{2} & \multicolumn{1}{c}{4} & \multicolumn{1}{c}{5} & \multicolumn{1}{c}{4} \\
\midrule
\multirow{2}{*}{\rotatebox[origin=c]{90}{UNI}} & Av.MED  & \textbf{0.00e+00} & 2.80e+03 & 9.03e+00 & \textbf{0.00e+00} & 1.16e+07 & 1.25e+07 & 4.18e+05 & 8.35e+05 & 6.84e+06 & 8.72e+06 & \textbf{0.00e+00} & \textbf{0.00e+00} & \textbf{0.00e+00} & \textbf{0.00e+00} \\
& Av.Ranking  & 3.33 & 5.67 & 7.00 & \textbf{2.00} & 12.00 & 13.00 & 9.67 & 10.67 & 10.33 & 13.33 & 6.67 & 3.67 & 4.67 & 3.33 \\
\midrule
\multirow{2}{*}{\rotatebox[origin=c]{90}{MUL}} & Av.MED  & 3.26e+03 & 3.26e+03 & 3.42e+03 & 3.35e+03 & 4.60e+03 & 4.92e+03 & 4.33e+03 & 6.42e+03 & 3.88e+03 & 6.69e+03 & 3.27e+03 & 3.35e+03 & 3.28e+03 & \textbf{3.24e+03} \\
& Av.Ranking  & 7.31 & \textbf{5.62} & 7.19 & 6.19 & 12.12 & 10.69 & 12.00 & 10.00 & 8.69 & 7.69 & 6.44 & 7.00 & 6.38 & 5.88 \\
\midrule
\multirow{2}{*}{\rotatebox[origin=c]{90}{HYB}} & Av.MED  & 1.91e+03 & 8.82e+03 & 5.14e+03 & \textbf{1.73e+03} & 3.07e+06 & 1.42e+06 & 3.28e+04 & 3.32e+04 & 5.93e+04 & 5.33e+04 & 1.78e+03 & \textbf{1.73e+03} & \textbf{1.73e+03} & 1.77e+03 \\
& Av.Ranking  & \textbf{2.82} & 2.86 & 4.00 & 4.23 & 7.14 & 7.55 & 8.55 & 8.14 & 8.73 & 9.86 & 9.55 & 10.18 & 10.55 & 11.95 \\
\midrule
\multirow{2}{*}{\rotatebox[origin=c]{90}{ROT}} & Av.MED  & 1.68e+03 & 2.29e+03 & 1.76e+03 & 1.74e+03 & 2.49e+06 & 2.68e+06 & 9.17e+04 & 1.81e+05 & 1.47e+06 & 1.87e+06 & 1.69e+03 & 1.73e+03 & 1.70e+03 & \textbf{1.67e+03} \\
& Av.Ranking  & 5.86 & 5.57 & 6.57 & 5.64 & 12.14 & 11.93 & 10.93 & 10.07 & 9.71 & 9.71 & 6.64 & 6.07 & 6.43 & \textbf{5.14} \\
\midrule
\multirow{2}{*}{\rotatebox[origin=c]{90}{ALL}} & Av.MED  & 2.68e+03 & 5.26e+03 & 3.81e+03 & 2.68e+03 & 2.56e+06 & 2.16e+06 & 7.01e+04 & 1.27e+05 & 9.53e+05 & 1.21e+06 & 2.64e+03 & 2.68e+03 & 2.64e+03 & \textbf{2.62e+03} \\
& Av.Ranking  & 6.68 & \textbf{5.50} & 6.82 & 5.86 & 12.27 & 11.14 & 11.68 & 9.91 & 8.86 & 8.86 & 6.23 & 6.18 & 5.73 & \textbf{5.50} \\
\midrule
& \multicolumn{1}{c}{Wilcoxon test} & \multicolumn{1}{c}{$\approx$} & \multicolumn{1}{c}{} & \multicolumn{1}{c}{$\approx$} & \multicolumn{1}{c}{$\approx$} & \multicolumn{1}{c}{$+$} & \multicolumn{1}{c}{$+$} & \multicolumn{1}{c}{$+$} & \multicolumn{1}{c}{$+$} & \multicolumn{1}{c}{$+$} & \multicolumn{1}{c}{$\approx$} & \multicolumn{1}{c}{$\approx$} & \multicolumn{1}{c}{$\approx$} & \multicolumn{1}{c}{$\approx$} & \multicolumn{1}{c}{$\approx$} \\
& \multicolumn{1}{c}{Wilcoxon test} & \multicolumn{1}{c}{$\approx$} & \multicolumn{1}{c}{$\approx$} & \multicolumn{1}{c}{$\approx$} & \multicolumn{1}{c}{$\approx$} & \multicolumn{1}{c}{$+$} & \multicolumn{1}{c}{$\approx$} & \multicolumn{1}{c}{$+$} & \multicolumn{1}{c}{$\approx$} & \multicolumn{1}{c}{$\approx$} & \multicolumn{1}{c}{$\approx$} & \multicolumn{1}{c}{$\approx$} & \multicolumn{1}{c}{$\approx$} & \multicolumn{1}{c}{$\approx$} & \multicolumn{1}{c}{} \\

\toprule
& {Wins \newline (MEDerr) } & \multicolumn{1}{c}{1} & \multicolumn{1}{c}{5} & \multicolumn{1}{c}{2} & \multicolumn{1}{c}{7} & \multicolumn{1}{c}{1} & \multicolumn{1}{c}{0} & \multicolumn{1}{c}{2} & \multicolumn{1}{c}{2} & \multicolumn{1}{c}{1} & \multicolumn{1}{c}{4} & \multicolumn{1}{c}{2} & \multicolumn{1}{c}{4} & \multicolumn{1}{c}{5} & \multicolumn{1}{c}{4} \\
\midrule
\multirow{2}{*}{\rotatebox[origin=c]{90}{UNI}} & Av.MEDerr  & \textbf{0.00e+00} & 2.80e+03 & 9.03e+00 & \textbf{0.00e+00} & 1.16e+07 & 1.25e+07 & 4.18e+05 & 8.35e+05 & 6.84e+06 & 8.72e+06 & \textbf{0.00e+00} & \textbf{0.00e+00} & \textbf{0.00e+00} & \textbf{0.00e+00} \\
& Av.Ranking  & 3.33 & 5.67 & 7.00 & \textbf{2.00} & 12.00 & 13.00 & 9.67 & 10.67 & 10.33 & 13.33 & 6.67 & 3.67 & 4.67 & 3.33 \\
\midrule
\multirow{2}{*}{\rotatebox[origin=c]{90}{MUL}} & Av.MEDerr  & 1.17e+02 & 1.19e+02 & 2.73e+02 & 2.10e+02 & 1.46e+03 & 1.78e+03 & 1.18e+03 & 3.28e+03 & 7.42e+02 & 3.55e+03 & 1.23e+02 & 2.03e+02 & 1.37e+02 & \textbf{9.42e+01} \\
& Av.Ranking  & 7.31 & \textbf{5.62} & 7.19 & 6.19 & 12.12 & 10.69 & 12.00 & 10.00 & 8.69 & 7.69 & 6.44 & 7.00 & 6.38 & 5.88 \\
\midrule
\multirow{2}{*}{\rotatebox[origin=c]{90}{HYB}} & Av.MEDerr  & 2.50e+02 & 7.16e+03 & 3.48e+03 & 6.45e+01 & 3.07e+06 & 1.42e+06 & 3.11e+04 & 3.15e+04 & 5.77e+04 & 5.17e+04 & 1.17e+02 & 6.59e+01 & \textbf{6.36e+01} & 1.09e+02 \\
& Av.Ranking  & \textbf{2.82} & 2.86 & 4.00 & 4.23 & 7.14 & 7.55 & 8.55 & 8.14 & 8.73 & 9.86 & 9.55 & 10.18 & 10.55 & 11.95 \\
\midrule
\multirow{2}{*}{\rotatebox[origin=c]{90}{ROT}} & Av.MEDerr  & 7.38e+01 & 6.87e+02 & 1.54e+02 & 1.32e+02 & 2.49e+06 & 2.68e+06 & 9.01e+04 & 1.79e+05 & 1.47e+06 & 1.87e+06 & 8.07e+01 & 1.24e+02 & 9.50e+01 & \textbf{5.91e+01} \\
& Av.Ranking  & 5.86 & 5.57 & 6.57 & 5.64 & 12.14 & 11.93 & 10.93 & 10.07 & 9.71 & 9.71 & 6.64 & 6.07 & 6.43 & \textbf{5.14} \\
\midrule
\multirow{2}{*}{\rotatebox[origin=c]{90}{ALL}} & Av.MEDerr  & 1.54e+02 & 2.73e+03 & 1.28e+03 & 1.54e+02 & 2.56e+06 & 2.16e+06 & 6.76e+04 & 1.24e+05 & 9.51e+05 & 1.21e+06 & 1.15e+02 & 1.50e+02 & 1.07e+02 & \textbf{9.45e+01} \\
& Av.Ranking  & 6.68 & \textbf{5.50} & 6.82 & 5.86 & 12.27 & 11.14 & 11.68 & 9.91 & 8.86 & 8.86 & 6.23 & 6.18 & 5.73 & \textbf{5.50} \\
\midrule
& \multicolumn{1}{c}{Wilcoxon test} & \multicolumn{1}{c}{$\approx$} & \multicolumn{1}{c}{} & \multicolumn{1}{c}{$\approx$} & \multicolumn{1}{c}{$\approx$} & \multicolumn{1}{c}{$+$} & \multicolumn{1}{c}{$+$} & \multicolumn{1}{c}{$+$} & \multicolumn{1}{c}{$+$} & \multicolumn{1}{c}{$+$} & \multicolumn{1}{c}{$\approx$} & \multicolumn{1}{c}{$\approx$} & \multicolumn{1}{c}{$\approx$} & \multicolumn{1}{c}{$\approx$} & \multicolumn{1}{c}{$\approx$} \\
& \multicolumn{1}{c}{Wilcoxon test} & \multicolumn{1}{c}{$\approx$} & \multicolumn{1}{c}{$\approx$} & \multicolumn{1}{c}{$\approx$} & \multicolumn{1}{c}{$\approx$} & \multicolumn{1}{c}{$+$} & \multicolumn{1}{c}{$\approx$} & \multicolumn{1}{c}{$+$} & \multicolumn{1}{c}{$\approx$} & \multicolumn{1}{c}{$\approx$} & \multicolumn{1}{c}{$\approx$} & \multicolumn{1}{c}{$\approx$} & \multicolumn{1}{c}{$\approx$} & \multicolumn{1}{c}{$\approx$} & \multicolumn{1}{c}{} \\

\toprule
& {Wins \newline (MAD) } & \multicolumn{1}{c}{0} & \multicolumn{1}{c}{4} & \multicolumn{1}{c}{3} & \multicolumn{1}{c}{1} & \multicolumn{1}{c}{0} & \multicolumn{1}{c}{0} & \multicolumn{1}{c}{1} & \multicolumn{1}{c}{0} & \multicolumn{1}{c}{0} & \multicolumn{1}{c}{3} & \multicolumn{1}{c}{0} & \multicolumn{1}{c}{4} & \multicolumn{1}{c}{4} & \multicolumn{1}{c}{3} \\
\midrule
\multirow{2}{*}{\rotatebox[origin=c]{90}{UNI}} & Av.MAD  & \textbf{0.00e+00} & 6.89e+02 & 7.93e+00 & \textbf{0.00e+00} & 1.04e+07 & 3.64e+06 & 7.53e+04 & 6.97e+04 & 1.45e+06 & 9.01e+05 & \textbf{0.00e+00} & \textbf{0.00e+00} & \textbf{0.00e+00} & \textbf{0.00e+00} \\
& Av.Ranking  & 4.33 & 5.67 & 7.00 & 3.33 & 12.33 & 12.67 & 10.00 & 10.33 & 10.67 & 13.00 & 6.67 & 3.00 & 3.67 & \textbf{2.33} \\
\midrule
\multirow{2}{*}{\rotatebox[origin=c]{90}{MUL}} & Av.MAD  & 3.60e+01 & 4.08e+01 & 4.51e+01 & 4.42e+01 & 2.58e+02 & 3.21e+02 & 4.86e+01 & 1.23e+03 & 1.33e+02 & 1.86e+03 & 3.66e+01 & 5.30e+01 & \textbf{2.81e+01} & 4.45e+01 \\
& Av.Ranking  & 8.12 & \textbf{4.94} & 7.12 & 7.44 & 10.00 & 9.50 & 10.25 & 10.44 & 7.75 & 7.88 & 7.38 & 6.06 & 5.31 & 6.00 \\
\midrule
\multirow{2}{*}{\rotatebox[origin=c]{90}{HYB}} & Av.MAD  & 8.68e+01 & 1.86e+03 & 3.87e+02 & 5.71e+01 & 2.85e+06 & 6.35e+05 & 1.18e+04 & 6.48e+03 & 5.54e+03 & 8.68e+03 & 4.79e+01 & 3.11e+01 & \textbf{5.28e+00} & 4.56e+01 \\
& Av.Ranking  & \textbf{2.32} & 3.18 & 3.59 & 5.23 & 6.27 & 7.00 & 8.09 & 8.82 & 8.82 & 9.86 & 9.59 & 10.18 & 10.77 & 11.36 \\
\midrule
\multirow{2}{*}{\rotatebox[origin=c]{90}{ROT}} & Av.MAD  & 2.16e+01 & 1.65e+02 & 1.91e+01 & 2.01e+01 & 2.22e+06 & 7.81e+05 & 1.61e+04 & 1.50e+04 & 3.12e+05 & 1.93e+05 & 2.72e+01 & 2.74e+01 & \textbf{1.10e+01} & 2.61e+01 \\
& Av.Ranking  & 7.43 & 5.07 & 7.14 & 6.36 & 11.29 & 10.36 & 10.29 & 10.07 & 8.57 & 9.43 & 7.50 & 5.07 & 5.29 & \textbf{4.71} \\
\midrule
\multirow{2}{*}{\rotatebox[origin=c]{90}{ALL}} & Av.MAD  & 5.05e+01 & 7.02e+02 & 1.53e+02 & 4.63e+01 & 2.32e+06 & 6.99e+05 & 1.40e+04 & 1.16e+04 & 2.00e+05 & 1.26e+05 & 3.84e+01 & 4.36e+01 & \textbf{1.96e+01} & 4.28e+01 \\
& Av.Ranking  & 6.95 & 5.55 & 6.95 & 7.00 & 10.73 & 10.00 & 10.41 & 10.64 & 8.14 & 8.68 & 6.77 & 5.41 & 5.18 & \textbf{5.00} \\
\midrule
& \multicolumn{1}{c}{Wilcoxon test} & \multicolumn{1}{c}{$\approx$} & \multicolumn{1}{c}{$\approx$} & \multicolumn{1}{c}{$\approx$} & \multicolumn{1}{c}{$\approx$} & \multicolumn{1}{c}{$\approx$} & \multicolumn{1}{c}{$\approx$} & \multicolumn{1}{c}{$\approx$} & \multicolumn{1}{c}{$\approx$} & \multicolumn{1}{c}{$\approx$} & \multicolumn{1}{c}{$\approx$} & \multicolumn{1}{c}{$\approx$} & \multicolumn{1}{c}{$\approx$} & \multicolumn{1}{c}{$\approx$} & \multicolumn{1}{c}{} \\

    \end{longtable} }
\end{ThreePartTable}

%% file: tables/tbl_cec14_2.tex
\begin{ThreePartTable}
    \setlength{\tabcolsep}{0.5pt} 
    {\fontsize{6.5}{6.2}\selectfont
    \begin{longtable}{m{0.1cm} C{1.25cm} m{0.88cm} m{0.88cm} m{0.88cm} m{0.88cm} m{0.88cm} m{0.88cm} m{0.88cm} m{0.88cm} m{0.88cm} m{0.88cm} m{0.88cm}m{0.88cm} m{0.88cm} m{0.88cm}} 
    \caption{Average median (Av.MED), average median error (Av.MEDerr) and average median absolute deviation (Av.MAD) obtained by the algorithms on the 22 functions of the CEC'14 test suite with $\mathbf{d = 100}$.} \label{table:MED-MEDerr-MAD-cec14-100}\\
    \toprule
    & $f\#$ &  
    \textbf{\MTFtfoTBL} & \textbf{\MTFldoTBL} & 
    \textbf{\CMAEStfoTBL} & \textbf{\CMAESldoTBL} & 
    \textbf{\DEtfoTBL} & \textbf{\DEldoTBL} & 
    \textbf{\PSOtfoTBL} & \textbf{\PSOldoTBL} & 
    \textbf{\PDHybtfoTBL} & \textbf{\PDHybldoTBL} & 
    \textbf{\DCHybtfoTBL} & \textbf{\DCHybldoTBL} &
    \textbf{\PCHybtfoTBL} & \textbf{\PCHybldoTBL} \\
    \midrule
    \endfirsthead
    
    \caption*{ Table \ref{table:MED-MEDerr-MAD-cec14-100} (Continued.)}\\
    \toprule
    & $f\#$ &  
    \textbf{\MTFtfoTBL} & \textbf{\MTFldoTBL} & 
    \textbf{\CMAEStfoTBL} & \textbf{\CMAESldoTBL} & 
    \textbf{\DEtfoTBL} & \textbf{\DEldoTBL} & 
    \textbf{\PSOtfoTBL} & \textbf{\PSOldoTBL} & 
    \textbf{\PDHybtfoTBL} & \textbf{\PDHybldoTBL} & 
    \textbf{\DCHybtfoTBL} & \textbf{\DCHybldoTBL} &
    \textbf{\PCHybtfoTBL} & \textbf{\PCHybldoTBL} \\
    \midrule
    \endhead

    \bottomrule
    \endfoot
    
    \bottomrule
    \endlastfoot

& {Wins \newline (MED) } & \multicolumn{1}{c}{4} & \multicolumn{1}{c}{4} & \multicolumn{1}{c}{4} & \multicolumn{1}{c}{8} & \multicolumn{1}{c}{1} & \multicolumn{1}{c}{1} & \multicolumn{1}{c}{1} & \multicolumn{1}{c}{2} & \multicolumn{1}{c}{2} & \multicolumn{1}{c}{4} & \multicolumn{1}{c}{3} & \multicolumn{1}{c}{6} & \multicolumn{1}{c}{8} & \multicolumn{1}{c}{5} \\
\midrule
\multirow{2}{*}{\rotatebox[origin=c]{90}{UNI}} & Av.MED  & 8.80e+03 & 8.57e+04 & 4.07e+04 & 2.38e+03 & 1.41e+08 & 9.93e+07 & 1.26e+06 & 3.10e+06 & 4.87e+07 & 4.64e+07 & 8.10e+03 & 2.11e+03 & \textbf{2.03e+03} & 6.07e+03 \\
& Av.Ranking  & 5.00 & 6.00 & 7.00 & \textbf{2.33} & 12.67 & 12.33 & 9.00 & 11.33 & 10.33 & 13.00 & 5.67 & 2.67 & 4.33 & 3.33 \\
\midrule
\multirow{2}{*}{\rotatebox[origin=c]{90}{MUL}} & Av.MED  & 6.75e+03 & 6.78e+03 & 7.04e+03 & 7.02e+03 & 9.23e+03 & 9.07e+03 & 8.74e+03 & 9.23e+03 & 7.61e+03 & 1.01e+04 & \textbf{6.63e+03} & 6.69e+03 & 6.68e+03 & 6.65e+03 \\
& Av.Ranking  & 7.56 & 6.50 & 7.38 & \textbf{6.00} & 12.00 & 10.25 & 11.56 & 10.00 & 8.56 & 8.12 & 6.88 & 7.06 & 6.69 & 6.44 \\
\midrule
\multirow{2}{*}{\rotatebox[origin=c]{90}{HYB}} & Av.MED  & 1.08e+04 & 2.80e+04 & 1.76e+04 & 9.18e+03 & 2.54e+07 & 1.32e+07 & 5.60e+04 & 7.09e+04 & 1.33e+05 & 1.54e+05 & 1.02e+04 & 8.01e+03 & \textbf{7.36e+03} & 9.58e+03 \\
& Av.Ranking  & \textbf{2.64} & 3.55 & 3.91 & 4.27 & 7.23 & 7.50 & 8.27 & 8.27 & 9.05 & 9.86 & 9.68 & 10.09 & 10.59 & 11.59 \\
\midrule
\multirow{2}{*}{\rotatebox[origin=c]{90}{ROT}} & Av.MED  & 5.33e+03 & 2.17e+04 & 1.23e+04 & 4.14e+03 & 3.01e+07 & 2.13e+07 & 2.75e+05 & 6.68e+05 & 1.04e+07 & 9.94e+06 & 5.09e+03 & 3.84e+03 & \textbf{3.82e+03} & 4.66e+03 \\
& Av.Ranking  & 6.64 & 6.43 & 6.71 & \textbf{5.07} & 12.00 & 11.64 & 10.64 & 10.50 & 9.43 & 10.21 & 6.57 & 6.00 & 6.57 & 5.64 \\
\midrule
\multirow{2}{*}{\rotatebox[origin=c]{90}{ALL}} & Av.MED  & 8.92e+03 & 2.48e+04 & 1.56e+04 & 7.71e+03 & 2.73e+07 & 1.77e+07 & 1.95e+05 & 4.50e+05 & 6.68e+06 & 6.38e+06 & 8.56e+03 & 7.09e+03 & \textbf{6.87e+03} & 8.11e+03 \\
& Av.Ranking  & 7.05 & 6.50 & 6.95 & \textbf{5.36} & 12.23 & 10.73 & 11.23 & 10.32 & 9.05 & 9.18 & 6.32 & 6.00 & 5.86 & 5.68 \\
\midrule
& \multicolumn{1}{c}{Wilcoxon test} & \multicolumn{1}{c}{$\approx$} & \multicolumn{1}{c}{$\approx$} & \multicolumn{1}{c}{$\approx$} & \multicolumn{1}{c}{} & \multicolumn{1}{c}{$+$} & \multicolumn{1}{c}{$\approx$} & \multicolumn{1}{c}{$+$} & \multicolumn{1}{c}{$+$} & \multicolumn{1}{c}{$\approx$} & \multicolumn{1}{c}{$\approx$} & \multicolumn{1}{c}{$\approx$} & \multicolumn{1}{c}{$\approx$} & \multicolumn{1}{c}{$\approx$} & \multicolumn{1}{c}{$\approx$} \\

\toprule
& {Wins \newline (MEDerr) } & \multicolumn{1}{c}{4} & \multicolumn{1}{c}{4} & \multicolumn{1}{c}{4} & \multicolumn{1}{c}{8} & \multicolumn{1}{c}{1} & \multicolumn{1}{c}{1} & \multicolumn{1}{c}{1} & \multicolumn{1}{c}{2} & \multicolumn{1}{c}{2} & \multicolumn{1}{c}{4} & \multicolumn{1}{c}{3} & \multicolumn{1}{c}{6} & \multicolumn{1}{c}{8} & \multicolumn{1}{c}{5} \\
\midrule
\multirow{2}{*}{\rotatebox[origin=c]{90}{UNI}} & Av.MEDerr  & 8.63e+03 & 8.53e+04 & 4.03e+04 & 2.20e+03 & 1.41e+08 & 9.93e+07 & 1.26e+06 & 3.10e+06 & 4.87e+07 & 4.64e+07 & 7.94e+03 & 1.93e+03 & \textbf{1.85e+03} & 5.90e+03 \\
& Av.Ranking  & 5.00 & 6.00 & 7.00 & \textbf{2.33} & 12.67 & 12.33 & 9.00 & 11.33 & 10.33 & 13.00 & 5.67 & 2.67 & 4.33 & 3.33 \\
\midrule
\multirow{2}{*}{\rotatebox[origin=c]{90}{MUL}} & Av.MEDerr  & 4.41e+02 & 4.70e+02 & 7.33e+02 & 7.03e+02 & 2.92e+03 & 2.76e+03 & 2.43e+03 & 2.92e+03 & 1.30e+03 & 3.80e+03 & \textbf{3.18e+02} & 3.78e+02 & 3.67e+02 & 3.40e+02 \\
& Av.Ranking  & 7.56 & 6.50 & 7.38 & \textbf{6.00} & 12.00 & 10.25 & 11.56 & 10.00 & 8.56 & 8.12 & 6.88 & 7.06 & 6.69 & 6.44 \\
\midrule
\multirow{2}{*}{\rotatebox[origin=c]{90}{HYB}} & Av.MEDerr  & 6.47e+03 & 2.37e+04 & 1.33e+04 & 4.87e+03 & 2.54e+07 & 1.32e+07 & 5.17e+04 & 6.66e+04 & 1.29e+05 & 1.50e+05 & 5.88e+03 & 3.70e+03 & \textbf{3.05e+03} & 5.27e+03 \\
& Av.Ranking  & \textbf{2.64} & 3.55 & 3.91 & 4.27 & 7.23 & 7.50 & 8.27 & 8.27 & 9.05 & 9.86 & 9.68 & 10.09 & 10.59 & 11.59 \\
\midrule
\multirow{2}{*}{\rotatebox[origin=c]{90}{ROT}} & Av.MEDerr  & 2.07e+03 & 1.85e+04 & 8.98e+03 & 8.88e+02 & 3.01e+07 & 2.13e+07 & 2.71e+05 & 6.65e+05 & 1.04e+07 & 9.94e+06 & 1.84e+03 & 5.88e+02 & \textbf{5.62e+02} & 1.40e+03 \\
& Av.Ranking  & 6.64 & 6.43 & 6.71 & \textbf{5.07} & 12.00 & 11.64 & 10.64 & 10.50 & 9.43 & 10.21 & 6.57 & 6.00 & 6.57 & 5.64 \\
\midrule
\multirow{2}{*}{\rotatebox[origin=c]{90}{ALL}} & Av.MEDerr  & 3.52e+03 & 1.93e+04 & 1.02e+04 & 2.30e+03 & 2.73e+07 & 1.77e+07 & 1.90e+05 & 4.45e+05 & 6.68e+06 & 6.37e+06 & 3.16e+03 & 1.68e+03 & \textbf{1.45e+03} & 2.70e+03 \\
& Av.Ranking  & 7.05 & 6.50 & 6.95 & \textbf{5.36} & 12.23 & 10.73 & 11.23 & 10.32 & 9.05 & 9.18 & 6.32 & 6.00 & 5.86 & 5.68 \\
\midrule
& \multicolumn{1}{c}{Wilcoxon test} & \multicolumn{1}{c}{$\approx$} & \multicolumn{1}{c}{$\approx$} & \multicolumn{1}{c}{$\approx$} & \multicolumn{1}{c}{} & \multicolumn{1}{c}{$+$} & \multicolumn{1}{c}{$\approx$} & \multicolumn{1}{c}{$+$} & \multicolumn{1}{c}{$+$} & \multicolumn{1}{c}{$\approx$} & \multicolumn{1}{c}{$\approx$} & \multicolumn{1}{c}{$\approx$} & \multicolumn{1}{c}{$\approx$} & \multicolumn{1}{c}{$\approx$} & \multicolumn{1}{c}{$\approx$} \\

\toprule
& {Wins \newline (MAD) } & \multicolumn{1}{c}{3} & \multicolumn{1}{c}{5} & \multicolumn{1}{c}{1} & \multicolumn{1}{c}{6} & \multicolumn{1}{c}{2} & \multicolumn{1}{c}{1} & \multicolumn{1}{c}{1} & \multicolumn{1}{c}{1} & \multicolumn{1}{c}{4} & \multicolumn{1}{c}{2} & \multicolumn{1}{c}{2} & \multicolumn{1}{c}{2} & \multicolumn{1}{c}{3} & \multicolumn{1}{c}{6} \\
\midrule
\multirow{2}{*}{\rotatebox[origin=c]{90}{UNI}} & Av.MAD  & 9.83e+02 & 6.71e+03 & 1.45e+04 & 8.90e+02 & 6.77e+07 & 9.00e+06 & 5.71e+04 & 1.29e+05 & 1.38e+07 & 1.26e+06 & 5.68e+03 & 1.20e+03 & 5.14e+03 & \textbf{6.47e+02} \\
& Av.Ranking  & 4.33 & 6.00 & 7.00 & \textbf{2.00} & 12.67 & 11.33 & 10.00 & 11.00 & 10.33 & 13.00 & 6.00 & 3.33 & 4.33 & 3.67 \\
\midrule
\multirow{2}{*}{\rotatebox[origin=c]{90}{MUL}} & Av.MAD  & 1.77e+02 & 2.13e+02 & 1.09e+02 & 2.47e+02 & 3.69e+02 & 3.43e+02 & 2.40e+02 & 4.06e+02 & 1.54e+02 & 1.18e+03 & 6.02e+01 & 2.65e+02 & \textbf{4.09e+01} & 5.83e+01 \\
& Av.Ranking  & 6.75 & 5.69 & 6.94 & 8.19 & 9.94 & 9.69 & 11.56 & 9.50 & 8.56 & 8.44 & 6.12 & 8.25 & 6.38 & \textbf{5.56} \\
\midrule
\multirow{2}{*}{\rotatebox[origin=c]{90}{HYB}} & Av.MAD  & 3.15e+03 & 2.85e+03 & 2.71e+03 & 4.79e+03 & 2.09e+07 & 3.48e+06 & 6.13e+03 & 6.19e+03 & 1.26e+04 & 3.66e+04 & 2.42e+03 & 2.21e+03 & \textbf{5.16e+02} & 3.15e+03 \\
& Av.Ranking  & \textbf{2.55} & 2.91 & 3.82 & 4.59 & 6.55 & 7.23 & 8.05 & 7.91 & 9.27 & 10.05 & 9.00 & 10.64 & 11.00 & 11.50 \\
\midrule
\multirow{2}{*}{\rotatebox[origin=c]{90}{ROT}} & Av.MAD  & 3.57e+02 & 1.46e+03 & 3.20e+03 & 4.06e+02 & 1.45e+07 & 1.93e+06 & 1.23e+04 & 2.77e+04 & 2.95e+06 & 2.69e+05 & 1.24e+03 & 4.93e+02 & 1.13e+03 & \textbf{1.64e+02} \\
& Av.Ranking  & 5.79 & 6.14 & 6.93 & 6.79 & 10.14 & 10.93 & 11.21 & 10.36 & 9.00 & 10.14 & 6.64 & 7.36 & 6.29 & \textbf{4.64} \\
\midrule
\multirow{2}{*}{\rotatebox[origin=c]{90}{ALL}} & Av.MAD  & 1.25e+03 & 1.85e+03 & 2.91e+03 & 1.80e+03 & 1.59e+07 & 2.33e+06 & 9.88e+03 & 1.96e+04 & 1.88e+06 & 1.83e+05 & 1.58e+03 & 1.03e+03 & \textbf{8.90e+02} & 1.13e+03 \\
& Av.Ranking  & 6.36 & 5.95 & 6.64 & 7.05 & 10.32 & 10.32 & 11.36 & 9.55 & 8.73 & 9.41 & 5.91 & 7.45 & 5.68 & \textbf{5.05} \\
\midrule
& \multicolumn{1}{c}{Wilcoxon test} & \multicolumn{1}{c}{$\approx$} & \multicolumn{1}{c}{$\approx$} & \multicolumn{1}{c}{$\approx$} & \multicolumn{1}{c}{$\approx$} & \multicolumn{1}{c}{$+$} & \multicolumn{1}{c}{$\approx$} & \multicolumn{1}{c}{$+$} & \multicolumn{1}{c}{$\approx$} & \multicolumn{1}{c}{$\approx$} & \multicolumn{1}{c}{$\approx$} & \multicolumn{1}{c}{$\approx$} & \multicolumn{1}{c}{$\approx$} & \multicolumn{1}{c}{$\approx$} & \multicolumn{1}{c}{} \\

    \end{longtable} }
\end{ThreePartTable}

%% file: tables/tbl_LD_3.tex
\begin{ThreePartTable}
\setlength{\tabcolsep}{0.5pt} 
    {\fontsize{6.5}{6.2}\selectfont
    \begin{longtable}{m{0.1cm} C{1.25cm} m{0.88cm} m{0.88cm} m{0.88cm} m{0.88cm} m{0.88cm} m{0.88cm} m{0.88cm} m{0.88cm} m{0.88cm} m{0.88cm} m{0.88cm} m{0.88cm} m{0.88cm} m{0.88cm}} 
    \caption{Average median (Av.MED), average median error (Av.MEDerr) and average median absolute deviation (Av.MAD) obtained by the algorithms on the 19 functions of the SOCO'10 test suite with $\mathbf{d = 750}$.}\label{table:MED-MEDerr-MAD-LD-750}\\
    \toprule
    & $f\#$ &  
    \textbf{\MTFtfoTBL} &  \textbf{\MTFldoTBL} & 
    \textbf{\CMAEStfoTBL} & \textbf{\CMAESldoTBL} & 
    \textbf{\DEtfoTBL} & \textbf{\DEldoTBL} & 
    \textbf{\PSOtfoTBL} & \textbf{\PSOldoTBL} & 
    \textbf{\PDHybtfoTBL} & \textbf{\PDHybldoTBL} & 
    \textbf{\DCHybtfoTBL} & \textbf{\DCHybldoTBL} &
    \textbf{\PCHybtfoTBL} & \textbf{\PCHybldoTBL} \\
    \midrule
    \endfirsthead
    
    \caption*{ Table \ref{table:MED-MEDerr-MAD-LD-750} (Continued.)}\\
    \toprule
    & $f\#$ &  
    \textbf{\MTFtfoTBL} &  \textbf{\MTFldoTBL} & 
    \textbf{\CMAEStfoTBL} & \textbf{\CMAESldoTBL} & 
    \textbf{\DEtfoTBL} & \textbf{\DEldoTBL} & 
    \textbf{\PSOtfoTBL} & \textbf{\PSOldoTBL} & 
    \textbf{\PDHybtfoTBL} & \textbf{\PDHybldoTBL} & 
    \textbf{\DCHybtfoTBL} & \textbf{\DCHybldoTBL} &
    \textbf{\PCHybtfoTBL} & \textbf{\PCHybldoTBL} \\
    \midrule
    \endhead
    
    \bottomrule
    \endfoot
    
    \bottomrule
    \endlastfoot

& {Wins \newline (MED) } & \multicolumn{1}{c}{0} & \multicolumn{1}{c}{0} & \multicolumn{1}{c}{2} & \multicolumn{1}{c}{2} & \multicolumn{1}{c}{4} & \multicolumn{1}{c}{0} & \multicolumn{1}{c}{1} & \multicolumn{1}{c}{0} & \multicolumn{1}{c}{0} & \multicolumn{1}{c}{0} & \multicolumn{1}{c}{2} & \multicolumn{1}{c}{1} & \multicolumn{1}{c}{8} & \multicolumn{1}{c}{0} \\
\midrule
\multirow{2}{*}{\rotatebox[origin=c]{90}{SHU}} & Av.MED  & 1.43e+09 & 1.43e+09 & 1.43e+09 & 1.43e+09 & 4.03e+05 & 3.69e+05 & \textbf{4.98e+04} & 1.43e+09 & 1.43e+09 & 1.43e+09 & 1.43e+09 & 1.43e+09 & 1.43e+09 & 1.43e+09 \\
& Av.Ranking  & 5.57 & 7.14 & 5.71 & 5.86 & 8.14 & 10.86 & 9.00 & 13.00 & 11.29 & 11.57 & \textbf{5.29} & 8.00 & 6.14 & \textbf{5.29} \\
\midrule
\multirow{2}{*}{\rotatebox[origin=c]{90}{MUL}} & Av.MED  & 7.83e+04 & 7.75e+04 & 7.91e+04 & 8.00e+04 & \textbf{7.34e+04} & 2.00e+09 & 3.37e+05 & 2.00e+09 & 1.26e+05 & 7.93e+04 & 7.46e+04 & 7.81e+04 & 7.42e+04 & 7.59e+04 \\
& Av.Ranking  & 7.91 & 6.00 & 6.73 & 7.64 & 5.36 & 12.45 & 11.73 & 14.00 & 8.73 & 7.09 & 4.09 & 6.91 & \textbf{3.36} & 5.09 \\
\midrule
\multirow{2}{*}{\rotatebox[origin=c]{90}{HYB}} & Av.MED  & 1.11e+09 & 1.11e+09 & 4.51e+04 & 4.57e+04 & \textbf{4.19e+04} & 2.22e+09 & 7.01e+04 & 4.44e+09 & 6.42e+04 & 4.70e+04 & 1.11e+09 & 4.46e+04 & 1.11e+09 & 1.11e+09 \\
& Av.Ranking  & 4.53 & \textbf{4.00} & 4.53 & 6.00 & 4.79 & 8.58 & 8.95 & 10.53 & 8.53 & 8.16 & 8.21 & 9.89 & 8.89 & 10.47 \\
\midrule
\multirow{2}{*}{\rotatebox[origin=c]{90}{ALL}} & Av.MED  & 1.05e+09 & 1.05e+09 & 5.26e+08 & 5.26e+08 & 1.69e+05 & 1.58e+09 & \textbf{1.14e+05} & 3.16e+09 & 5.26e+08 & 5.26e+08 & 1.05e+09 & 5.26e+08 & 1.05e+09 & 1.05e+09 \\
& Av.Ranking  & 7.11 & 6.47 & 6.32 & 7.11 & 6.37 & 11.84 & 10.79 & 13.63 & 9.79 & 8.47 & 4.47 & 7.37 & \textbf{4.26} & 5.11 \\
\midrule
& \multicolumn{1}{c}{Wilcoxon test} & \multicolumn{1}{c}{$\approx$} & \multicolumn{1}{c}{$\approx$} & \multicolumn{1}{c}{$\approx$} & \multicolumn{1}{c}{$\approx$} & \multicolumn{1}{c}{$\approx$} & \multicolumn{1}{c}{$\approx$} & \multicolumn{1}{c}{$\approx$} & \multicolumn{1}{c}{$+$} & \multicolumn{1}{c}{$\approx$} & \multicolumn{1}{c}{$\approx$} & \multicolumn{1}{c}{$\approx$} & \multicolumn{1}{c}{$\approx$} & \multicolumn{1}{c}{} & \multicolumn{1}{c}{$\approx$} \\

\toprule
& {Wins \newline (MEDerr) } & \multicolumn{1}{c}{0} & \multicolumn{1}{c}{0} & \multicolumn{1}{c}{2} & \multicolumn{1}{c}{2} & \multicolumn{1}{c}{4} & \multicolumn{1}{c}{0} & \multicolumn{1}{c}{1} & \multicolumn{1}{c}{0} & \multicolumn{1}{c}{0} & \multicolumn{1}{c}{0} & \multicolumn{1}{c}{2} & \multicolumn{1}{c}{1} & \multicolumn{1}{c}{8} & \multicolumn{1}{c}{0} \\
\midrule
\multirow{2}{*}{\rotatebox[origin=c]{90}{SHU}} & Av.MEDerr  & 1.43e+09 & 1.43e+09 & 1.43e+09 & 1.43e+09 & 3.70e+05 & 3.36e+05 & \textbf{1.71e+04} & 1.43e+09 & 1.43e+09 & 1.43e+09 & 1.43e+09 & 1.43e+09 & 1.43e+09 & 1.43e+09 \\
& Av.Ranking  & 5.57 & 7.14 & 5.71 & 5.86 & 8.14 & 10.86 & 9.00 & 13.00 & 11.29 & 11.57 & \textbf{5.29} & 8.00 & 6.14 & \textbf{5.29} \\
\midrule
\multirow{2}{*}{\rotatebox[origin=c]{90}{MUL}} & Av.MEDerr  & 9.97e+03 & 9.14e+03 & 1.07e+04 & 1.16e+04 & \textbf{5.03e+03} & 2.00e+09 & 2.69e+05 & 2.00e+09 & 5.77e+04 & 1.10e+04 & 6.28e+03 & 9.76e+03 & 5.88e+03 & 7.51e+03 \\
& Av.Ranking  & 7.91 & 6.00 & 6.73 & 7.64 & 5.36 & 12.45 & 11.73 & 14.00 & 8.73 & 7.09 & 4.09 & 6.91 & \textbf{3.36} & 5.09 \\
\midrule
\multirow{2}{*}{\rotatebox[origin=c]{90}{HYB}} & Av.MEDerr  & 1.11e+09 & 1.11e+09 & 6.82e+03 & 7.36e+03 & \textbf{3.58e+03} & 2.22e+09 & 3.17e+04 & 4.44e+09 & 2.58e+04 & 8.70e+03 & 1.11e+09 & 6.26e+03 & 1.11e+09 & 1.11e+09 \\
& Av.Ranking  & 4.53 & \textbf{4.00} & 4.53 & 6.00 & 4.79 & 8.58 & 8.95 & 10.53 & 8.53 & 8.16 & 8.21 & 9.89 & 8.89 & 10.47 \\
\midrule
\multirow{2}{*}{\rotatebox[origin=c]{90}{ALL}} & Av.MEDerr  & 1.05e+09 & 1.05e+09 & 5.26e+08 & 5.26e+08 & 1.38e+05 & 1.58e+09 & \textbf{8.41e+04} & 3.16e+09 & 5.26e+08 & 5.26e+08 & 1.05e+09 & 5.26e+08 & 1.05e+09 & 1.05e+09 \\
& Av.Ranking  & 7.11 & 6.47 & 6.32 & 7.11 & 6.37 & 11.84 & 10.79 & 13.63 & 9.79 & 8.47 & 4.47 & 7.37 & \textbf{4.26} & 5.11 \\
\midrule
& \multicolumn{1}{c}{Wilcoxon test} & \multicolumn{1}{c}{$\approx$} & \multicolumn{1}{c}{$\approx$} & \multicolumn{1}{c}{$\approx$} & \multicolumn{1}{c}{$\approx$} & \multicolumn{1}{c}{$\approx$} & \multicolumn{1}{c}{$\approx$} & \multicolumn{1}{c}{$\approx$} & \multicolumn{1}{c}{$+$} & \multicolumn{1}{c}{$\approx$} & \multicolumn{1}{c}{$\approx$} & \multicolumn{1}{c}{$\approx$} & \multicolumn{1}{c}{$\approx$} & \multicolumn{1}{c}{} & \multicolumn{1}{c}{$\approx$} \\

\toprule
& {Wins \newline (MAD) } & \multicolumn{1}{c}{4} & \multicolumn{1}{c}{4} & \multicolumn{1}{c}{2} & \multicolumn{1}{c}{3} & \multicolumn{1}{c}{1} & \multicolumn{1}{c}{3} & \multicolumn{1}{c}{0} & \multicolumn{1}{c}{6} & \multicolumn{1}{c}{1} & \multicolumn{1}{c}{1} & \multicolumn{1}{c}{2} & \multicolumn{1}{c}{3} & \multicolumn{1}{c}{2} & \multicolumn{1}{c}{5} \\
\midrule
\multirow{2}{*}{\rotatebox[origin=c]{90}{SHU}} & Av.MAD  & 3.63e+03 & 3.56e+03 & 3.84e+03 & \textbf{2.33e+03} & 2.19e+04 & 3.61e+03 & 3.38e+04 & 7.42e+04 & 2.51e+03 & 4.12e+04 & 5.20e+03 & 2.87e+03 & 3.07e+03 & 4.26e+03 \\
& Av.Ranking  & 5.86 & 5.86 & 6.43 & 6.57 & 10.14 & 10.71 & 11.00 & 12.71 & 9.43 & 10.57 & 6.57 & 6.71 & \textbf{5.14} & 5.29 \\
\midrule
\multirow{2}{*}{\rotatebox[origin=c]{90}{MUL}} & Av.MAD  & 7.63e+02 & 3.40e+02 & 6.62e+02 & \textbf{3.29e+02} & 6.57e+02 & 1.16e+03 & 7.70e+04 & 5.80e+03 & 9.19e+03 & 3.69e+03 & 6.54e+02 & 5.94e+02 & 6.35e+02 & 6.99e+02 \\
& Av.Ranking  & 6.82 & 5.73 & 5.73 & 6.00 & 8.91 & 8.18 & 12.18 & 9.27 & 10.18 & 9.00 & \textbf{5.27} & 7.55 & 6.55 & 5.45 \\
\midrule
\multirow{2}{*}{\rotatebox[origin=c]{90}{HYB}} & Av.MAD  & 4.00e+02 & 2.00e+02 & 3.76e+02 & \textbf{1.98e+02} & 9.69e+02 & 3.12e+03 & 9.03e+03 & 3.76e+04 & 4.94e+03 & 2.26e+03 & 3.77e+02 & 3.81e+02 & 3.91e+02 & 4.00e+02 \\
& Av.Ranking  & \textbf{3.16} & 3.63 & 3.89 & 5.37 & 7.11 & 7.32 & 9.05 & 8.00 & 9.68 & 9.21 & 8.37 & 10.58 & 10.37 & 10.21 \\
\midrule
\multirow{2}{*}{\rotatebox[origin=c]{90}{ALL}} & Av.MAD  & 1.55e+03 & 1.41e+03 & 1.60e+03 & \textbf{9.56e+02} & 8.53e+03 & 3.02e+03 & 3.51e+04 & 4.65e+04 & 5.52e+03 & 1.64e+04 & 2.10e+03 & 1.24e+03 & 1.32e+03 & 1.76e+03 \\
& Av.Ranking  & 6.58 & 5.84 & 6.00 & 6.42 & 9.32 & 9.42 & 11.79 & 10.11 & 9.95 & 9.74 & 5.58 & 7.05 & 5.84 & \textbf{5.42} \\
\midrule
& \multicolumn{1}{c}{Wilcoxon test} & \multicolumn{1}{c}{$\approx$} & \multicolumn{1}{c}{$\approx$} & \multicolumn{1}{c}{$\approx$} & \multicolumn{1}{c}{$\approx$} & \multicolumn{1}{c}{$\approx$} & \multicolumn{1}{c}{$\approx$} & \multicolumn{1}{c}{$+$} & \multicolumn{1}{c}{$\approx$} & \multicolumn{1}{c}{$\approx$} & \multicolumn{1}{c}{$\approx$} & \multicolumn{1}{c}{$\approx$} & \multicolumn{1}{c}{$\approx$} & \multicolumn{1}{c}{$\approx$} & \multicolumn{1}{c}{} \\
    \end{longtable} }
\end{ThreePartTable}

%% file: tables/tbl_LD_4.tex
\begin{ThreePartTable}
\setlength{\tabcolsep}{0.5pt} 
    {\fontsize{6.5}{6.2}\selectfont
    \begin{longtable}{m{0.1cm} C{1.25cm} m{0.88cm} m{0.88cm} m{0.88cm} m{0.88cm} m{0.88cm} m{0.88cm} m{0.88cm} m{0.88cm} m{0.88cm} m{0.88cm} m{0.88cm} m{0.88cm} m{0.88cm} m{0.88cm}} 
    \caption{Average median (Av.MED), average median error (Av.MEDerr) and average median absolute deviation (Av.MAD) obtained by the algorithms on the 19 functions of the SOCO'10 test suite with $\mathbf{d = 1250}$.}\label{table:MED-MEDerr-MAD-LD-1250}\\
    \toprule
    & $f\#$ &  
    \textbf{\MTFtfoTBL} &  \textbf{\MTFldoTBL} & 
    \textbf{\CMAEStfoTBL} & \textbf{\CMAESldoTBL} & 
    \textbf{\DEtfoTBL} & \textbf{\DEldoTBL} & 
    \textbf{\PSOtfoTBL} & \textbf{\PSOldoTBL} & 
    \textbf{\PDHybtfoTBL} & \textbf{\PDHybldoTBL} & 
    \textbf{\DCHybtfoTBL} & \textbf{\DCHybldoTBL} &
    \textbf{\PCHybtfoTBL} & \textbf{\PCHybldoTBL} \\
    \midrule
    \endfirsthead
    
    \caption*{ Table \ref{table:MED-MEDerr-MAD-LD-1250} (Continued.)}\\
    \toprule
    & $f\#$ &  
    \textbf{\MTFtfoTBL} &  \textbf{\MTFldoTBL} & 
    \textbf{\CMAEStfoTBL} & \textbf{\CMAESldoTBL} & 
    \textbf{\DEtfoTBL} & \textbf{\DEldoTBL} & 
    \textbf{\PSOtfoTBL} & \textbf{\PSOldoTBL} & 
    \textbf{\PDHybtfoTBL} & \textbf{\PDHybldoTBL} & 
    \textbf{\DCHybtfoTBL} & \textbf{\DCHybldoTBL} &
    \textbf{\PCHybtfoTBL} & \textbf{\PCHybldoTBL} \\
    \midrule
    \endhead
    
    \bottomrule
    \endfoot
    
    \bottomrule
    \endlastfoot

& {Wins \newline (MED) } & \multicolumn{1}{c}{3} & \multicolumn{1}{c}{3} & \multicolumn{1}{c}{2} & \multicolumn{1}{c}{1} & \multicolumn{1}{c}{2} & \multicolumn{1}{c}{0} & \multicolumn{1}{c}{1} & \multicolumn{1}{c}{0} & \multicolumn{1}{c}{0} & \multicolumn{1}{c}{0} & \multicolumn{1}{c}{4} & \multicolumn{1}{c}{0} & \multicolumn{1}{c}{8} & \multicolumn{1}{c}{1} \\
\midrule
\multirow{2}{*}{\rotatebox[origin=c]{90}{SHU}} & Av.MED  & 1.43e+09 & 1.43e+09 & 1.43e+09 & 1.43e+09 & \textbf{1.05e+06} & 1.43e+09 & 1.43e+09 & 1.52e+06 & 1.43e+09 & 1.43e+09 & 1.43e+09 & 1.43e+09 & 1.43e+09 & 1.43e+09 \\
& Av.Ranking  & \textbf{4.86} & 5.00 & 5.86 & 5.71 & 8.29 & 12.00 & 9.71 & 11.00 & 10.57 & 11.14 & 5.29 & 12.71 & 6.86 & 5.43 \\
\midrule
\multirow{2}{*}{\rotatebox[origin=c]{90}{MUL}} & Av.MED  & 1.32e+05 & 1.31e+05 & 1.33e+05 & 1.35e+05 & 9.16e+05 & 2.00e+09 & 1.92e+08 & 2.00e+09 & 1.72e+08 & 4.26e+08 & \textbf{1.27e+05} & 2.00e+09 & \textbf{1.27e+05} & 1.29e+05 \\
& Av.Ranking  & 5.91 & 4.73 & 6.91 & 7.00 & 7.09 & 11.91 & 9.73 & 12.73 & 10.09 & 10.18 & 3.55 & 13.36 & \textbf{3.45} & 4.73 \\
\midrule
\multirow{2}{*}{\rotatebox[origin=c]{90}{HYB}} & Av.MED  & 1.11e+09 & 1.11e+09 & 1.11e+09 & 1.11e+09 & \textbf{9.11e+06} & 2.22e+09 & 2.47e+07 & 3.33e+09 & 1.28e+09 & 1.17e+09 & 1.11e+09 & 3.33e+09 & 1.11e+09 & 1.11e+09 \\
& Av.Ranking  & \textbf{3.32} & \textbf{3.32} & 4.74 & 5.53 & 5.74 & 8.26 & 8.05 & 9.95 & 9.53 & 10.11 & 7.74 & 12.53 & 8.74 & 10.16 \\
\midrule
\multirow{2}{*}{\rotatebox[origin=c]{90}{ALL}} & Av.MED  & 1.05e+09 & 1.05e+09 & 1.05e+09 & 1.05e+09 & \textbf{4.91e+06} & 2.11e+09 & 5.89e+08 & 2.11e+09 & 1.18e+09 & 1.19e+09 & 1.05e+09 & 2.63e+09 & 1.05e+09 & 1.05e+09 \\
& Av.Ranking  & 5.53 & 4.79 & 6.42 & 6.53 & 7.58 & 11.89 & 9.84 & 12.16 & 10.21 & 10.53 & \textbf{4.11} & 13.11 & 4.58 & 4.89 \\
\midrule
& \multicolumn{1}{c}{Wilcoxon test} & \multicolumn{1}{c}{$\approx$} & \multicolumn{1}{c}{$\approx$} & \multicolumn{1}{c}{$\approx$} & \multicolumn{1}{c}{$\approx$} & \multicolumn{1}{c}{$\approx$} & \multicolumn{1}{c}{$\approx$} & \multicolumn{1}{c}{$\approx$} & \multicolumn{1}{c}{$\approx$} & \multicolumn{1}{c}{$+$} & \multicolumn{1}{c}{$+$} & \multicolumn{1}{c}{} & \multicolumn{1}{c}{$+$} & \multicolumn{1}{c}{$\approx$} & \multicolumn{1}{c}{$\approx$} \\

\toprule
& {Wins \newline (MEDerr) } & \multicolumn{1}{c}{3} & \multicolumn{1}{c}{3} & \multicolumn{1}{c}{2} & \multicolumn{1}{c}{1} & \multicolumn{1}{c}{2} & \multicolumn{1}{c}{0} & \multicolumn{1}{c}{1} & \multicolumn{1}{c}{0} & \multicolumn{1}{c}{0} & \multicolumn{1}{c}{0} & \multicolumn{1}{c}{4} & \multicolumn{1}{c}{0} & \multicolumn{1}{c}{8} & \multicolumn{1}{c}{1} \\
\midrule
\multirow{2}{*}{\rotatebox[origin=c]{90}{SHU}} & Av.MEDerr  & 1.43e+09 & 1.43e+09 & 1.43e+09 & 1.43e+09 & \textbf{9.61e+05} & 1.43e+09 & 1.43e+09 & 1.43e+06 & 1.43e+09 & 1.43e+09 & 1.43e+09 & 1.43e+09 & 1.43e+09 & 1.43e+09 \\
& Av.Ranking  & \textbf{4.86} & 5.00 & 5.86 & 5.71 & 8.29 & 12.00 & 9.71 & 11.00 & 10.57 & 11.14 & 5.29 & 12.71 & 6.86 & 5.43 \\
\midrule
\multirow{2}{*}{\rotatebox[origin=c]{90}{MUL}} & Av.MEDerr  & 8.06e+03 & 6.40e+03 & 8.84e+03 & 1.04e+04 & 7.92e+05 & 2.00e+09 & 1.92e+08 & 2.00e+09 & 1.72e+08 & 4.26e+08 & 2.89e+03 & 2.00e+09 & \textbf{2.88e+03} & 4.15e+03 \\
& Av.Ranking  & 5.91 & 4.73 & 6.91 & 7.00 & 7.09 & 11.91 & 9.73 & 12.73 & 10.09 & 10.18 & 3.55 & 13.36 & \textbf{3.45} & 4.73 \\
\midrule
\multirow{2}{*}{\rotatebox[origin=c]{90}{HYB}} & Av.MEDerr  & 1.11e+09 & 1.11e+09 & 1.11e+09 & 1.11e+09 & \textbf{9.04e+06} & 2.22e+09 & 2.46e+07 & 3.33e+09 & 1.28e+09 & 1.17e+09 & 1.11e+09 & 3.33e+09 & 1.11e+09 & 1.11e+09 \\
& Av.Ranking  & \textbf{3.32} & \textbf{3.32} & 4.74 & 5.53 & 5.74 & 8.26 & 8.05 & 9.95 & 9.53 & 10.11 & 7.74 & 12.53 & 8.74 & 10.16 \\
\midrule
\multirow{2}{*}{\rotatebox[origin=c]{90}{ALL}} & Av.MEDerr  & 1.05e+09 & 1.05e+09 & 1.05e+09 & 1.05e+09 & \textbf{4.84e+06} & 2.11e+09 & 5.89e+08 & 2.11e+09 & 1.18e+09 & 1.19e+09 & 1.05e+09 & 2.63e+09 & 1.05e+09 & 1.05e+09 \\
& Av.Ranking  & 5.53 & 4.79 & 6.42 & 6.53 & 7.58 & 11.89 & 9.84 & 12.16 & 10.21 & 10.53 & \textbf{4.11} & 13.11 & 4.58 & 4.89 \\
\midrule
& \multicolumn{1}{c}{Wilcoxon test} & \multicolumn{1}{c}{$\approx$} & \multicolumn{1}{c}{$\approx$} & \multicolumn{1}{c}{$\approx$} & \multicolumn{1}{c}{$\approx$} & \multicolumn{1}{c}{$\approx$} & \multicolumn{1}{c}{$\approx$} & \multicolumn{1}{c}{$\approx$} & \multicolumn{1}{c}{$\approx$} & \multicolumn{1}{c}{$+$} & \multicolumn{1}{c}{$+$} & \multicolumn{1}{c}{} & \multicolumn{1}{c}{$+$} & \multicolumn{1}{c}{$\approx$} & \multicolumn{1}{c}{$\approx$} \\

\toprule
& {Wins \newline (MAD) } & \multicolumn{1}{c}{4} & \multicolumn{1}{c}{3} & \multicolumn{1}{c}{2} & \multicolumn{1}{c}{4} & \multicolumn{1}{c}{0} & \multicolumn{1}{c}{4} & \multicolumn{1}{c}{2} & \multicolumn{1}{c}{4} & \multicolumn{1}{c}{2} & \multicolumn{1}{c}{2} & \multicolumn{1}{c}{3} & \multicolumn{1}{c}{5} & \multicolumn{1}{c}{4} & \multicolumn{1}{c}{6} \\
\midrule
\multirow{2}{*}{\rotatebox[origin=c]{90}{SHU}} & Av.MAD  & 3.35e+04 & 2.12e+04 & 4.09e+04 & 1.92e+04 & 4.03e+04 & 5.88e+04 & 3.73e+04 & 3.00e+05 & 2.50e+04 & 5.18e+04 & 1.83e+04 & 6.53e+06 & \textbf{5.53e+03} & 2.39e+04 \\
& Av.Ranking  & 5.14 & 5.86 & 8.57 & \textbf{4.00} & 9.57 & 11.00 & 8.86 & 10.71 & 9.57 & 9.57 & 6.29 & 13.00 & 6.29 & 6.14 \\
\midrule
\multirow{2}{*}{\rotatebox[origin=c]{90}{MUL}} & Av.MAD  & 4.87e+02 & 4.55e+02 & 1.07e+03 & 8.30e+02 & 6.42e+05 & 5.13e+04 & 4.86e+08 & 8.09e+03 & 9.98e+08 & 4.18e+08 & \textbf{4.16e+02} & 7.39e+05 & 1.00e+04 & 7.14e+02 \\
& Av.Ranking  & 5.18 & 4.73 & 6.64 & 5.55 & 10.18 & 9.91 & 10.45 & 8.18 & 11.00 & 10.82 & 6.18 & 10.18 & 6.64 & \textbf{4.36} \\
\midrule
\multirow{2}{*}{\rotatebox[origin=c]{90}{HYB}} & Av.MAD  & 5.66e+02 & \textbf{4.03e+02} & 4.19e+03 & 6.78e+02 & 1.46e+08 & 6.44e+04 & 3.05e+07 & 4.67e+04 & 2.82e+08 & 8.04e+08 & 4.63e+02 & 6.84e+05 & 5.58e+03 & 4.84e+02 \\
& Av.Ranking  & \textbf{2.95} & 3.37 & 4.63 & 4.84 & 7.16 & 7.74 & 8.32 & 8.00 & 9.79 & 10.42 & 9.16 & 11.47 & 9.79 & 10.05 \\
\midrule
\multirow{2}{*}{\rotatebox[origin=c]{90}{ALL}} & Av.MAD  & 1.26e+04 & 8.00e+03 & 1.71e+04 & 7.39e+03 & 6.94e+07 & 6.37e+04 & 1.42e+08 & 1.35e+05 & 3.96e+08 & 4.91e+08 & 7.00e+03 & 2.92e+06 & \textbf{4.78e+03} & 9.04e+03 \\
& Av.Ranking  & 5.05 & 5.16 & 7.37 & \textbf{4.89} & 9.95 & 10.42 & 10.00 & 8.74 & 10.47 & 10.26 & 6.21 & 11.42 & 6.58 & \textbf{4.89} \\
\midrule
& \multicolumn{1}{c}{Wilcoxon test} & \multicolumn{1}{c}{$\approx$} & \multicolumn{1}{c}{$\approx$} & \multicolumn{1}{c}{$\approx$} & \multicolumn{1}{c}{} & \multicolumn{1}{c}{$+$} & \multicolumn{1}{c}{$\approx$} & \multicolumn{1}{c}{$+$} & \multicolumn{1}{c}{$\approx$} & \multicolumn{1}{c}{$+$} & \multicolumn{1}{c}{$+$} & \multicolumn{1}{c}{$\approx$} & \multicolumn{1}{c}{$\approx$} & \multicolumn{1}{c}{$\approx$} & \multicolumn{1}{c}{$\approx$} \\
& \multicolumn{1}{c}{Wilcoxon test} & \multicolumn{1}{c}{$\approx$} & \multicolumn{1}{c}{$\approx$} & \multicolumn{1}{c}{$\approx$} & \multicolumn{1}{c}{$\approx$} & \multicolumn{1}{c}{$\approx$} & \multicolumn{1}{c}{$\approx$} & \multicolumn{1}{c}{$\approx$} & \multicolumn{1}{c}{$\approx$} & \multicolumn{1}{c}{$\approx$} & \multicolumn{1}{c}{$+$} & \multicolumn{1}{c}{$\approx$} & \multicolumn{1}{c}{$\approx$} & \multicolumn{1}{c}{$\approx$} & \multicolumn{1}{c}{} \\

    \end{longtable} }
\end{ThreePartTable}